%% file: cpal_2025.tex
\documentclass{article}

\usepackage[preprint]{cpal_2025}

\usepackage[unicode]{hyperref}
\usepackage{url}
\usepackage{booktabs}       
\usepackage{amsfonts}       
\usepackage[pdftex]{graphicx}
\usepackage{natbib}
\usepackage{amsmath}
\usepackage{bbm}
\usepackage{algorithm}
\usepackage{algorithmic}
\usepackage{wrapfig}
\usepackage{makecell}
\usepackage{caption}
\usepackage{subcaption}
\usepackage{siunitx}
\usepackage{tablefootnote}
\usepackage{lipsum}

\usepackage[textsize=tiny]{todonotes}

\newcommand{\nl}{\\[2pt]}

\title{What Scalable Second-Order Information Knows for Pruning at Initialization}

\author{Ivo Gollini Navarrete, Nicolás Mauricio Cuadrado Ávila, Martin Takáč \& Samuel Horváth \\
Mohamed bin Zayed University of Artificial Intelligence \\
Abu Dhabi, UAE \\
\texttt{\{ivo.navarrete,nicolas.avila\}@mbzuai.ac.ae} \\
}

\begin{document}
\maketitle

\input{sections/00_abstract}
\input{sections/01_introduction}
\input{sections/02_background}
\input{sections/03_method}
\input{sections/04_results}
\input{sections/05_conclusion}
\input{sections/06_end_of_document}
\appendix
\input{sections/07_appendix}

\end{document}

%% file: sections/00_abstract.tex
\begin{abstract}
Pruning remains an effective strategy for reducing both the costs and environmental impact associated with deploying large neural networks (NNs) while maintaining performance. Classical methods, such as OBD \cite{lecun1989optimal} and OBS \cite{hassibi1992second}, demonstrate that utilizing curvature information can significantly enhance the balance between network complexity and performance. However, the computation and storage of the Hessian matrix make it impractical for modern NNs, motivating the use of approximations.
Recent research \cite{gur2018gradient, karakida2019universal} suggests that the top eigenvalues guide optimization in a small subspace, are identifiable early, and remain consistent during training. Motivated by these findings, we revisit pruning at initialization (PaI) to evaluate scalable, unbiased second-order approximations, such as the Empirical Fisher and Hutchinson diagonals. Our experiments show that these methods capture sufficient curvature information to improve the identification of critical parameters compared to first-order baselines, while maintaining linear complexity.
Additionally, we empirically demonstrate that updating batch normalization statistics as a warmup phase improves the performance of data-dependent criteria and mitigates the issue of layer collapse. Notably, Hutchinson-based criteria consistently outperformed or matched existing PaI algorithms across various models (including VGG, ResNet, and ViT) and datasets (such as CIFAR-10/100, TinyImageNet, and ImageNet).
Our findings suggest that scalable second-order approximations strike an effective balance between computational efficiency and accuracy, making them a valuable addition to the pruning toolkit. We make our code available \footnote{\url{https://github.com/Gollini/Scalable_Second_Order_PaI}}.
\end{abstract}

%% file: sections/01_introduction.tex
\section{Introduction}
Advancements in computing and data availability have enabled large Neural Networks (NNs) to achieve exceptional progress in fields such as robotics \citep{soori2023artificial}, computer vision \citep{khan2020machine}, and natural language processing \citep{torfi2020natural}. However, the increasing size of these models raises concerns about computing costs and energy consumption \citep{han2015learning}, making them difficult to implement on edge devices or in resource-constrained settings \citep{cheng2024survey}. In addition, large-scale training and inference have become another key focus area in climate change awareness \citep{wu2022sustainable, chien2023reducing}.
\nl
Model compression techniques, such as quantization \citep{dettmers2023qlora}, low-rank factorization \citep{denton2014exploiting}, knowledge distillation \citep{xu2024survey}, neural architecture search \citep{zhang2021idarts}, and neural network pruning \citep{lecun1989optimal}, aim to reduce model complexity while preserving performance. In particular, neural network pruning effectively decreases the model size and computational workload with minimal performance loss. Cheng et al. \cite{cheng2024survey} classify pruning strategies based on three questions: (1) Is the acceleration universal or specific? (2) When does pruning occur in the training pipeline? (3) Is the pruning criterion predefined or learned during training?
\nl
Researchers mainly focus on pruning after training (PaT) because converged models produce more accurate estimates of parameter importance compared to randomly initialized ones \citep{kumar2024no}. Pruning criteria can range from simple random selection or parameter magnitude \citep{han2015learning} to principled approaches that evaluate changes in loss \cite{singh2020woodfisher}. Classical methods, such as Optimal Brain Damage (OBD) \citep{lecun1989optimal} and Optimal Brain Surgery (OBS) \citep{hassibi1992second}, utilize curvature information to improve the balance between network complexity and performance. However, the large number of parameters in modern NNs makes the computation of the Hessian matrix infeasible, leading to the use of approximations \citep{singh2020woodfisher}.
\nl
The benefits of PaT come at the cost of training a dense model and subsequent parameter adjustments. This has sparked interest in Pruning at Initialization (PaI), which seeks to identify subnetworks before training to save computation. Frankle et al.~\cite{frankle2018lottery} introduced the idea of ``winning tickets,'' sparse subnetworks that can outperform dense models with the same initialization. However, finding these requires a computationally heavy prune-retrain strategy, prompting the search for more efficient algorithms \citep{sreenivasan2022rare, you2022supertickets} and one-shot approaches to reduce overhead \citep{lee2018snip, wang2020picking, tanaka2020pruning}.
\nl
Incorporating second-derivative information directly at initialization seems to be a motivated step. However, there are reservations about relying on information from a randomly initialized model, given the high variance and the mismatch between the information of the early and converged curvature~\cite{liao2021hessian}. As a result, PaI research has favored zero and first-order methods. Nevertheless, the findings of Gur et al.~\cite{gur2018gradient} show that gradient descent rapidly concentrates in a tiny subspace dominated by the top eigenvectors and is preserved throughout training. Furthermore, Karakida et al.~\cite{karakida2019universal} analyzed the spectrum of the Fisher Information Matrix (FIM) at initialization. They determined that for randomly initialized wide networks, the local geometry is dominated by a small number of directions. Together, these results suggest that even at initialization, curvature information already identifies important directions to preserve during training.
\begin{wrapfigure}{R}{0.425\textwidth}  
    \centering
    \includegraphics[width=\linewidth]{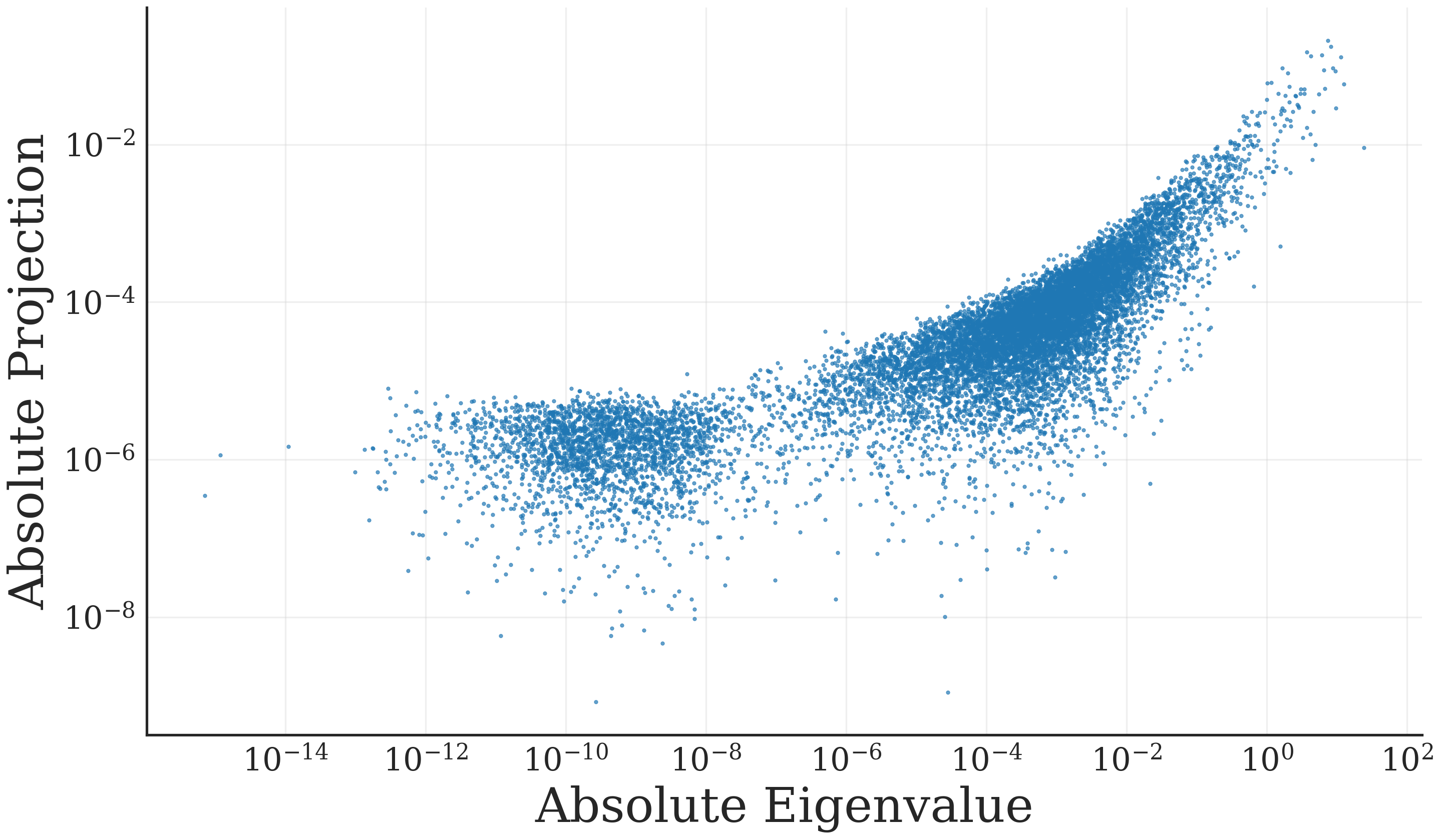}
    \caption{Relationship between curvature and parameter displacement. Each point corresponds to one eigendirection of the Hessian calculated at initialization.}
    \label{fig:spectra}
\end{wrapfigure}

To validate our assumptions, we conducted a small-scale experiment using two classes from the MNIST dataset~\cite{lecun1998mnist}. At the beginning of the experiment, we computed the Hessian and obtained its eigendecomposition, \(H_0 = U_0\Sigma_0 U^{T}_{0}\), where \(\Sigma\) is the diagonal matrix of eigenvalues and \(U\) is the orthogonal matrix of corresponding eigenvectors. We then measured the movement of the parameters from initialization to the end of training (\(\Delta w = w_0 - w_t\)). As shown in Figure \ref{fig:spectra}, projecting this displacement onto the Hessian eigenbasis, denoted as \(proj = |U^{T}_{0} \Delta w|\), indicates that the largest movements in parameter space relate to the directions associated with the largest magnitude eigenvalues \(|\lambda|\) of \(H_0\). A detailed description of the experimental setup and results is provided in Appendix \ref{sec:HessianEigenspectrum}, where the reader can observe that this trend emerges after the first optimization step and is maintained throughout the training process.
\nl
Motivated by these findings, we revisit the one-shot PaI setting and examine scalable estimators of the Hessian matrix for pruning. Specifically, we evaluate unbiased estimators of the Hessian diagonal, such as the Empirical FIM diagonal and Hutchinson's estimator diagonal, effectively reducing the computation overhead of second-order methods from quadratic to linear \citep{yao2021adahessian}.
\nl
We also propose a method to alleviate \textit{layer collapse}, a failure mode in \textit{data-dependent} methods that disrupts information flow between layers due to bottlenecks or complete removal of a layer \citep{tanaka2020pruning}. Gradient-based PaI methods often assign disproportionately low scores to wide layers, pruning them first, and making the model untrainable \citep{kumar2024no}. We show that a simple warmup phase to update batch normalization statistics yields more reliable gradient estimates and effectively mitigates layer collapse.
\nl
Our experiments used CIFAR-10/100, TinyImageNet, and ImageNet on a range of models (VGG, ResNet, and ViT). Our results show that coarse second-order information consistently improves pruning outcomes compared to long-standing one-shot pruning baselines. Moreover, the proposed warmup phase reduced the effect of layer collapse and improved the performance of data-dependent methods. Notably, the Hutchinson diagonal approximation with warmup surpasses magnitude-based PaT in extreme sparsities ($\geq 90\%$), a long-standing barrier for PaI methods \citep{frankle2020pruning}.
\nl
Our key contributions are summarized as follows: (1) We revisit one-shot PaI with scalable second-order approximations for a balance between performance and computation overhead. (2) Our experiments show that coarse curvature information outperforms long-standing PaI one-shot baselines. (3) We show that updating batch normalization statistics provides a simple and effective fix to layer collapse and improves the performance of data-dependent criteria. And (4) We empirically demonstrate that principled pruning methods narrow the gap between PaI and PaT. 

%% file: sections/02_background.tex
\section{Background \& Related Work}
\textbf{Problem Definition}.
Given access to the training set $\mathcal{D} = \{(x_n, y_n)\}_{n=1}^N$, composed of tuples of input $x_n \in \mathcal{X}$ and output $y_n \in \mathcal{Y}$, the goal is to learn a model parameterized by $w \in \mathbb{R}^d$ that maps $f: \mathcal{X} \rightarrow \mathcal{Y}$ by minimizing the objective function:
\begin{align}
    \textstyle
    \mathcal{L}(w) = \frac{1}{N} \sum_{n=1}^N l(y_n, f(x_n;w)).
    \label{eq:obj_func}
\end{align}
We denote $q \in \mathcal{Q}$, with $\mathcal{Q} = \{1,2, \dots,d\}$, as an index to refer to an element $w_q$ in the parameter vector $w$.

\textbf{The Hessian Matrix}.
Consider the scalar-valued function \eqref{eq:obj_func} is twice differentiable. We refer to the matrix of second partial derivatives as the Hessian:
\begin{align}
    H(w):= \nabla^2 \mathcal{L}(w) \in \mathbb{R}^{d \times d}.
    \label{eq:hessian_matrix}
\end{align}
It captures the local curvature of the loss landscape around $w$, indicating how rapidly the gradient changes in different directions. Direct computation and storage of the entire matrix requires \(\mathcal{O}(d^2)\) time and memory, infeasible for modern networks.
Consequently, many second-order methods rely on approximations, especially in optimization research. Hessian-free methods exploit the Hessian-vector product (HVP), but require many iterations or additional techniques for stability \citep{pearlmutter1994fast, martens2011learning}. Other approaches, such as the Generalized Gauss-Newton (GGN) \citep{schraudolph2002fast}, approximate layer-wise blocks to approximate only parts of the Hessian. The Kronecker-factored Approximate Curvature method (KFAC) reduces the GGN calculation of each block by writing it as a product of two smaller matrices \citep{martens2015optimizing}. However, storing KFAC matrices can become prohibitively expensive for large models. 

A scalable alternative is to approximate only the diagonal of the Hessian. Although the exact calculation of the Hessian diagonal remains quadratic, stochastic methods yield unbiased estimates \citep{elsayed2024revisiting}. For example, Curvature Propagation (CP) \citep{martens2012estimating} provides such an estimator at the cost of two gradient evaluations. More recently, Yao et al. \cite{yao2021adahessian} used Hutchinson's estimator to approximate $\mathrm{diag}(H)$ through randomized probes $z$ drawn from a Rademacher distribution.

\textbf{The Fisher Information Matrix (FIM)}. The FIM has been used as an approximation of the Hessian to increase computational speed \citep{vacar2011langevin}. It is defined as the expectation of the second moment of the score function. Based on the probabilistic concept that minimizing the loss function $l(y, f(x;w)$ is equivalent to maximizing the negative log-likelihood $- \log p(y \mid x, w)$, we can express the FIM in terms of the Hessian under regularity conditions \citep{Schervish2012-xv}:
\begin{align}
    \textstyle 
    F(w) 
    = - \mathbb{E} \left[ 
        \nabla^2 \log p(y \mid x, w)
    \right]
    =\mathbb{E} \left[ 
        \nabla^2 l(y, f(x;w))
    \right]
    .
    \label{eq:fisher_information_hessian}
\end{align}
The FIM approximation reduces computational demands, but still requires \(\mathcal{O}(d^2)\) memory. Soen et al. \cite{soen2024tradeoffs} discussed the trade-offs of using only the diagonal of the FIM, which reduces the complexity to \(\mathcal{O}(d)\). In practice, the FIM diagonal is estimated from the empirical training distribution, providing an unbiased plug-in estimator that retains essential geometric information. This leads to the following formulation:
\begin{align}
    \textstyle 
    \operatorname{diag}(\hat{F})
    = 
    \frac{1}{N} \sum_{n=1}^N 
    \nabla l(y_n, f(x_n;w))^2.
    \label{eq:empirical_fim_diag}
\end{align}
This approximation can be intuitively understood as follows: Each entry in \(\operatorname{diag}(\hat{F})\) is the average of the squared gradient of the model's output with respect to a parameter. Larger entries indicate parameters that have a greater influence on the model's output.

\textbf{PaT Methods.}
Early pruning focused on removing parameters with low performance impact (low saliency). First, Hanson et al. \cite{hanson1988comparing} found that parameters with larger magnitudes are generally more significant. Then, LeCun et al. \cite{lecun1989optimal} introduced OBD, a more principled saliency measure that uses a second-order Taylor series to assess the effect of removing parameters under three assumptions: (1) the Hessian matrix is approximated using only its diagonal, (2) the first-order term of the Taylor series is negligible for a converged model, and (3) the local loss model is assumed to be quadratic, discarding higher-order terms. The saliency of the parameter $w_q$ is defined as:
\begin{align}
    \textstyle 
    s_q = \frac{1}{2} w_q^2 H_{qq},
    \label{eq:obd_final}
\end{align}
A few years later, Hassibi et al. \citep{hassibi1992second} presented OBS, highlighting the importance of a more comprehensive representation of second-order information, which includes off-diagonal elements, and criticized the need to fine-tune the subnetwork.
They derived a general expression for saliency that includes \eqref{eq:obd_final} as a special case, and an expression to recalculate the magnitude of all parameters after removing a parameter $w_q$:
\begin{align}
    \textstyle
    s_q = \frac{w_q^2}{2[H^{-1}]_{qq}},
    ~~~&~~~
    \textstyle
    \delta w = -\frac{w_q H^{-1} e_q}{[H^{-1}]_{qq}}.
    \label{eq:obs_final}
\end{align}
More recently, Theis et al. \citep{theis2018faster} adapted OBD to the size of current models by approximating the Hessian diagonal using the empirical FIM diagonal. As in \eqref{eq:obd_final}, the first term of the Taylor expansion vanishes, yielding the saliency metric Fisher Pruning (FP):
\begin{align}
    \textstyle 
    s_q = \frac{1}{2} w_q^2 \hat{F}_{qq}. 
    \label{eq:faster_gaze_fisher}
\end{align}
Extending FP to structured pruning, Liu et al.~\cite{liu2021group} employed Fisher information to estimate the importance of channels identified by a layer grouping algorithm that exploits the network's computation graph. Singh et al. \citep{singh2020woodfisher} adapted OBS to modern times with an iterative blockwise method to approximate the inverse of the Hessian. The authors demonstrated the relationship between the empirical FIM inverse $\hat{F}^{-1}$ and the Hessian inverse $H^{-1}$, concluding that the former is a good approximation of the latter as long as the application is scale-invariant. 
 
\textbf{PaI Methods.}
When referring to methods that prune at initialization time, it is necessary to mention the \textit{Lottery Ticket Hypothesis} introduced by Frankle et al. \citep{frankle2018lottery}. Certain initializations can reveal smaller subnetworks that match or surpass the performance of the original dense network. Their iterative pruning approach identified performant subnetworks in ResNet18 and VGG19 with compression rates of $80-90\%$ for the CIFAR-10 classification task. However, they required a computationally expensive process, opening the question: If winning tickets exist, can we find them inexpensively?

Lee et al. \citep{lee2018snip} addressed this question with SNIP, a single-shot pruning method that measures \textit{connection sensitivity}. They introduce an auxiliary indicator vector $c \in \{0,1\}^d$ that specifies whether a connection is active with a Hadamard product $c \odot w$. From this perspective, the effect of removing the connection $q$ on the loss is approximated as a directional derivative $g_q(w;\mathcal{D})$ with respect to $c_q$:
\begin{align}
    \textstyle 
    \Delta \mathcal{L}_q(w) 
    = 
    \lim_{\epsilon \rightarrow 0} \frac{\mathcal{L}(c \odot w) - \mathcal{L}((c - \epsilon e_q) \odot w)}{\epsilon}\Big\rvert_{c = \mathbbm{1}}
    = 
    w_q \frac{\partial \mathcal{L}(w)}{\partial c_q} = g_q(w).
    \label{eq:snip_prev}
\end{align}
This method can be intuitively understood as follows: The magnitude of the gradient with respect to $c$ indicates how each connection affects the loss, regardless of the direction. The sensitivity is defined as:
\begin{align}
    \textstyle 
    s_q = \frac{|g_q(c \odot w)|}{\sum_{k=1}^{m} |g_k(c \odot w)|}.
    \label{eq:snip}
\end{align}
Wang et al. \citep{wang2020picking} proposed Gradient Signal Preservation (GraSP), a method based on the concept of neural tangent kernel (NTK) \citep{jacot2018neural}, under the premise that effective training requires preserving gradient flow through the model. GraSP selects parameters that encourage the NTK to be large in the direction corresponding to the gradients of the output space. They derived a sensitivity metric that measures the response to a stimulus $\delta$:
\begin{align}
    \textstyle 
    S(\delta) 
    = \Delta \mathcal{L}(w_0 + \delta) - \Delta \mathcal{L}(w_0) 
    = 2\delta^\top H \nabla \mathcal{L}(w) + \mathcal{O}(||\delta||_2^2).
\end{align}
Tanaka et al. \citep{tanaka2020pruning} proposed SynFlow, a data-agnostic pruning method that identifies sparse, trainable subnetworks at initialization. Unlike data-dependent methods, which can induce \textit{layer collapse} by removing entire layers, SynFlow iteratively preserves the total “synaptic flow,” i.e., the cumulative strength of connections, thereby maintaining information flow and network trainability.

%% file: sections/03_method.tex
\section{Methodology}
\textbf{Setting Definition}.
Following the taxonomy in \cite{wang2022recentadvancesneuralnetwork}, the main setting focuses on unstructured pruning. A binary mask $m \in \{0,1\}^d$ is applied to induce sparsity, effectively reducing the parameter count of the model. We construct $m$ using an importance criterion and the target sparsity level. The pruned model is defined as $f(x_n;m \odot w)$, where $\odot$ denotes the Hadamard product between $m$ and the model weights $w$. The objective \eqref{eq:obj_func} then becomes:
\begin{align}
    \textstyle 
    \mathcal{L}(m \odot w) = \frac{1}{N} \sum_{n=1}^N l(y_n, f(x_n; m \odot w)).
    \label{eq:mask_obj_func}
\end{align}

\textbf{Sensitivity Score.}
Following OBD \citep{lecun1989optimal}, we start by approximating the objective \eqref{eq:obj_func} using a Taylor series. The perturbation $\delta w$ of the parameter vector will change $\mathcal{L}$ by
\begin{align}
    \textstyle
    \delta \mathcal{L} 
    = \mathcal{L}(w) - \mathcal{L}(w + \delta w)
    = \delta w^T \nabla \mathcal{L}(w) + \frac{1}{2} \delta w^T H \delta w + \mathcal{O}(||\delta w||^3).
    \label{eq:taylor_series}
\end{align}
Unlike OBD, the gradients are far from zero in the PaI setting, and neglecting the first-order term is not viable. Still, we operate under the assumption that the local error is quadratic, discarding higher-order components. This approach assumes that individually deleting the parameters yields the same perturbation as removing them simultaneously. 
We aim to maintain a linear complexity of $O(d)$ to make the sensitivity score practical and minimize computational overhead. Therefore, we restrict the computation of the Hessian to diagonal approximations that capture enough curvature information. Then, \eqref{eq:taylor_series} becomes:
\begin{align}
    \delta \mathcal{L} 
    = \sum_{q \in \mathcal{Q}} \delta w_q \frac{\partial \mathcal{L}(w)}{\partial w_q} 
    + \frac{1}{2} \sum_{q \in \mathcal{Q}} \delta w_q^2 H_{qq}.
    \label{eq:diag_taylor_series_sum}
\end{align}
As explained in \cite{lee2018snip}, induced perturbations at initialization will cause $\mathcal{L}$ to increase, decrease, or remain the same. Changes of high magnitude to \eqref{eq:diag_taylor_series_sum} (positive or negative) mean that the objective function is significantly \textit{sensitive} to the parameters $q$ and should be preserved for the pruned model to learn. 
We define the sensitivity of the parameter $w_q$ as follows:
\begin{align}
\textstyle
    s_q = \left| w_q \frac{\partial \mathcal{L}(w)}{\partial w_q} + \frac{1}{2} w_{q}^{2} H_{qq} \right|.
    \label{eq:sensitivity}
\end{align}

\textbf{Scalable Hessian approximations}.
Firstly, we consider the Hutchinson approximation that AdaHessian employs \citep{yao2021adahessian}. We sample a random Rademacher vector $z \in \{-1,1\}^d$ and compute the HVP $Hz$. In practice, the computation of the HVP is performed via a single backpropagation trick, thereby keeping the cost on the order of a gradient pass. The elementwise product of $Hz$ and $z$, averaged over random draws, yields an unbiased estimator of the Hessian diagonal:
\begin{align}
    \operatorname{diag}(\hat{H}) = \mathbb{E}_{z}\!\big[(Hz) \odot z\big].   
    \label{eq:hutch_diag}
\end{align}
With the incorporation of $\operatorname{diag}(\hat{H})$ into \eqref{eq:sensitivity}, we define Hutchinson-Taylor Sensitivity (HTS) as our first parameter importance score:
\begin{align}
\textstyle
    s_q = \left| w_q \frac{\partial \mathcal{L}(w)}{\partial w_q} + \frac{1}{2} w_q^2 \hat{H}_{qq} \right|.
    \label{eq:hutch_taylor_sensitivity}
\end{align}

Secondly, the equivalence between FIM and Hessian described in \eqref{eq:fisher_information_hessian} only holds in convergence when the parameter vector $w$ maximizes the likelihood $\mathbb{E}[\nabla \log p(y \mid x, w)] = 0$. However, Karikada et al. \cite{karakida2019universal} showed that even at initialization, the FIM captures essential geometric properties of the parameter space. Some FIM eigenvalues are close to zero and indicate local flatness, while others are significantly large and induce substantial distortions in specific directions. We test whether the signals from the empirical FIM diagonal are strong enough to identify important parameters. In addition, the FIM has the desirable property of being PSD by construction, ensuring a stable representation. We define Fisher-Taylor Sensitivity (FTS) incorporating $\operatorname{diag}(\hat{F})$ into \eqref{eq:sensitivity}:
\begin{align}
\textstyle
    s_q = \left| w_q \frac{\partial \mathcal{L}(w)}{\partial w_q} + \frac{1}{2} w_q^2 \hat{F}_{qq} \right|.
    \label{eq:fisher_taylor_sensitivity}
\end{align}

\textbf{Baselines comparison and Taylor Series Ablation}.
For comparison, we evaluate our proposed criteria against the following pruning methods: random, parameter magnitude, gradient norm (GN), SNIP \cite{lee2018snip}, GraSP \cite{wang2020picking}, and SynFlow \cite{tanaka2020pruning}. Additionally, we decompose the different elements of \eqref{eq:sensitivity} to gain a better understanding of the elements of the Taylor expansion.
We do not isolate the first-order component since Wang et al. \citep{wang2020picking} show that SNIP is equivalent to computing its absolute value:
\begin{align*}
    s_q = \left| w_q \frac{\partial \mathcal{L}(w)}{\partial w_q} \right|.
\end{align*}
For the second-order component, we extract two extra sensitivity criteria. First, we directly evaluate the diagonal approximations of the Hessian, referring to them as Hutchinson Diagonal (HD) and Fisher Diagonal (FD):
\begin{align}
    \textstyle 
    s_q = \hat{H}_{qq}, 
    ~~~&~~~
    s_q = \hat{F}_{qq}.
\end{align}
Second, we evaluate the effect of using only the second-order term as in OBD \citep{lecun1989optimal} or Fisher Pruning \citep{theis2018faster}, referring to them as Hutchinson Pruning (HP) and Fisher Pruning (FP):
\begin{align}
    \textstyle 
    s_q = w_{q}^{2}\hat{H}_{qq}, 
    ~~~&~~~
    s_q = w_{q}^{2}\hat{F}_{qq}.
\end{align}

\textbf{Pruning Mask}.
Given a data set partition, we compute the vector $s$ that contains the sensitivity scores $s_q$ for each parameter $w_q$. We generate the saliency scores using a subset of the training set as in \cite{frankle2020pruning}. To create the pruning mask $m$, we define a percentile $p$ to narrow the subset containing the parameter index to retain:
\[
    \textstyle
    \mathcal{R} = \{ q \mid s_q \text{ is in the top } (1 - p) \text{ of scores} \}.
\]
Using this subset $\mathcal{R}$, the elements of the binary mask $m$ are defined using the following rule: $m_q = 1$ if $q \in \mathcal{R}$ and $m_q = 0$ otherwise. We produce the pruned model $f(x; m \odot w_0)$ with the Hadamard product between the binary mask $m$ and the vector of the initial parameters of the model $w_0$, with the sparsity ratio defined as:
\[
    \textstyle
    \text{sparsity} = \frac{1}{d}\sum_{q} m_q,
\]
where $d$ is the total number of parameters of the dense model. Once the mask is applied, the pruned model is optimized utilizing stochastic gradient descent to minimize the objective function \eqref{eq:mask_obj_func}. It is important to note that our pruning setting excludes parameters from batch normalization and the output layers, as we consider them crucial for enabling learning and performing the designed task. We also skip bias parameters, as they are initialized to zero and would be immediately pruned. 

%% file: sections/04_results.tex
\section{Results and Discussion}
\label{sec:results}
\begin{figure*}[tb]
    \centering
    \begin{subfigure}{0.49\textwidth}
        \includegraphics[width=\linewidth]{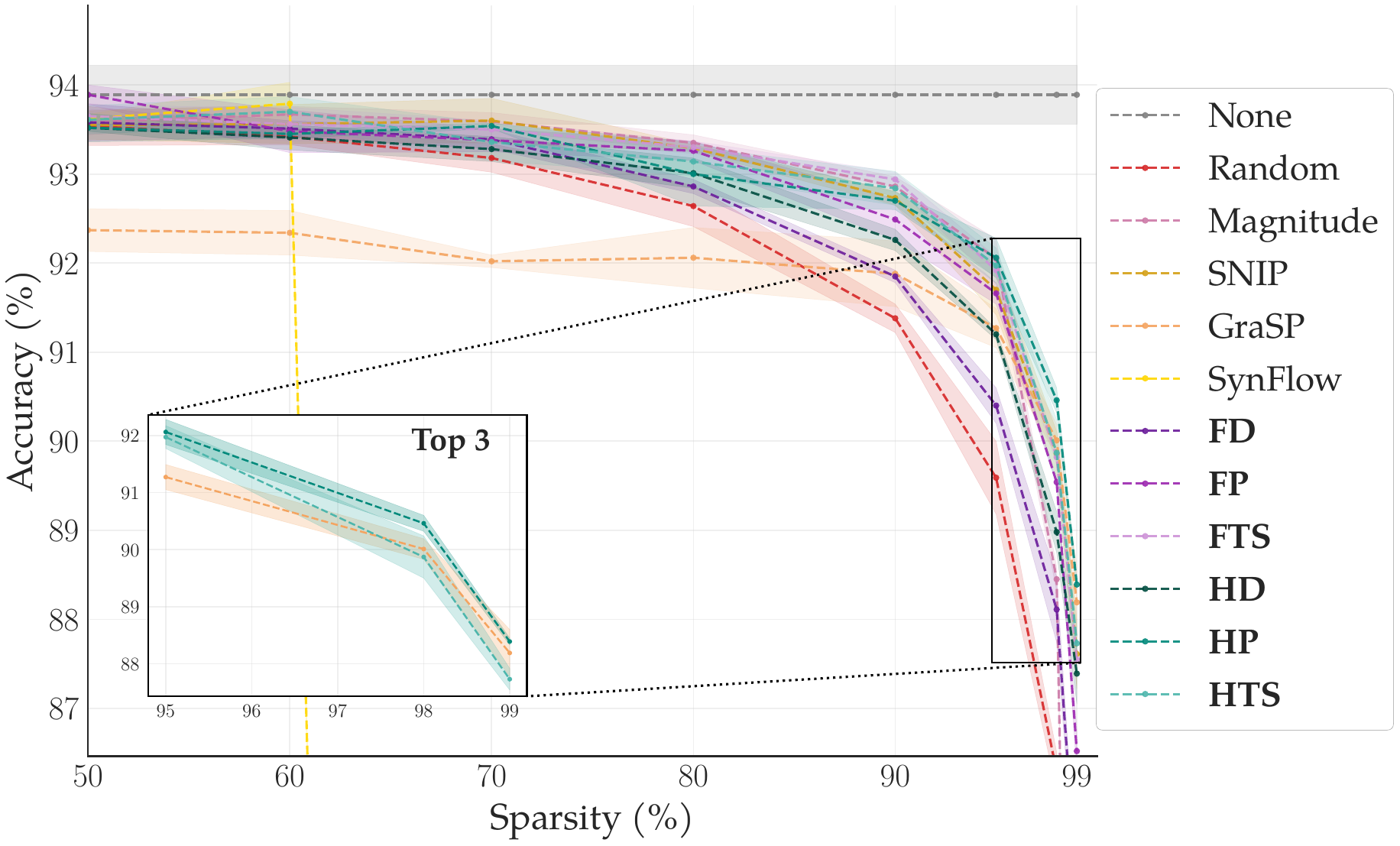}
        \caption{ResNet18/CIFAR-10 without warmup.}
        \label{fig:ResNet_18_CIFAR_10_w=0}
    \end{subfigure}
    \hfill
    \hfill
    \begin{subfigure}{0.49\textwidth}
        \includegraphics[width=\linewidth]{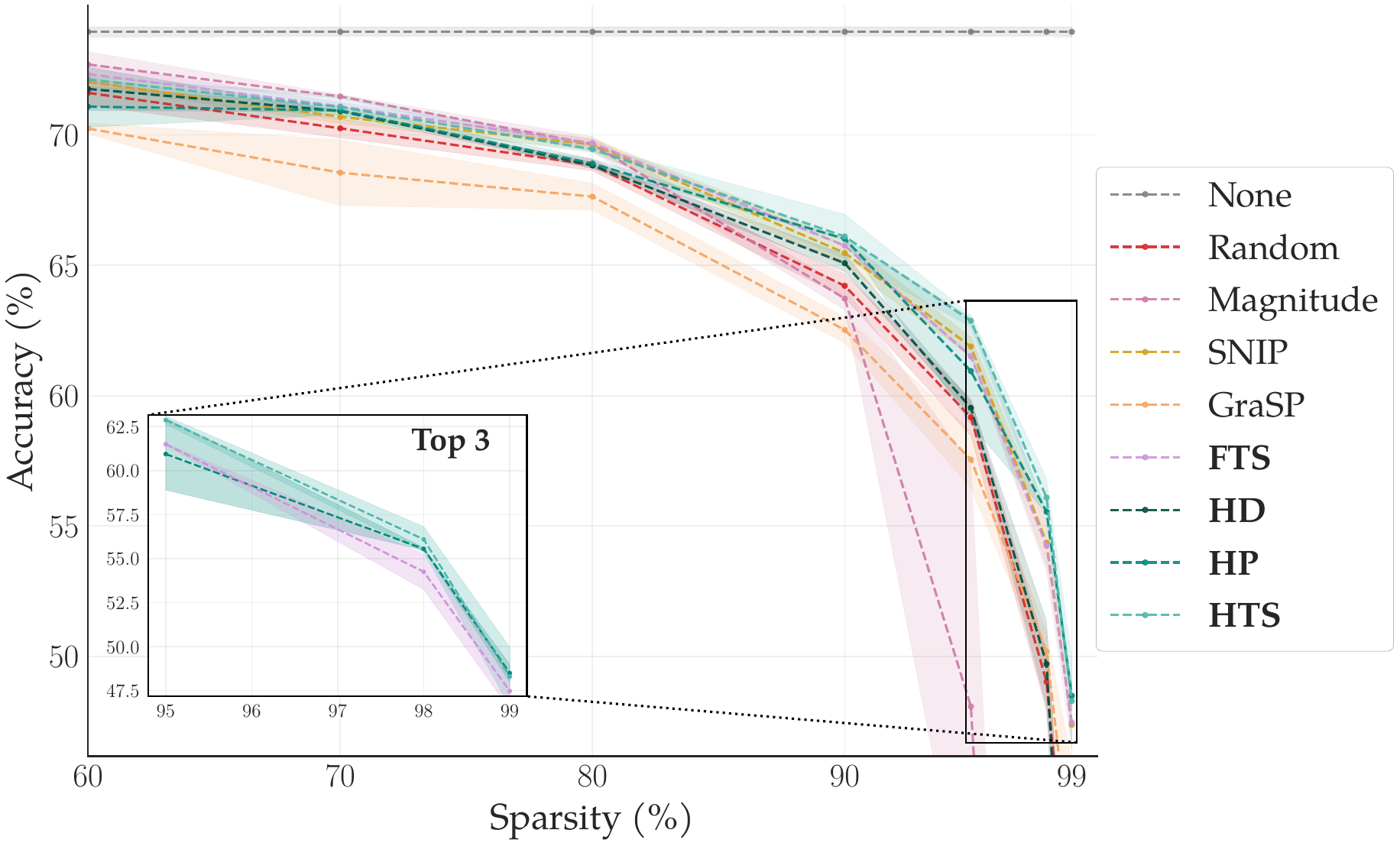}
        \caption{ResNet50/Imagenet-1K without warmup.}
        \label{fig:ResNet_50_ImageNet}
    \end{subfigure}
    \caption{Test accuracy across sparsity levels under various PaI methods for our base reference case of CIFAR-10 with ResNet18 on the left, and the more complex Imagenet-1K with ResNet50. The dashed gray line denotes the baseline accuracy without pruning.}
    \label{fig:test_accuracies_plots}
\end{figure*}
\begin{figure*}[t]
    \centering
    \begin{subfigure}{0.49\textwidth}
        \includegraphics[width=0.99\linewidth]{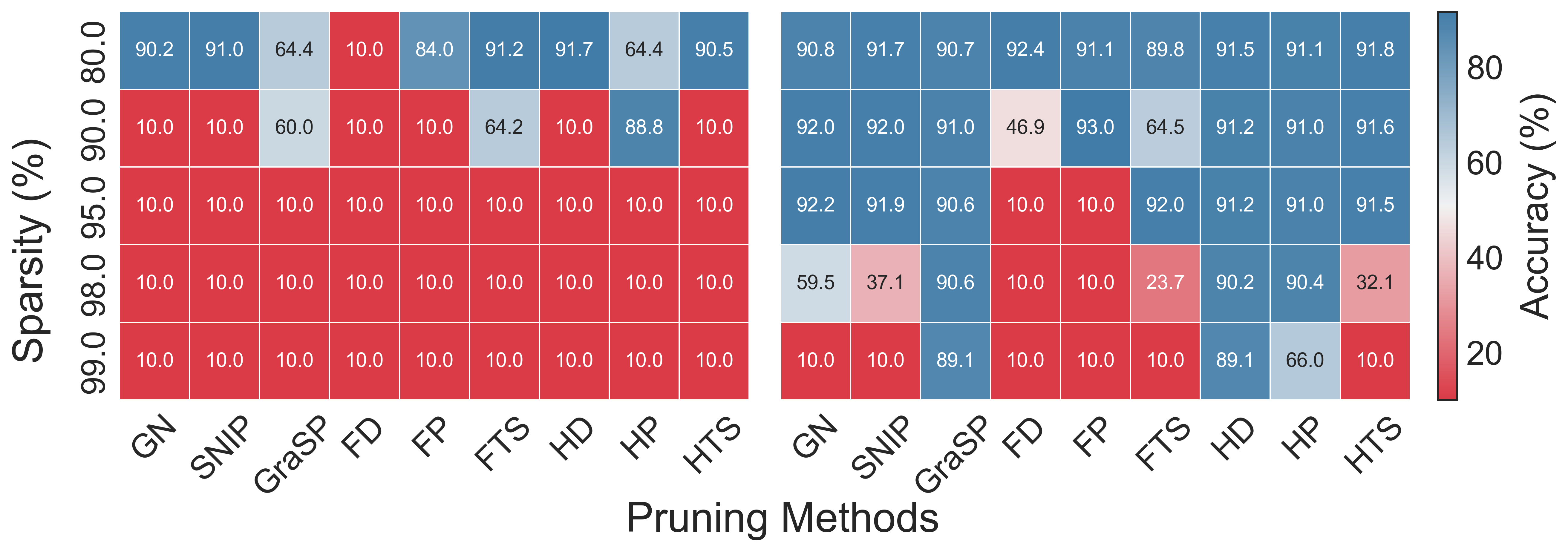}
        \caption{}
        \label{fig:vgg_exps_heatmaps}
    \end{subfigure}
    \hfill
    \begin{subfigure}{0.49\textwidth}
        \includegraphics[width=\linewidth]{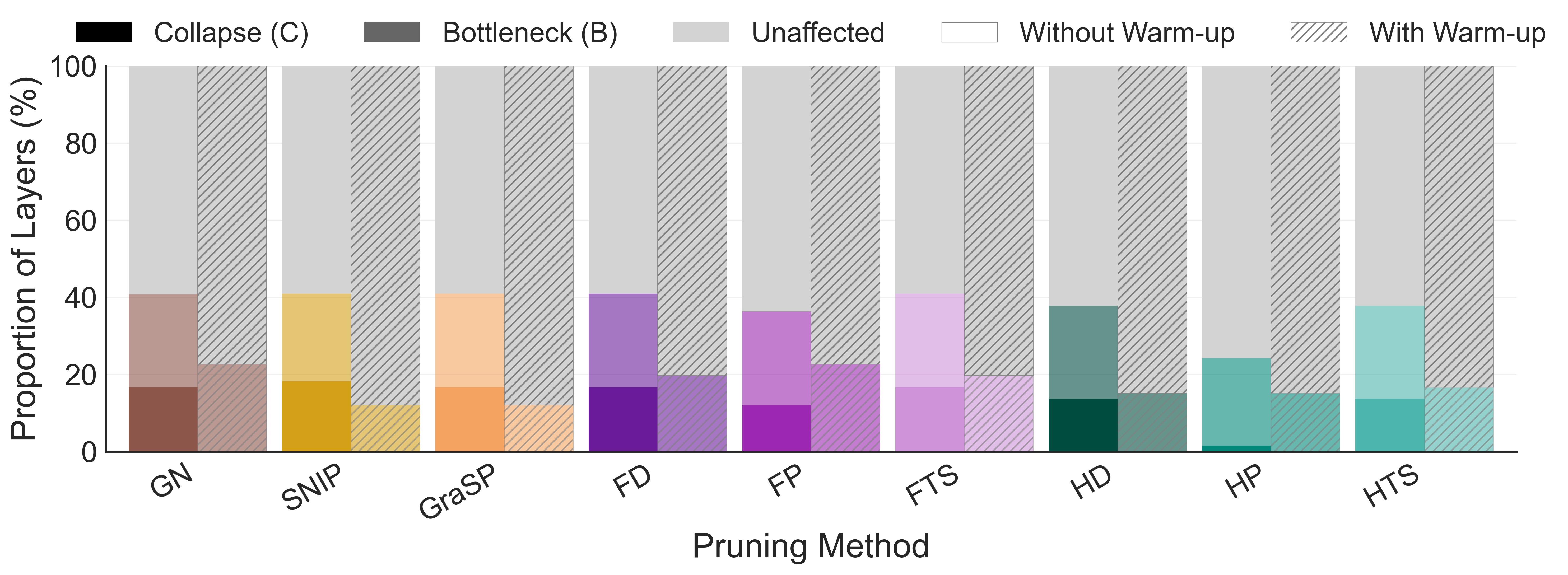}
        \caption{}
        \label{fig:layer_collapse_analysis}
    \end{subfigure}
    \caption{\textbf{(a)} Effect of a warmup phase on training stability in CIFAR-10 with VGG19 at extreme sparsity ratios. Several methods exhibit systematic layer collapse without a warmup (left), leading to near-random performance. A warmup phase (right) largely prevents collapse and preserves accuracy across methods. \textbf{(b)} Importance of batch normalization statistics (BNS) update to prevent layer collapse. We studied the case of VGG19 pruning at $95\%$ sparsity.}
    \label{fig:side_by_side}
    \vskip -15pt
\end{figure*}

We progressively evaluate the effectiveness of the presented pruning criteria by increasing model and task complexities in various setups using benchmarks commonly employed in the unstructured pruning literature \citep{frankle2020pruning}: CIFAR-10/100 \citep{krizhevsky2009learning}, TinyImageNet \citep{deng2009imagenet}, and ImageNet-1K \cite{deng2009imagenet} datasets in ResNet \citep{he2016deep} and VGG \citep{simonyan2014very} architectures (training details in Appendix \ref{appendix:training_parameters}). Additionally, we evaluated the performance of our proposed criteria in a pre-finetuning structured pruning setup for the visual transformer ViT-B/16 \citep{dosovitskiy2021imageworth16x16words}. We report all metrics across three random seeds.
\nl
\textbf{Computational Complexity}.
First, we compare the pruning criteria by complexity. Table \ref{tab:pruning_complexity} (Appendix \ref{appendix:computational_complexity}) details the practical and theoretical time complexity, as well as the space complexity.
The proposed diagonal estimations reduce the complexity of the Hessian matrix to $\mathcal{O}(d)$, and reduce the computational cost to just $\mathcal{O}(Nd)$, the same as first-order methods such as SNIP or GN.
\nl
\textbf{Base Reference Case}.
We consider the CIFAR-10 classification with ResNet18 as our base reference case. As seen in Table \ref{tab:resnet18_cifar10_compressors_warmup_0} (Appendix~\ref{appendix:CIFAR10_ResNet18}), random pruning is a trivial baseline that any method should surpass. Up to $70\%$ sparsity, random's performance is comparable to baseline accuracy ($93.89$). After that point, the accuracy degrades rapidly as we increase the sparsity, as shown in Figure \ref{fig:ResNet_18_CIFAR_10_w=0}. Therefore, our analysis focuses on the high sparsity regime ($\geq 80\%$), where the differences and limitations of the methods are most pronounced.
\nl
Magnitude pruning is a second point of reference, as it is the most widely studied method in the literature. It performs well up to $80\%$ sparsity, where it is the peak performer among the evaluated criteria. However, its performance decays rapidly as the sparsity increases. We observe a similar behavior for GN, SNIP, and GraSP, but with a slower decay, surpassing both basic baselines. The data-agnostic method SynFlow struggles to pass from low sparsities, after which a sudden drop of performance is observed\footnote{Reported issues in the repository of the original implementation highlight inconsistencies due to zero gradients that prune whole layers.}. We highlight that across the sparsity range, Hutchinson-based criteria are top-performers or matching metrics. Most importantly, HP dominates the high-sparsity regime, suggesting that even coarse second derivative information improves the trade-off between model complexity and performance in PaI. We refer to Appendix \ref{appendix:data_estimate_score} for a sensitivity analysis for the amount of data used to estimate scores and the effect of the number of Rademacher probes on ranking stability for Hutchinson diagonal approximation.
\nl
\textbf{Increasing Model Complexity}.
We evaluated the consistency of our results by replacing the model in our reference case with ResNet50 (Table \ref{tab:resnet50_cifar10_compressors_warmup_0} in Appendix~\ref{appendix:CIFAR10_ResNet50}), a deeper and wider architecture commonly used in large-scale vision tasks. Similarly, the performance of random pruning is comparable to the $93.17$ baseline accuracy up to $70\%$ sparsity. Magnitude pruning suffered from layer collapse as the model size increased. GN, SNIP, and GraSP showed a behavior consistent with the base case. SynFlow is the best-performing model for low sparsities, but the same failure mode appears as the sparsity increases. The HP and FTS criteria consistently appear among the top performers at high sparsity. Results continue to support the relevance of second-order information.
\nl
\textbf{Increasing Task Complexity}.
We increase the difficulty of the classification task by evaluating the CIFAR-100 and TinyImageNet datasets with ResNet-18. We observe the same trends, with the HP criterion notoriously outperforming in the high sparsity regime on both datasets, as shown in Figure \ref{fig:ResNet_18_CIFAR_100} and Table \ref{tab:resnet18_cifar100_compressors_warmup_0} for CIFAR-100 (Appendix \ref{sec:resnet_cifar-100}), and Figure \ref{fig:ResNet_18_TinyImageNet} and Table \ref{tab:resnet18_tinyimagenet_tinyimagenet_compressors_warmup_0} for TinyImageNet (Appendix \ref{appendix:TinyImageNet_ResNet18}), respectively. Notably, as task complexity increases, the advantage of Hutchinson-based criteria becomes more noticeable. For CIFAR-100, HP's accuracy increases $\approx2$ points with respect to the closest non-Hutchinson method. For TinyImageNet, the gap increases by $\approx3$ points.
\nl
\textbf{Increasing Model and Task Complexity}. 
Finally, for the last set of experiments in the unstructured PaI domain, we increased the complexity by evaluating ImageNet-1K with ResNet-50, resulting in a significant increase in the number of classes. As seen in Figure \ref{fig:ResNet_50_ImageNet} and Table \ref{tab:resnet50_imagenet_compressors_warmup_0} (Appendix \ref{appendix:ImageNet}), HTS and HP are the constant top-performers across the evaluated sparsities.
\nl
\textbf{Consolidated Analysis}. We empirically showed that Hutchinson's diagonal approximation improves the trade-off between sparsity and performance at initialization. While the theoretical motivation for the Taylor expansion suggests the inclusion of first-order terms (HTS/FTS), the empirical strength of HP suggests that the high variance and poor alignment of gradients at initialization might destabilize the first-order contribution. Also, Fisher-based criteria lagged behind Hutchinson variants. The reliance on solely first-order statistics, as defined in Equation \eqref{eq:empirical_fim_diag}, suggests the FIM diagonal variants might be better suited for later stages of training. For example, near convergence, where both the FIM and the Hessian often become diagonal dominant, as shown in \cite{singh2020woodfisher}, and the FIM more closely approximates the Hessian, as formalized in Equation \eqref{eq:fisher_information_hessian}. Regardless, our results align with the observations of \cite{yvinec2022singe}, which indicated that magnitude-based pruning may remove low-magnitude parameters that contribute to performance. In Appendix \ref{appendix:parameter_selection_comparison}, we illustrate the difference in parameter selection between magnitude-based and principled approaches, showing how magnitude might not be the best indication of importance.
\nl 
\textbf{Preventing Layer Collapse.} 
Table \ref{tab:vgg19_cifar10_compressors_warmup_0} (Appendix~\ref{appendix:CIFAR10_VGG19}) and Table \ref{tab:vgg19_cifar100_compressors_warmup_0} (Appendix~\ref{appendix:vgg_cifar-100}) show the performance of different pruning criteria evaluated in CIFAR-10 and CIFAR-100, respectively, with VGG19. All data-dependent methods suffer a drastic performance drop at higher sparsities. This behavior is consistent with the \textit{layer collapse} phenomenon described by Tanaka et al. \citep{tanaka2020pruning}, where pruning removes entire layers (or most of their parameters), disrupting the flow of information, and rendering the model untrainable. We propose updating the batch normalization statistics before forming the pruning mask as a warm-up phase.
\nl
Figure \ref{fig:vgg_exps_heatmaps} illustrates how our proposed warmup step can mitigate layer collapse. Notably, the methods that can mitigate layer collapse up to the highest sparsity rely solely on second-order information (GraSP, HD, HP), further supporting the interpretation that second-order signals prove a more stable criterion in the PaI setting. Figure \ref{fig:layer_collapse_analysis} presents a visual representation of the results in Table \ref{tab:bn_layer_collapse} (Appendix~\ref{appendix:bn_study}). It is clearly observed that at a $95\%$ sparsity, the update of the batch-norm statistics reduces the number of collapsed layers to $0\%$ and significantly reduces the percentage of bottlenecks, enabling the training of the pruned model. We refer the reader to Tables \ref{tab:vgg19_cifar10_compressors_warmup_1} (Appendix~\ref{appendix:CIFAR10_VGG19}) and Table \ref{tab:vgg19_cifar100_compressors_warmup_1} (Appendix~\ref{appendix:vgg_cifar-100}) for detailed results on CIFAR-10 and CIFAR-100, respectively. 
\nl
\textbf{Improving Gradient Estimation.} 
We further analyze the effect of the warmup step using ResNet18 in the base reference case with CIFAR-10 (Table \ref{tab:resnet18_cifar10_compressors_warmup_1} in Appendix \ref{appendix:CIFAR10_ResNet18}), and the increased complexity tasks with CIFAR-100 and TinyImageNet (Table \ref{tab:resnet18_cifar100_compressors_warmup_1} in Appendix \ref{sec:resnet_cifar-100} and Table \ref{tab:resnet18_tinyimagenet_tinyimagenet_compressors_warmup_1} in Appendix \ref{appendix:TinyImageNet_ResNet18}, respectively). We observe a constant improvement across all gradient-dependent methods. In particular, the HD criterion significantly improves the estimation of the Hessian diagonal, becoming the top performer across the three experiments. Performance increased by $1.57$, $2.39$, and $1.96$ for sparsity ratios $95\%$, $98\%$, and $99\%$, respectively. This simple yet effective step not only mitigates the effect of layer collapse but also enables more accurate gradient estimation at initialization.
\nl
\textbf{PaI vs. PaT comparison}.
Frankle et al. \cite{frankle2020pruning} strongly criticized PaI methodologies for consistently underperforming compared to magnitude-based PaT. However, their analysis did not assess how the PaI-designed criteria perform in the post-training setting. Table \ref{tab:resnet18_cifar10_pat_compressors}  (Appendix~\ref{appendix:PAT_CIFAR10_ResNet18}) evaluates all criteria in the reference case under the same PaT retraining protocol as \cite{frankle2020pruning}. Our results show that principled criteria outperform magnitude pruning at higher sparsities, with HP consistently ranking as a top performer. Notably, FIM-based variants greatly improve in the PaT setting, even match the Hutchinson-based criteria. This aligns with our interpretation that the FIM diagonal limits in PaI are not intrinsic but rather stage-dependent.
\nl
In Table \ref{tab:resnet18_cifar10_pat_vs_pbt_delta} (Appendix~\ref{appendix:PAT_CIFAR10_ResNet18}), we visualize the performance gain of the PaT setting over the PaI setting with warmup. GN and FD are the methods that benefit the most from the PaT setting, given the complete overcoming of layer collapse. Nevertheless, as we increase the sparsity for magnitude-based pruning, it becomes clear that the advantage over PaI methodologies stems from training a fully dense model and then retraining, rather than the inherent strength of the method. For example, the performance gain at $99\%$ sparsity is greater than 12 points. Most notably, the warmup phase significantly improves the HD criterion, enabling the PaI method to slightly outperform its own PaT counterpart at the highest sparsities, and even exceeding the performance of magnitude-based PaT. These results solidify the value of scalable second-order information and provide insights into potentially breaking the long-standing wall for PaI methods without training a fully dense model.
\nl
\begin{wrapfigure}{R}{0.5\textwidth}  
    \centering
    \includegraphics[width=\linewidth]{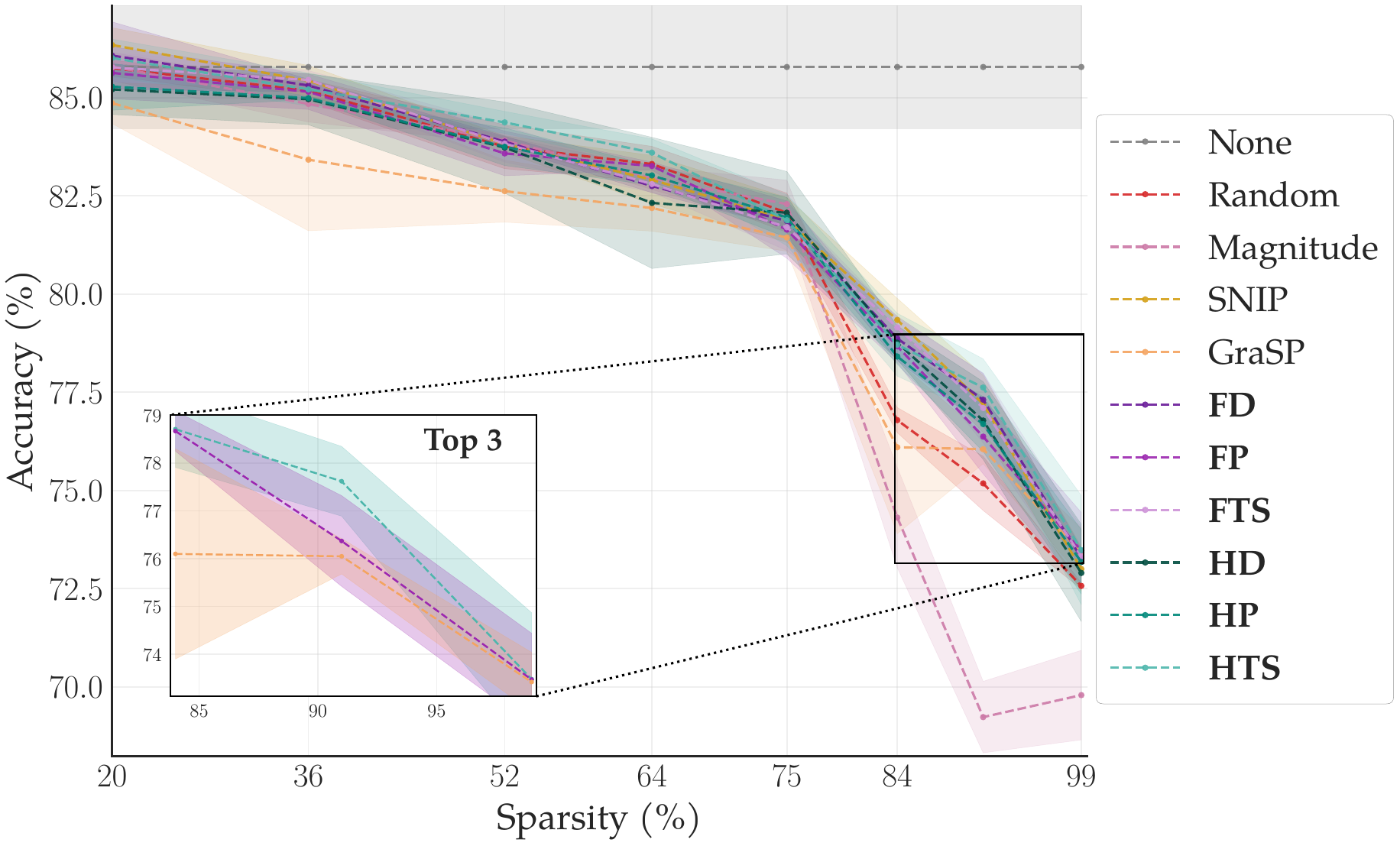}
    \caption{Structured pruning of ViT-B/16 in CIFAR-10: Test accuracy across sparsities for all the importance-based metrics.}
    \label{fig:vit_cifar10_lineplot}
\end{wrapfigure} 
\textbf{Pre-Finetune Pruning for Transformers}. 
Most PaI methods assume that unstructured sparsity is a suitable setup for any model. However, this becomes problematic in transformer-based architectures \cite{cheng2024survey}. We evaluate CIFAR-10 with ViT-B/16 to determine whether our proposed pruning metrics remain informative in a structured setting where decisions operate over larger computationally meaningful units rather than individual weights. 
Transformers are known to be data-hungry and it is not possible to achieve state-of-the-art performance on CIFAR-10 training from scratch. Therefore, we follow \cite{dosovitskiy2021imageworth16x16words} and define a pre-finetune pruning setting for our transformer experiments. Specifically, we pruned the default Torch initialization, which is pretrained on a different distribution (ImageNet-1K), and fine-tuned the subnetwork for 30 epochs. More details in the Appendix \ref{appendix:structured}.
\nl
As seen in Figure \ref{fig:vit_cifar10_lineplot}, at low and moderate sparsity levels, most criteria perform similarly, suggesting that structured decisions are relatively trivial in this regime. As we increase target sparsity, first- and second-order criteria exhibit greater robustness to degradation, consistently achieving higher accuracy. We highlight that, although the performances in our experiment remain very comparable, methods that utilize second-order information, in this case HTS, consistently yield a higher average accuracy for most evaluated sparsity levels. This further solidifies that second-order information at initialization remains relevant in the pruning task. A detailed analysis is available in Appendix \ref{appendix:structured}.

%% file: sections/05_conclusion.tex
\section{Conclusion}
\label{sec:conclusion}
In this work, we validate the assumption that second-order information provides valuable information from a random initialization in a small-scale experiment. Then, we proposed a suite of one-shot PaI methods based on scalable diagonal approximations of second-order information. We demonstrate the effectiveness of approximations in both structured and unstructured PaI and PaT settings, especially in the extreme sparsity regime. The results show that Hutchinson variants consistently outperform or match the SOTA PaI methods across different models, task complexities, and sparsities. We demonstrated that a warmup phase to update batch normalization statistics mitigates layer collapse when using data-dependent methods and enhances the performance of the Hutchinson estimator. Limitations described in Appendix \ref{sec:limitations}.
\nl
Our work contributes to advancing efficient deep learning and resource-aware model deployment. We demonstrate that even coarse second-order methods can reduce the performance gap between PaI and PaT, providing a step toward more efficient and theoretically grounded model compression techniques. We also demonstrated that second-order information is relevant to pruning transformers. Future work includes refining the approximations to capture off-diagonal interactions, as well as exploring the integration with other compression techniques, such as quantization.

%% file: sections/06_end_of_document.tex
\newpage

\bibliography{references}

%% file: sections/07_appendix.tex
\newpage
\appendix
\onecolumn

\renewcommand{\thetable}{A\arabic{table}} 
\renewcommand{\thefigure}{A\arabic{figure}} 
\renewcommand{\theequation}{A\arabic{equation}} 

\setcounter{table}{0} 
\setcounter{figure}{0} 
\setcounter{equation}{0} 

\section*{Appendix}
\input{sections/appendix/00_misc}
\input{sections/appendix/01_hessian_spectrum}
\input{sections/appendix/03_training_details}
\input{sections/appendix/02_complexity}
\input{sections/appendix/16_amount_data_estimate_scores}
\input{sections/appendix/04_bn_study}
\input{sections/appendix/05_cifar10_resnet_18}
\input{sections/appendix/06_cifar10_vgg_19}
\input{sections/appendix/07_cifar10_resnet_50}
\input{sections/appendix/08_cifar100_resnet_18}
\input{sections/appendix/09_cifar100_vgg_19}
\input{sections/appendix/11_tinyimagenet_resnet_18}
\input{sections/appendix/12_imagenet_resnet_50}
\input{sections/appendix/13_pat_resnet_18}
\input{sections/appendix/14_vit_structured}
\input{sections/appendix/15_criteria_vs_magnitude}

%% file: sections/appendix/00_misc.tex
\section{Limitations}
\label{sec:limitations}
(1) The validation of the findings described in \cite{gur2018gradient} was performed in a relatively easy small-scale setting. As complexity increases, the experiments become computationally infeasible to validate and may not be reliable. We consider the extended analysis of this phenomenon an interesting research direction for future work. (2) We operate under the assumption that the FIM and Hutchinson diagonals are good enough approximations of second-order information at initialization. The diagonal approximation assumes non-interaction between parameters at the second-order level score. As we mentioned in the conclusion, future work should expand to better approximations of the Hessian, even if the computational cost is no longer linear. (3) The proposed warmup step can be applied only to architectures with batch normalization layers. While batch normalization layers are integrated into most modern architectures, future research can explore alternatives for architectures that don't incorporate this type of layer.

%% file: sections/appendix/01_hessian_spectrum.tex
\section{Hessian Spectrum}
\label{sec:HessianEigenspectrum}

This section complements the curvature discussion in the main text by showing how parameter updates align with the eigendirections of the Hessian at initialization in a small-scale setting. We consider binary classification in the MNIST~\cite{lecun1998mnist} dataset, restricted to digits $(4,7)$ (mapped to labels 0,1). The data is normalized using the standard MNIST statistics, and we use a batch size of 256. We use a two-layer fully connected network with 12,577 parameters initialized with Xavier-uniform weights and zero biases. We train for $10$ epochs with SGD (learning rate $10^{-4}$, momentum $0.9$, weight decay $10^{-4}$) and Cross-Entropy loss. For clarity, we will stick to the notation in the main body of the manuscript.

\paragraph{Hessian and Eigendecomposition at Initialization.}
At initialization $w_0$, we compute the full empirical Hessian column-wise using Hessian–vector products throughout the training set: for each basis vector $e_i \in \mathbb{R}^P$ we evaluate $H_0 e_i$ by differentiating a scalar product $g(w_0)^\top e_i$, where $g(w_0)$ is the mean loss gradient. The columns are accumulated and symmetrized as $(H_0 + H_0^\top)/2$. We then perform a dense symmetric eigendecomposition using Torch.

\paragraph{Parameter Projection in the Hessian Eigenbasis.}
We compute $\Delta w_t = w_t - w_0$ after the first optimization step and after the termination of each subsequent epoch. For each checkpoint, we
flatten to $\mathbb{R}^P$, and project it onto the eigenbasis of $H_0$ to obtain $z_t = U_0^\top \Delta w_t$. We focus on the absolute coefficients $|z_{t,i}|$, which measure how far training has moved along the eigenvector $i$.

\begin{figure}[t]
    \centering
    \begin{minipage}[b]{0.49\textwidth}
        \centering
        \includegraphics[width=\linewidth]{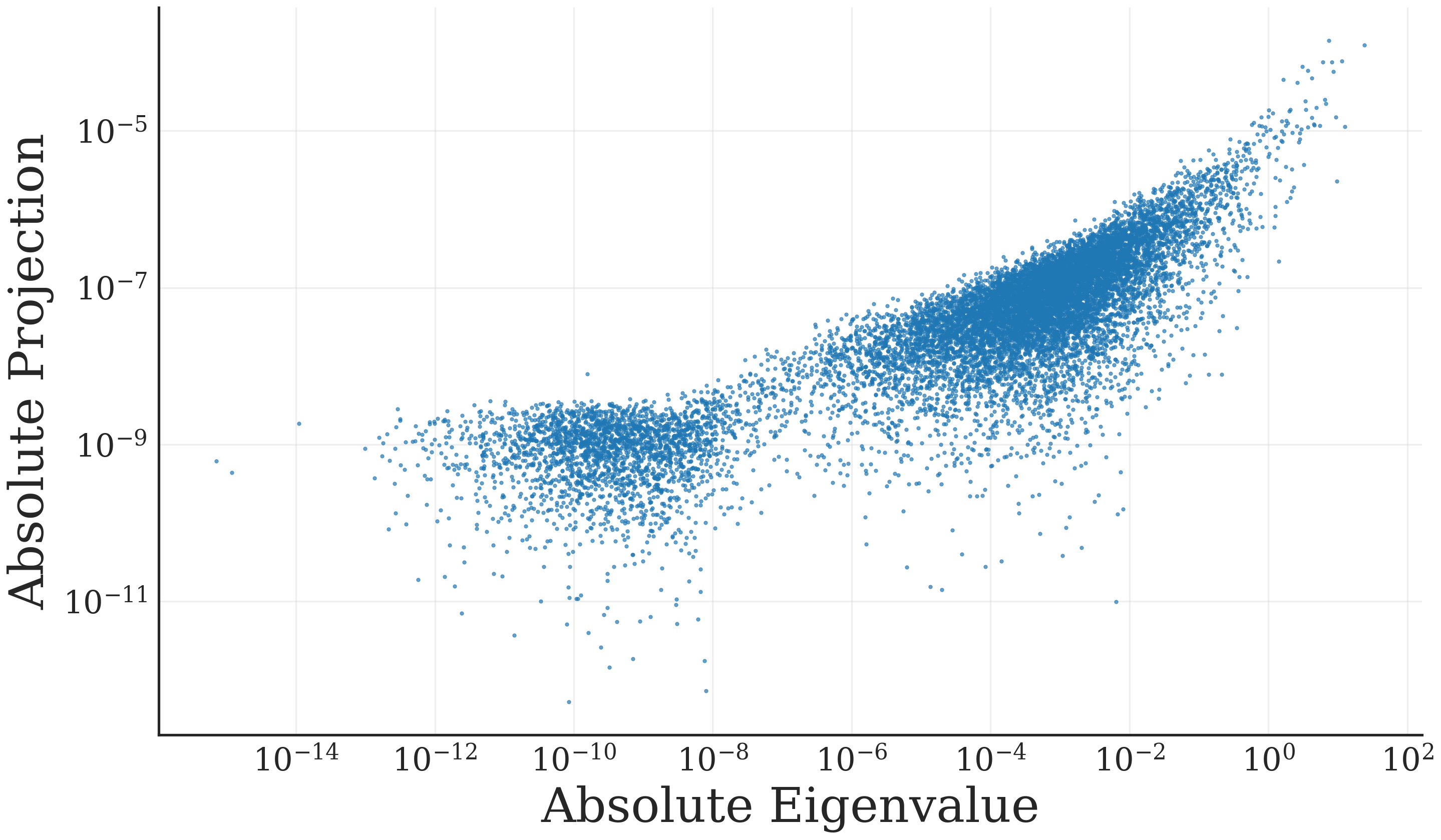}
        \caption*{(a) Projection of $\Delta w$ after the first optimization step.}
    \end{minipage}
    \hfill
    \begin{minipage}[b]{0.49\textwidth}
        \centering
        \includegraphics[width=\linewidth]{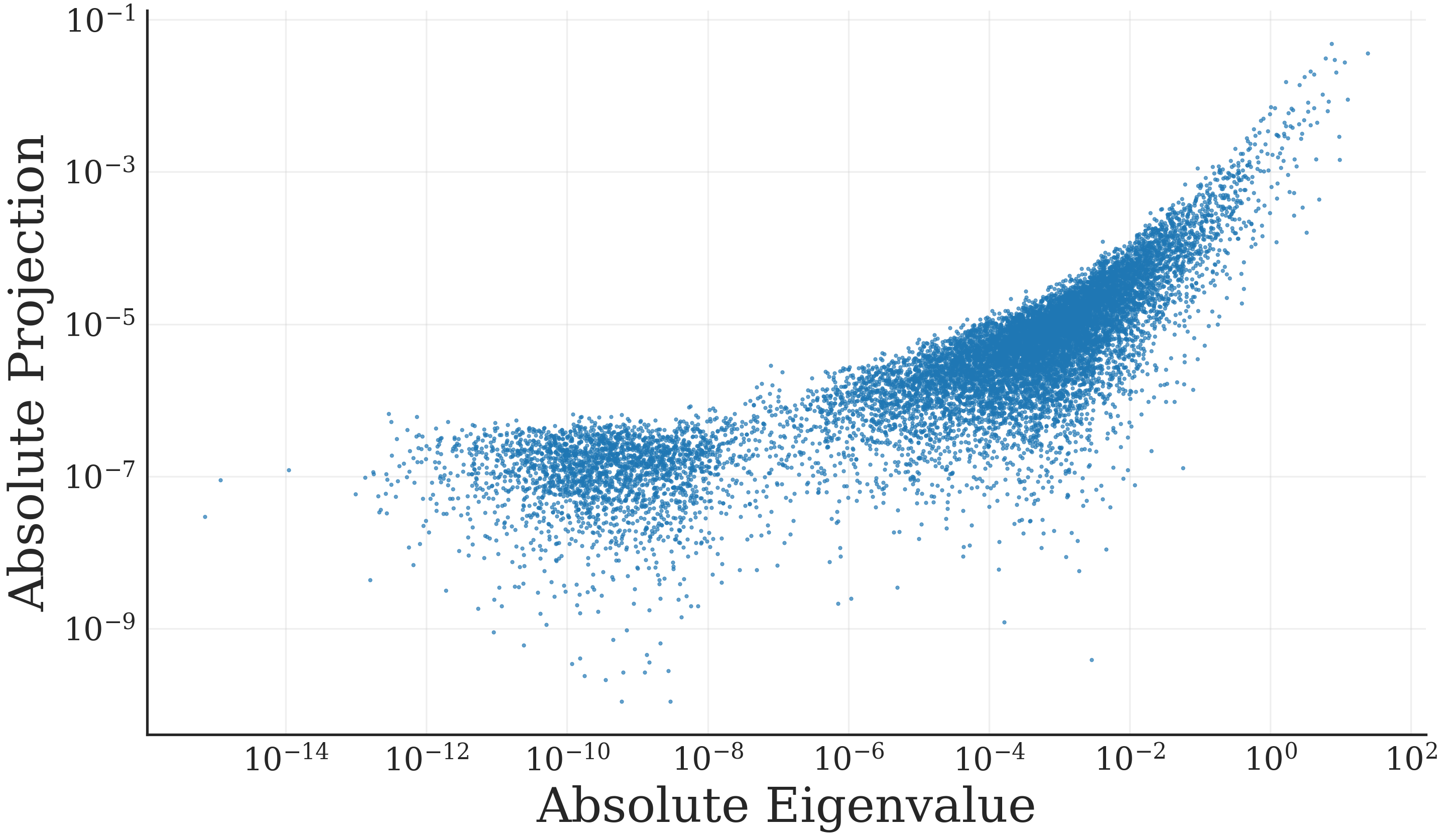}
        \caption*{(b) Projection of $\Delta w$ after the first training epoch.}
    \end{minipage}

    \vspace{0.4em}

    \begin{minipage}[b]{0.49\textwidth}
        \centering
        \includegraphics[width=\linewidth]{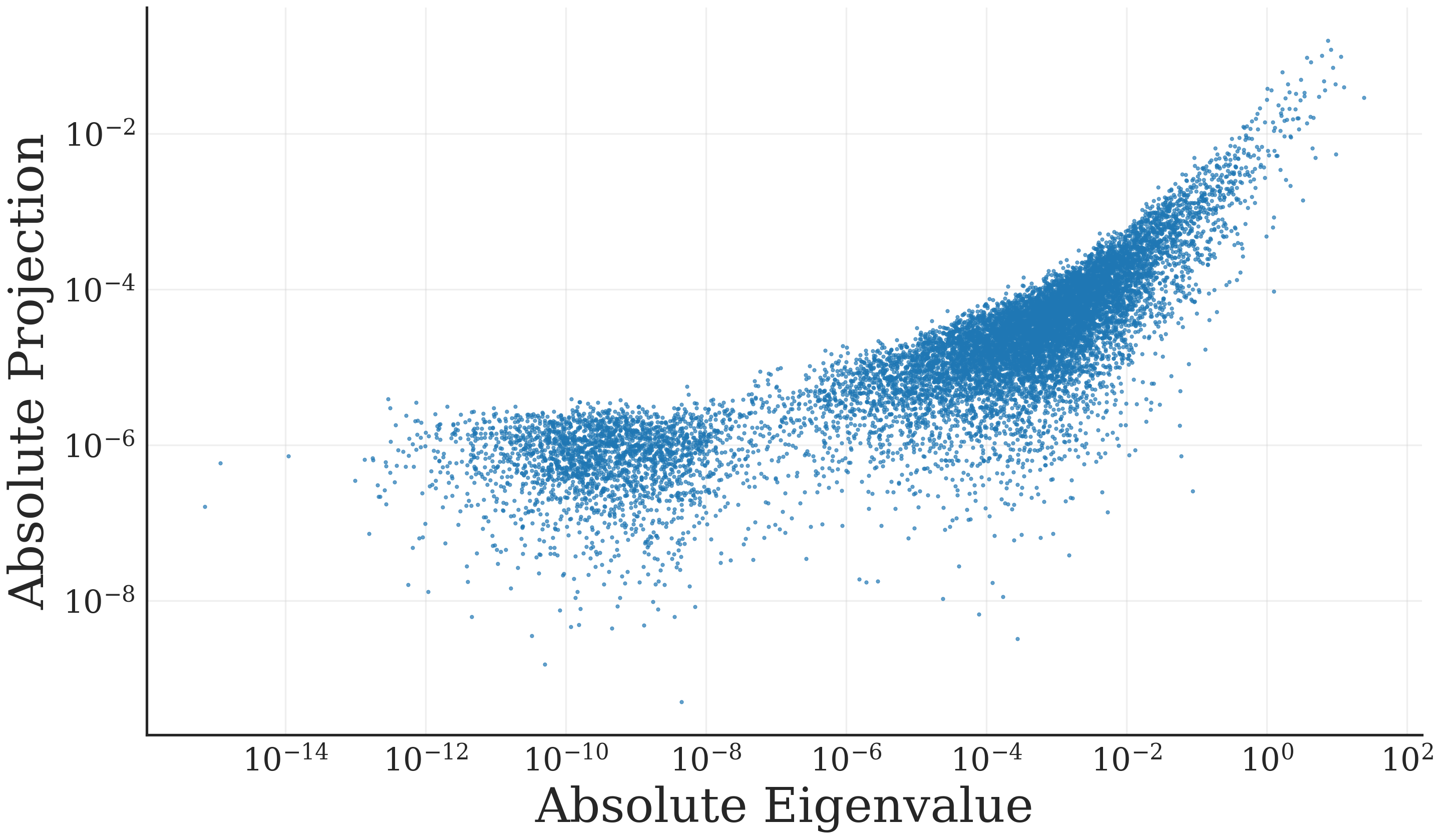}
        \caption*{(c) Projection of $\Delta w$ after the fifth training epoch.}
    \end{minipage}
    \hfill
    \begin{minipage}[b]{0.49\textwidth}
        \centering
        \includegraphics[width=\linewidth]{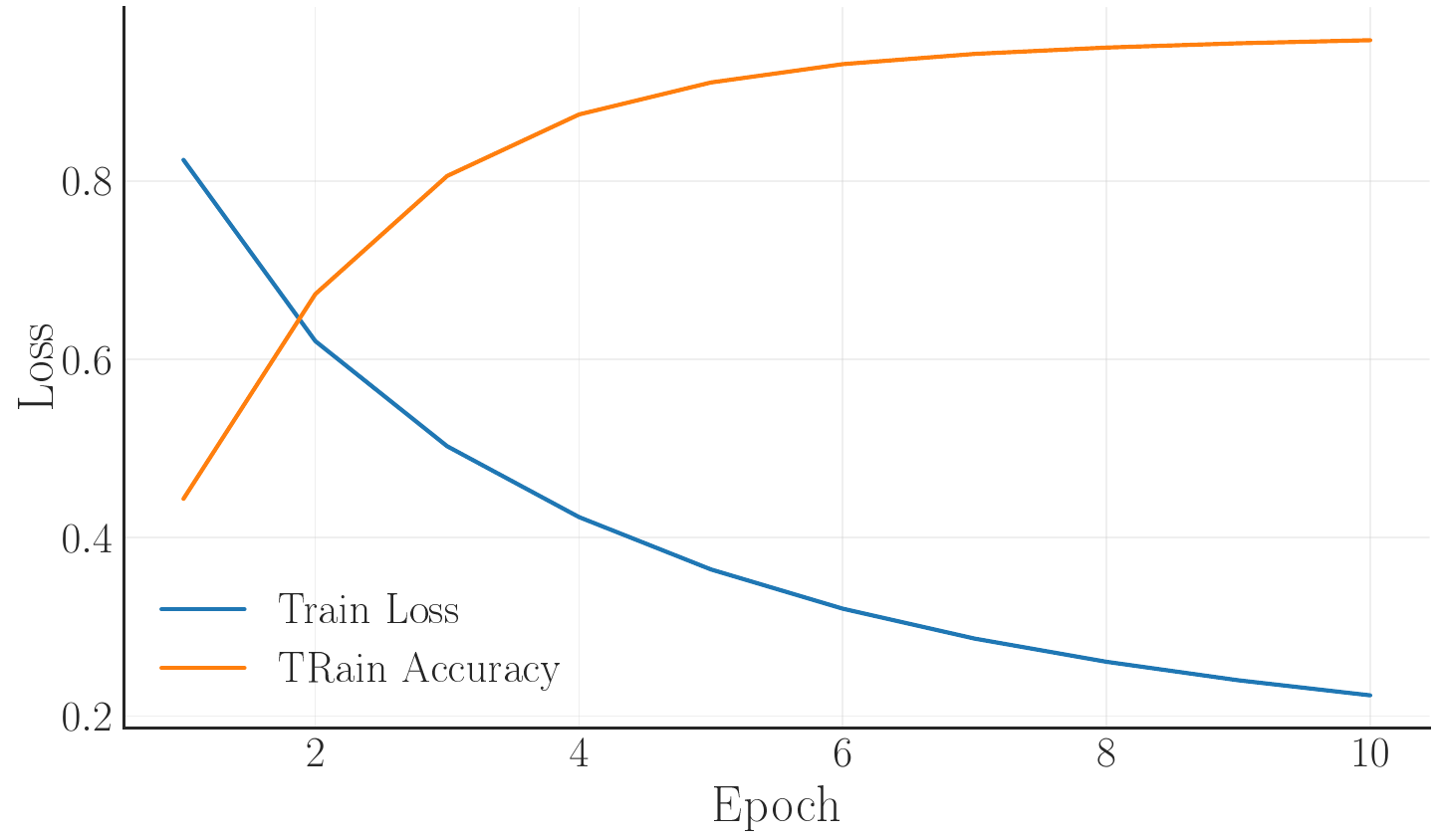}
        \caption*{(d) Loss and Accuracy across training epochs for a single random seed.}
    \end{minipage}

    \caption{Hessian spectrum and parameter displacement in the eigenbasis at different points in training. Loss and Accuracy progress throughout training.}
    \label{fig:hessian_quadrants}
\end{figure}

Figures \ref{fig:hessian_quadrants} (a), (b), and (c) show the projection after the first optimization step, after epoch $1$, and in epoch $5$. In all three cases, most of the displacement mass is concentrated in the leading directions, while the remaining coordinates remain close to zero. The overall shape of this profile becomes visible after the first update and remains qualitatively stable throughout training, indicating that gradient-based optimization quickly becomes confined to a low-dimensional subspace spanned by the top eigendirections of the Hessian at (or very near) initialization.

\clearpage

%% file: sections/appendix/03_training_details.tex
\newpage
\section{Training and Testing Details}
\label{appendix:training_parameters}
Following the setting is \cite{su2020sanity}, testing is performed after each training epoch, and we report the best top-1 test accuracy. Moreover, we ran each experiment on three seeds and reported the average test accuracy with the corresponding standard deviations. Table \ref{tab:table_training_parameters} shows the training parameters.
\begin{table}[h]
\caption{Training configurations for various networks and datasets.}
\label{tab:table_training_parameters}
\vskip 0.15in
\begin{center}
\resizebox{\textwidth}{!}{%
\begin{tabular}{lcccccccccc}
\toprule
Network & Dataset & Epochs & Batch & Opt. & Mom. & LR & Weight Decay & LR Drops & Drop Factor  \\
\midrule
ResNet-18   & CIFAR-10/100  & 160 & 512  & SGD & 0.9 & 0.03 & 5e-4  & 80, 120       & 0.2 \\
ResNet-50   & CIFAR-10      & 160 & 512  & SGD & 0.9 & 0.03 & 5e-4  & 80, 120       & 0.2 \\
VGG-19      & CIFAR-10/100  & 160 & 512  & SGD & 0.9 & 0.6  & 1e-4  & 80, 120       & 0.1 \\
ResNet-18   & TinyImageNet  & 100 & 512  & SGD & 0.9 & 0.03 & 1e-4  & 30, 60, 80    & 0.1 \\
ResNet-50   & ImageNet      & 90  & 256  & SGD & 0.9 & 0.1  & 1e-4  & 30, 60, 80    & 0.1 \\
\bottomrule
\end{tabular}
}
\end{center}
\vskip -0.1in
\end{table}

All CIFAR-10, CIFAR-100, and TinyImageNet experiments were conducted in single GPUs NVIDIA RTX 5000. The ImageNet experiments were conducted in a distributed setting, utilizing 4 NVIDIA A100-SXM4 GPUs. Each GPU was assigned a batch size of 256 and a learning rate of 0.1. This corresponds to a global batch size of 1024 and a global learning rate of 0.4.

%% file: sections/appendix/02_complexity.tex
\clearpage
\section{Computational Complexity}
\label{appendix:computational_complexity}
For clarity, we will stick to the notation in the main body of the manuscript: $d$ denotes the number of parameters in the network, $B$ is the number of batches, and $N$ is the number of data samples. Following \cite{frankle2020pruning}, we use a subset with 10 samples per class and a batch size of 1 to calculate the importance scores, so $ B = N$. Given the small subset, the wall-clock time added to the training pipeline is negligible. Some precision about the columns:
\begin{itemize}
    \item \textbf{Practical Time Complexity} corresponds to real computation, determined by the batch size used to create the mask and limited by available RAM/VRAM constraints.
    \item \textbf{Theoretical Time Complexity} indicates the worst-case scenario.
    \item \textbf{Space Complexity} indicates the storage requirements for estimating the metrics, considering that some require keeping intermediate results that will iteratively accumulate until the best metric estimation is obtained.
\end{itemize}

\begin{table}[h!]
    \centering
    \caption{Comparison of pruning methods in terms of computational complexity and runtime characteristics.}
    \label{tab:pruning_complexity}
    \begin{sc}
        \resizebox{\textwidth}{!}{%
            \begin{tabular}{ccccccc}
                \toprule
                Method & \makecell{Practical \\ Time \\ Complexity} & \makecell{Theoretical \\ Time \\ Complexity} & \makecell{Space \\ Complexity} & \makecell{Depends on \\ Data?} & \makecell{Uses \\ Gradients?} & \makecell{Hessian \\ Use} \\
                \midrule
                Random          & $\mathcal{O}(1)$     & $\mathcal{O}(1)$     & $\mathcal{O}(1)$ & No       & No        & None             \\
                Magnitude       & $\mathcal{O}(d)$     & $\mathcal{O}(d)$     & $\mathcal{O}(d)$ & No       & No        & None             \\
                GN              & $\mathcal{O}(Bd)$    & $\mathcal{O}(Nd)$    & $\mathcal{O}(d)$ & Yes      & Yes       & None             \\
                SNIP            & $\mathcal{O}(Bd)$    & $\mathcal{O}(Nd)$    & $\mathcal{O}(d)$ & Yes      & Yes       & None             \\
                Synflow         & $\mathcal{O}(d)$     & $\mathcal{O}(d)$     & $\mathcal{O}(d)$ & No       & Yes       & None             \\
                GraSP           & $\mathcal{O}(Bd^2)$  & $\mathcal{O}(Nd^2)$  & $\mathcal{O}(d)$ & Yes      & Yes       & Hessian-vector   \\
                \textbf{FD}     & $\mathcal{O}(Bd)$    & $\mathcal{O}(Nd)$    & $\mathcal{O}(d)$ & Yes      & Yes       & Diag. approx.$^{F^*}$    \\
                \textbf{FP}     & $\mathcal{O}(Bd)$    & $\mathcal{O}(Nd)$    & $\mathcal{O}(d)$ & Yes      & Yes       & Diag. approx.$^{F^*}$    \\
                \textbf{FTS}    & $\mathcal{O}(Bd)$    & $\mathcal{O}(Nd)$    & $\mathcal{O}(d)$ & Yes      & Yes       & Diag. approx.$^{F^*}$    \\
                \textbf{HD}     & $\mathcal{O}(Bd)$    & $\mathcal{O}(Nd)$    & $\mathcal{O}(d)$ & Yes      & Yes       & Diag. approx.$^{H^*}$     \\
                \textbf{HP}     & $\mathcal{O}(Bd)$    & $\mathcal{O}(Nd)$    & $\mathcal{O}(d)$ & Yes      & Yes       & Diag. approx.$^{H^*}$     \\
                \textbf{HTS}    & $\mathcal{O}(Bd)$    & $\mathcal{O}(Nd)$    & $\mathcal{O}(d)$ & Yes      & Yes       & Diag. approx.$^{H^*}$     \\
                OBD             & $\mathcal{O}(Bd)$    & $\mathcal{O}(Nd)$    & $\mathcal{O}(d)$   & Yes$^a$  & Yes$^b$   & Diag.$^a$        \\
                OBS             & $\mathcal{O}(Bd^2)$  & $\mathcal{O}(Nd^2)$  & $\mathcal{O}(d^2)$ & Yes$^a$  & Yes$^c$   & Full inverse$^d$ \\
                Full Hessian    & $\mathcal{O}(Bd^2)$  & $\mathcal{O}(Nd^2)$  & $\mathcal{O}(d^2)$ & Yes      & Yes       & Full             \\
                \bottomrule
            \end{tabular}
        }
    \end{sc}
\end{table}
{
    \tiny{
    $^{F^*}$  FIM Approximation, $^{H^*}$ Hutchinson Approximation, $^a$ Post-Training, $^b$ For diagonal, $^c$ Outer products, $^d$ Kailath.
    }
}

\textbf{Notes:}

\begin{enumerate}
    \item Synflow does not depend on data, but calculates gradients based on a dummy input. Runtime linearly increases with the number of pruning steps to reach the target sparsity (fixed at 100 as in the original implementation).
    \item F* methods use the empirical Fisher diagonal (computed from squared gradients on data batches), not the true Hessian. Practical runtime scales with the number of batches used for estimation.
    \item H* methods use Hutchinson’s stochastic diagonal approximation of the Hessian. The runtime scales linearly with the number of probe vectors (fixed at 10 here).
    \item Depends on Data? indicates whether the method requires explicit forward/backward passes on input data to estimate its metric, not just parameter values.
    \item Space Complexity: Entries such as $\mathcal{O}(d^2)$ (OBS, Full Hessian) are correct theoretically but infeasible in practice for modern networks, since storing the full Hessian or its inverse is memory-prohibitive.
\end{enumerate}

%% file: sections/appendix/16_amount_data_estimate_scores.tex
\clearpage
\section{Sensitivity Analysis To Samples and Hutchinson Probes}
\label{appendix:data_estimate_score}

As mentioned in Section \ref{appendix:computational_complexity} of the Appendix, we used a small subset of the data (10 samples per class) to compute the pruning scores following the protocol defined in \cite{frankle2020pruning}, which in turn follows the work in \cite{wang2020picking}, which argues it is a sufficiently large subset. To verify this assumption, we conducted experiments on the base reference case (CIFAR-10 on ResNet18). First, we defined a single sparsity (95\%), and then we evaluated data-dependent methods with an increasing number of samples per class, up to utilizing the whole training set (5k samples per class). We maintain a batch size of 1 for the score computation and use the same initialization seeds to ensure consistency with the experiments in the main body. We also incorporate the proposed warmup step, as it has been shown to improve method performance.
\nl
\begin{table}[h]
\caption{Effect of the number of samples per class used to compute the pruning mask (CIFAR-10, ResNet-18, sparsity 0.95, batch size 1).}
\label{tab:class_samples_cifar10_resnet18_s095}
\vskip 0.15in
\begin{center}
\begin{small}
\begin{sc}
\resizebox{\textwidth}{!}{%
\begin{tabular}{cccccccccc}
\toprule
Samples per class & GN & SNIP & GraSP & \textbf{FD} & \textbf{FP} & \textbf{FTS} & \textbf{HD} & \textbf{HP} & \textbf{HTS}\\
\midrule
1 & 92.56 ± 0.16 & 92.38 ± 0.04 & 91.18 ± 0.42 & 92.04 ± 0.13 & 92.07 ± 0.16 & 92.37 ± 0.17 & 92.47 ± 0.08 & 92.36 ± 0.11 & 92.36 ± 0.07 \\
10 & 92.15 ± 0.33 & 92.19 ± 0.13 & 90.86 ± 0.30 & 79.83 ± 19.10 & 91.36 ± 0.34 & 92.27 ± 0.06 & 92.30 ± 0.23 & 92.32 ± 0.21 & 92.18 ± 0.37 \\
100 & 91.28 ± 0.52 & 92.06 ± 0.15 & 90.48 ± 0.12 & 64.46 ± 2.90 & 91.43 ± 0.14 & 91.91 ± 0.30 & 92.36 ± 0.28 & 92.27 ± 0.24 & 92.28 ± 0.27 \\
1000 & 90.03 ± 0.27 & 91.87 ± 0.18 & 90.80 ± 0.27 & 34.51 ± 42.46 & 91.16 ± 0.06 & 91.71 ± 0.17 & 92.42 ± 0.15 & 92.28 ± 0.24 & 92.15 ± 0.20 \\
5000 & 89.87 ± 0.38 & 91.69 ± 0.23 & 90.76 ± 0.17 & 89.98 ± 0.28 & 91.75 ± 0.14 & 91.65 ± 0.20 & 92.45 ± 0.27 & 92.28 ± 0.10 & 92.13 ± 0.08 \\
\bottomrule
\end{tabular}}
\end{sc}
\end{small}
\end{center}
\vskip -0.1in
\end{table}

The results in Table \ref{tab:class_samples_cifar10_resnet18_s095} show that all criteria based on non-Hutchinson follow the same trend: as the number of samples for computation increases, there is a slight decrease in performance. FD is observed to suffer the most as the number of samples increases. However, Hutchinson-based criteria are more robust, further solidifying the value of second-order information. Given the defined trend, we question whether the batch size of 1 played a role in the lower performance when using the entire training set. In the same setup, we evaluated increasing the batch size for score computation (1, 10, 100, 1000). The selected batch sizes were defined to avoid mini-batches with fewer samples that misalign the average computation.
\nl
\begin{table}[h]
\caption{Effect of batch size used to compute the pruning mask (CIFAR-10, ResNet-18, sparsity 0.95, 5K samples/class).}
\label{tab:batch_sizes_cifar10_resnet18_s095}
\vskip 0.15in
\begin{center}
\begin{small}
\begin{sc}
\resizebox{\textwidth}{!}{%
\begin{tabular}{cccccccccc}
\toprule
Batch size & GN & SNIP & GraSP & \textbf{FD} & \textbf{FP} & \textbf{FTS} & \textbf{HD} & \textbf{HP} & \textbf{HTS}\\
\midrule
1 & 90.04 ± 0.30 & 91.63 ± 0.09 & 90.40 ± 0.35 & 34.25 ± 42.00 & 91.48 ± 0.08 & 91.86 ± 0.37 & 92.10 ± 0.12 & 92.34 ± 0.19 & 92.24 ± 0.15 \\
10 & 89.97 ± 0.22 & 91.56 ± 0.24 & 90.69 ± 0.07 & 86.40 ± 3.60 & 91.52 ± 0.38 & 91.74 ± 0.24 & 92.44 ± 0.14 & 92.33 ± 0.12 & 92.01 ± 0.18 \\
100 & 90.06 ± 0.12 & 91.80 ± 0.24 & 90.80 ± 0.26 & 90.82 ± 0.15 & 91.80 ± 0.11 & 91.91 ± 0.23 & 92.25 ± 0.15 & 92.25 ± 0.10 & 92.01 ± 0.17 \\
1000 & 89.93 ± 0.41 & 91.60 ± 0.24 & 90.87 ± 0.19 & 90.57 ± 0.24 & 91.72 ± 0.13 & 91.73 ± 0.07 & 92.26 ± 0.15 & 92.39 ± 0.12 & 92.19 ± 0.16 \\
\bottomrule
\end{tabular}}
\end{sc}
\end{small}
\end{center}
\vskip -0.1in
\end{table}

The results in Table \ref{tab:batch_sizes_cifar10_resnet18_s095} show that batch size does not affect the previously observed trend, as all methods (except FD) produce close results regardless of the batch size. Still, we observe that increasing the batch size alleviates the performance drop of FD. Considering the heuristics involved, it is challenging to draw a definitive conclusion. However, we can offer a general recommendation to increase the batch size of score computation in the PaI setting if the practitioner chooses to utilize a larger subset, as this can prevent the performance drop observed in FD and accelerate computation time.

\newpage
Regarding the number of probes (random Rademacher vectors), we used a fixed number of 10 probes to compute the Hutchinson Diagonal (HD) approximation across all experiments. We defined this constant based on the linear increase in computation as the number of probes increased. We performed experiments in the same setting as the base case (10 samples per class), with the only change being the number of random probes used to compute the score.
\nl
\begin{table}[h]
\caption{Effect of the number of random probes on Hutchinson diagonal mask computation (CIFAR-10, ResNet-18, sparsity 0.95).}
\label{tab:rand_probes_cifar10_resnet18_s095}
\vskip 0.15in
\begin{center}
\begin{small}
\begin{sc}
\begin{tabular}{ccccc}
\toprule
No. Probes & Avg. Acc. (w/o warmup) & Avg. Acc. (w/ warmup) & Computation Time (s)\\
\midrule
1 & 91.26 ± 0.15 & 92.37 ± 0.22 & 1.75 \\
10 & 91.01 ± 0.33 & 92.36 ± 0.09 & 9.26 \\
50 & 91.13 ± 0.02 & 92.44 ± 0.19 & 44.07 \\
100 & 91.13 ± 0.15 & 92.18 ± 0.12 & 87.65 \\
\bottomrule
\end{tabular}
\end{sc}
\end{small}
\end{center}
\vskip -0.1in
\end{table}

The results in Table \ref{tab:rand_probes_cifar10_resnet18_s095} show that the approximation is very robust to the number of probes used; even a single probe is sufficient to find a performant subnetwork. The warm-up step improved performance, but there was no improvement when the number of probes was increased. This aligns with the AdaHessian implementation \cite{yao2021adahessian}, which uses only one probe to compute the approximation used in the optimization step.

%% file: sections/appendix/04_bn_study.tex
\newpage

\section{Study on Batch-Norm Statistics}
\label{appendix:bn_study}

All models in our experiments use batch-norm (BN) layers; however, only VGG-19 experiences layer collapse. We presumed that better statistics yield better gradient estimations, which is paramount when using data-based criteria in pruning. To validate this insight, we ran an experiment comparing the resulting pruning masks before and after updating the BN statistics.

\begin{table}[h]
    \caption{Importance of batch normalization layers' statistics (BNS) in preventing layer collapse. 
    We studied the case of VGG-19 pruning at 95\% sparsity, which is the level that started showing 
    layer collapse. \textit{C} indicates the percentage of layers that collapsed after pruning 
    (100\% of the layer is pruned), and \textit{B} indicates the percentage leading to a bottleneck 
    (80\% of the layer is pruned).}
    \label{tab:bn_layer_collapse}
    \begin{sc}
        \resizebox{\textwidth}{!}{%
            \begin{tabular}{ccccccccccccccccccccccc}
                \toprule
                & \multicolumn{2}{c}{GN} & \multicolumn{2}{c}{SNIP} & \multicolumn{2}{c}{GraSP} 
                & \multicolumn{2}{c}{\textbf{FD}} & \multicolumn{2}{c}{\textbf{FP}} & \multicolumn{2}{c}{\textbf{FTS}} 
                & \multicolumn{2}{c}{\textbf{HD}} & \multicolumn{2}{c}{\textbf{HP}} & \multicolumn{2}{c}{\textbf{HTS}}  \\
                \midrule
                BNS Status & C (\%) & B (\%) & C (\%) & B (\%) & C (\%) & B (\%) 
                           & C (\%) & B (\%) & C (\%) & B (\%) & C (\%) & B (\%) 
                           & C (\%) & B (\%) & C (\%) & B (\%) & C (\%) & B (\%) \\
                \midrule
                Without Warm-Up   
                    & 16.67 & 24.20  
                    & 18.18 & 22.73  
                    & 16.67 & 24.24  
                    & 16.67 & 24.24  
                    & 12.12 & 24.24  
                    & 16.67 & 24.24  
                    & 13.64 & 24.24  
                    & 1.52  & 22.73  
                    & 13.64 & 24.24  
                    \\
                With Warm-Up      
                    & 0     & 22.73  
                    & 0     & 12.12  
                    & 0     & 12.12  
                    & 0     & 19.70  
                    & 0     & 22.73  
                    & 0     & 19.70  
                    & 0     & 15.15  
                    & 0     & 15.15  
                    & 0     & 16.67  
                    \\
                \bottomrule
            \end{tabular}
        }
    \end{sc}
\end{table}

The Table \ref{tab:bn_layer_collapse} compares the percentage of collapsed layers in VGG19 resulting from the pruning mask before and after updating only the BN statistics and leaving the initial parameters frozen using one complete pass through the data. Note that $\% C$ indicates the percentage of layers that collapsed after pruning (100\% of the layer is pruned), and $\%B$ indicates the percentage leading to a bottleneck ($80\%$ of the layer is pruned). This insight allowed us to propose the warmup process to mitigate layer collapse.

%% file: sections/appendix/05_cifar10_resnet_18.tex
\clearpage
\section{Results CIFAR10}
\subsection{ResNet18}
\label{appendix:CIFAR10_ResNet18}
\begin{table}[h]
\caption{Performance of different pruning methods for CIFAR-10 on ResNet18 without warmup. Bold and underlined values highlight the top performer. Bold values show methods whose confidence intervals overlap with the best performer. Baseline accuracy (no pruning): 93.89 ± 0.33.}
\label{tab:resnet18_cifar10_compressors_warmup_0}
\vskip 0.15in
\begin{center}
\begin{small}
\begin{sc}
\resizebox{\textwidth}{!}{%
\begin{tabular}{ccccccccccccc}
\toprule
Sparsity (\%) & Random & Magnitude & GN & SNIP & GraSP & SynFlow & \textbf{FD} & \textbf{FP} & \textbf{FTS} & \textbf{HD} & \textbf{HP} & \textbf{HTS}\\
\midrule
10 & \textbf{93.77 ± 0.15} & \textbf{93.81 ± 0.08} & \textbf{93.72 ± 0.15} & \textbf{93.68 ± 0.13} & 92.84 ± 0.12 & \underline{\textbf{93.89 ± 0.18}} & \textbf{93.83 ± 0.24} & \textbf{93.77 ± 0.11} & \textbf{93.87 ± 0.01} & \textbf{93.86 ± 0.18} & \textbf{93.85 ± 0.05} & \textbf{93.73 ± 0.15} \\
20 & \textbf{93.81 ± 0.17} & \textbf{93.70 ± 0.09} & \textbf{93.86 ± 0.25} & \textbf{93.74 ± 0.18} & 92.69 ± 0.26 & 93.63 ± 0.06 & \textbf{93.67 ± 0.19} & \textbf{93.82 ± 0.18} & \textbf{93.78 ± 0.08} & \textbf{93.69 ± 0.20} & \textbf{93.72 ± 0.18} & \underline{\textbf{93.87 ± 0.14}} \\
30 & \underline{\textbf{93.81 ± 0.05}} & \textbf{93.75 ± 0.15} & \textbf{93.77 ± 0.08} & \textbf{93.77 ± 0.17} & 92.69 ± 0.08 & 93.45 ± 0.11 & 93.58 ± 0.07 & \textbf{93.66 ± 0.22} & \textbf{93.80 ± 0.06} & \textbf{93.77 ± 0.10} & \textbf{93.58 ± 0.29} & \textbf{93.79 ± 0.18} \\
40 & \textbf{93.58 ± 0.07} & \textbf{93.63 ± 0.09} & \textbf{93.70 ± 0.13} & \textbf{93.58 ± 0.06} & 92.45 ± 0.05 & \textbf{93.71 ± 0.12} & \textbf{93.55 ± 0.14} & \underline{\textbf{93.93 ± 0.29}} & \textbf{93.74 ± 0.07} & \textbf{93.64 ± 0.12} & \textbf{93.55 ± 0.14} & \textbf{93.73 ± 0.12} \\
50 & 93.54 ± 0.22 & 93.58 ± 0.14 & \textbf{93.64 ± 0.16} & 93.55 ± 0.16 & 92.37 ± 0.24 & 93.62 ± 0.05 & \textbf{93.58 ± 0.21} & \underline{\textbf{93.89 ± 0.11}} & 93.65 ± 0.12 & 93.52 ± 0.04 & 93.52 ± 0.16 & 93.61 ± 0.16 \\
60 & 93.42 ± 0.09 & \textbf{93.67 ± 0.09} & \textbf{93.48 ± 0.09} & \textbf{93.56 ± 0.22} & 92.34 ± 0.25 & \underline{\textbf{93.79 ± 0.24}} & \textbf{93.51 ± 0.09} & \textbf{93.48 ± 0.24} & \textbf{93.56 ± 0.19} & 93.41 ± 0.14 & 93.45 ± 0.05 & \textbf{93.70 ± 0.17} \\
70 & 93.18 ± 0.16 & \textbf{93.59 ± 0.10} & \textbf{93.45 ± 0.07} & \underline{\textbf{93.60 ± 0.25}} & 92.02 ± 0.07 & 10.00 ± 0.00 & \textbf{93.39 ± 0.11} & \textbf{93.38 ± 0.13} & \textbf{93.50 ± 0.10} & \textbf{93.28 ± 0.14} & \textbf{93.54 ± 0.03} & \textbf{93.36 ± 0.21} \\
80 & 92.64 ± 0.23 & \underline{\textbf{93.35 ± 0.09}} & \textbf{93.22 ± 0.06} & \textbf{93.29 ± 0.04} & 92.06 ± 0.34 & 10.00 ± 0.00 & 92.86 ± 0.08 & \textbf{93.26 ± 0.11} & \textbf{93.30 ± 0.07} & 93.01 ± 0.17 & \textbf{93.00 ± 0.36} & \textbf{93.14 ± 0.16} \\
90 & 91.38 ± 0.16 & \textbf{92.86 ± 0.11} & 92.35 ± 0.22 & 92.73 ± 0.08 & 91.88 ± 0.37 & 10.00 ± 0.00 & 91.85 ± 0.07 & 92.49 ± 0.23 & \underline{\textbf{92.94 ± 0.08}} & 92.26 ± 0.12 & 92.70 ± 0.10 & \textbf{92.84 ± 0.19} \\
95 & 89.59 ± 0.41 & \textbf{92.05 ± 0.22} & 91.43 ± 0.21 & \textbf{91.70 ± 0.28} & 91.27 ± 0.22 & 10.00 ± 0.00 & 90.40 ± 0.20 & 91.66 ± 0.12 & \textbf{91.92 ± 0.15} & 91.20 ± 0.06 & \underline{\textbf{92.06 ± 0.22}} & \textbf{91.97 ± 0.20} \\
98 & 86.29 ± 0.22 & 88.45 ± 0.14 & 89.03 ± 0.14 & 89.86 ± 0.25 & 90.01 ± 0.18 & 10.00 ± 0.00 & 88.11 ± 0.37 & 89.54 ± 0.02 & 89.82 ± 0.08 & 88.98 ± 0.23 & \underline{\textbf{90.46 ± 0.14}} & 89.87 ± 0.37 \\
99 & 82.92 ± 0.23 & 76.54 ± 0.24 & 86.70 ± 0.53 & 87.61 ± 0.39 & \textbf{88.19 ± 0.41} & 10.00 ± 0.00 & 84.98 ± 0.12 & 86.52 ± 0.24 & 87.55 ± 0.21 & 87.39 ± 0.34 & \underline{\textbf{88.39 ± 0.06}} & 87.73 ± 0.19 \\
\bottomrule
\end{tabular}}
\end{sc}
\end{small}
\end{center}
\vskip -0.1in
\end{table}
Table~\ref{tab:resnet18_cifar10_compressors_warmup_0} shows the complete sparsity spectrum for ResNet-18 with CIFAR-10. We highlight the model's resilience given that random pruning has a negligible drop in performance up to $0.7$ sparsity compared to the baseline. After this point, we observe a significant degradation for naive methods (random and magnitude). The proposed second-order-based criteria (in bold) demonstrate their robustness by consistently ranking among the top performers across sparsity ratios.
\begin{table}[h]
\caption{Performance of different pruning methods for CIFAR-10 on ResNet-18 with warmup. Bold and underlined values highlight the top performer. Bold values indicate methods whose confidence intervals overlap with those of the best performer.}
\label{tab:resnet18_cifar10_compressors_warmup_1}
\vskip 0.15in
\begin{center}
\begin{small}
\begin{sc}
\resizebox{\textwidth}{!}{%
\begin{tabular}{ccccccccccc}
\toprule
Sparsity (\%) & GN & SNIP & GraSP & SynFlow & \textbf{FD} & \textbf{FP} & \textbf{FTS} & \textbf{HD} & \textbf{HP} & \textbf{HTS}\\
\midrule
80 & \textbf{93.32 ± 0.12} & \textbf{93.30 ± 0.24} & 91.73 ± 0.46 & 10.00 ± 0.00 & \textbf{93.52 ± 0.17} & \textbf{93.36 ± 0.08} & \textbf{93.47 ± 0.30} & \textbf{93.28 ± 0.08} & \textbf{93.32 ± 0.10} & \underline{\textbf{93.52 ± 0.03}} \\
90 & \textbf{92.71 ± 0.27} & \textbf{93.12 ± 0.02} & 91.59 ± 0.06 & 10.00 ± 0.00 & 92.89 ± 0.04 & \textbf{93.09 ± 0.26} & \textbf{93.00 ± 0.19} & 92.64 ± 0.19 & \textbf{92.87 ± 0.26} & \underline{\textbf{93.15 ± 0.21}} \\
95 & \textbf{92.16 ± 0.08} & \underline{\textbf{92.32 ± 0.16}} & 90.89 ± 0.29 & 10.00 ± 0.00 & \textbf{79.58 ± 18.25} & 91.52 ± 0.16 & \textbf{92.27 ± 0.10} & \textbf{92.20 ± 0.13} & \textbf{92.18 ± 0.08} & \textbf{92.23 ± 0.31} \\
98 & 90.28 ± 0.22 & 90.29 ± 0.25 & 89.78 ± 0.14 & 10.00 ± 0.00 & 10.00 ± 0.00 & 89.31 ± 0.14 & 90.27 ± 0.07 & \underline{\textbf{91.24 ± 0.27}} & 90.55 ± 0.18 & 90.25 ± 0.22 \\
99 & 60.07 ± 43.38 & 88.36 ± 0.44 & 87.98 ± 0.25 & 10.00 ± 0.00 & 10.00 ± 0.00 & 85.39 ± 0.34 & 88.02 ± 0.17 & \underline{\textbf{89.47 ± 0.15}} & 88.79 ± 0.20 & 88.06 ± 0.20 \\
\bottomrule
\end{tabular}}
\end{sc}
\end{small}
\end{center}
\vskip -0.1in
\end{table}

Table \ref{tab:resnet18_cifar10_compressors_warmup_1} shows how the proposed warmup step improves the gradient estimation and improves the performance of first- and second-order criteria. Notably, HD shows a significant improvement in performance.

\begin{figure*}[tb]
    \centering
    \begin{subfigure}{0.85\textwidth}
        \includegraphics[width=\linewidth]{imgs/ResNet_18_CIFAR_10_w=0_lineplot.pdf}
        \caption{Without warmup}
        \label{fig:ResNet_18_CIFAR_10_w=0_appendix}
    \end{subfigure}
    \hfill
    \begin{subfigure}{0.85\textwidth}
        \includegraphics[width=\linewidth]{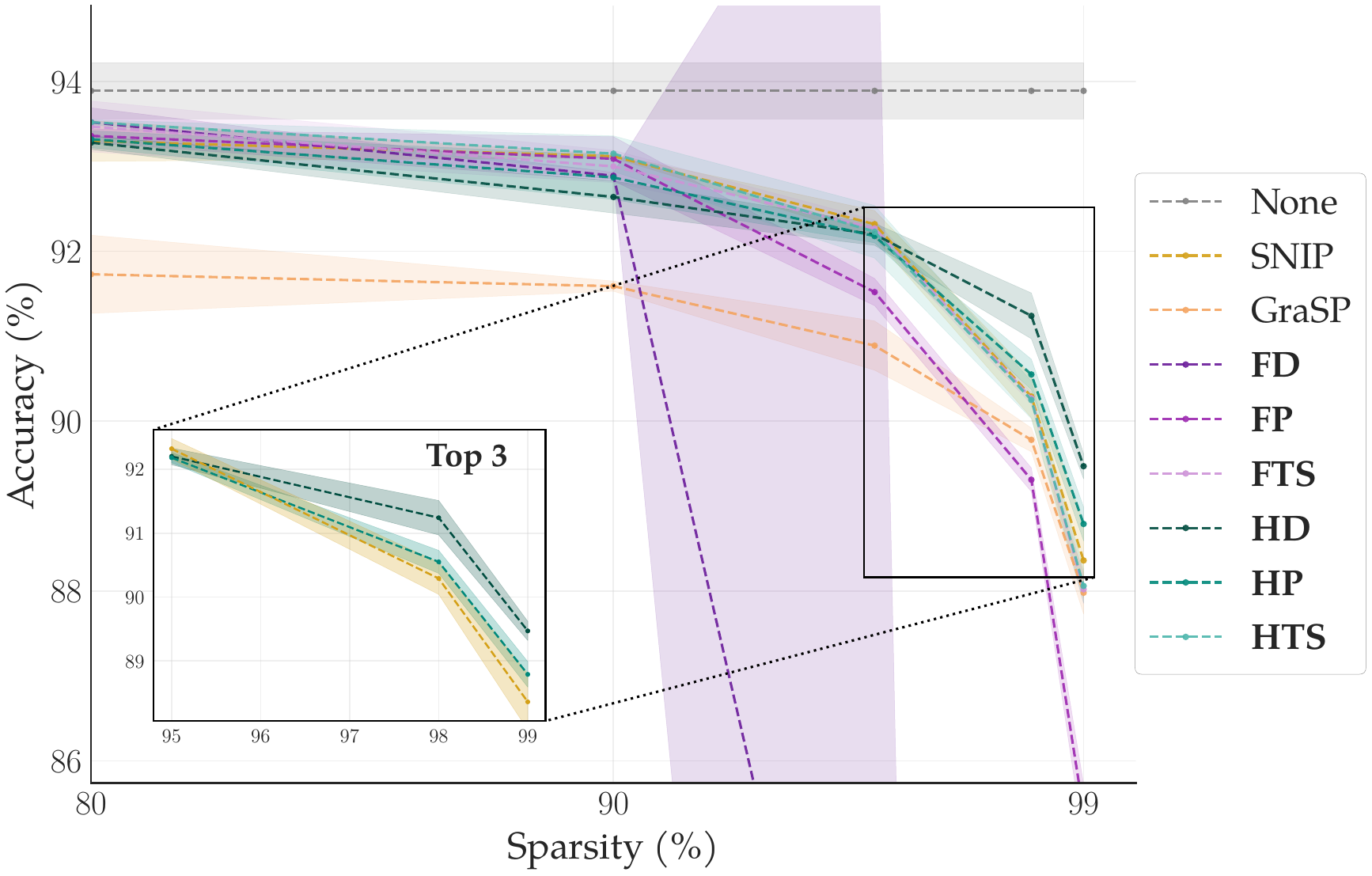}
        \caption{With warmup}
        \label{fig:ResNet_18_CIFAR_10_w=1_appendix}
    \end{subfigure}
    \caption{Test accuracy for ResNet-18 for the CIFAR-10 dataset with and without warmup. For extreme sparsities, we zoom in on the top 3 performers.}
    \label{fig:ResNet_18_CIFAR_10}
\end{figure*}

%% file: sections/appendix/06_cifar10_vgg_19.tex
\clearpage
\subsection{VGG19}
\label{appendix:CIFAR10_VGG19}
\begin{table}[h]
\caption{Performance of different pruning methods for CIFAR-10 on VGG-19 without warmup. Bold and underlined values highlight the top performer. Bold values show methods whose confidence intervals overlap with the best performer. Baseline accuracy (no pruning): 91.57 ± 0.19.}
\label{tab:vgg19_cifar10_compressors_warmup_0}
\vskip 0.15in
\begin{center}
\begin{small}
\begin{sc}
\resizebox{\textwidth}{!}{%
\begin{tabular}{ccccccccccccc}
\toprule
Sparsity (\%) & Random & Magnitude & GN & SNIP & GraSP & SynFlow & \textbf{FD} & \textbf{FP} & \textbf{FTS} & \textbf{HD} & \textbf{HP} & \textbf{HTS}\\
\midrule
10 & 91.46 ± 0.46 & \textbf{90.90 ± 1.50} & \textbf{91.91 ± 0.35} & \textbf{91.95 ± 0.26} & \textbf{64.12 ± 46.87} & \underline{\textbf{92.36 ± 0.36}} & \textbf{91.96 ± 0.16} & \textbf{91.87 ± 0.28} & \textbf{64.53 ± 47.23} & \textbf{91.31 ± 0.87} & \textbf{91.40 ± 0.61} & 91.76 ± 0.16 \\
20 & \textbf{91.58 ± 0.30} & 91.11 ± 0.54 & \textbf{90.27 ± 3.09} & \textbf{91.91 ± 0.20} & \textbf{64.41 ± 47.12} & 10.00 ± 0.00 & \textbf{92.08 ± 0.18} & \textbf{92.09 ± 0.15} & \textbf{90.60 ± 3.09} & \textbf{89.58 ± 3.07} & 91.14 ± 0.26 & \underline{\textbf{92.17 ± 0.33}} \\
30 & 91.42 ± 0.48 & \textbf{91.89 ± 0.28} & 37.36 ± 47.39 & 91.93 ± 0.07 & 10.00 ± 0.00 & 10.00 ± 0.00 & \textbf{91.82 ± 0.50} & \textbf{92.09 ± 0.17} & \underline{\textbf{92.13 ± 0.03}} & 91.74 ± 0.20 & 91.85 ± 0.18 & \textbf{90.19 ± 3.25} \\
40 & 91.68 ± 0.18 & \textbf{91.60 ± 0.42} & \underline{\textbf{92.19 ± 0.29}} & \textbf{92.07 ± 0.19} & \textbf{57.55 ± 42.32} & 10.00 ± 0.00 & \textbf{92.11 ± 0.20} & \textbf{92.09 ± 0.11} & \textbf{89.85 ± 4.13} & 91.58 ± 0.14 & \textbf{90.30 ± 2.37} & \textbf{92.16 ± 0.22} \\
50 & 90.94 ± 0.39 & \textbf{90.61 ± 1.80} & \textbf{92.03 ± 0.45} & \textbf{90.73 ± 2.14} & \textbf{64.21 ± 46.95} & 10.00 ± 0.00 & \textbf{92.02 ± 0.37} & \textbf{92.15 ± 0.32} & \textbf{92.00 ± 0.34} & \textbf{91.82 ± 0.33} & 91.67 ± 0.28 & \underline{\textbf{92.25 ± 0.25}} \\
60 & 90.95 ± 0.42 & \textbf{91.53 ± 0.72} & \textbf{92.16 ± 0.26} & \textbf{64.87 ± 47.52} & 37.14 ± 47.00 & 10.00 ± 0.00 & \textbf{92.09 ± 0.13} & \textbf{92.20 ± 0.11} & \underline{\textbf{92.21 ± 0.21}} & \textbf{64.56 ± 47.25} & 37.31 ± 47.30 & \textbf{92.09 ± 0.06} \\
70 & 90.72 ± 0.10 & \textbf{90.32 ± 2.33} & \underline{\textbf{92.25 ± 0.39}} & \textbf{92.04 ± 0.14} & 91.27 ± 0.58 & 10.00 ± 0.00 & 91.49 ± 0.27 & \textbf{91.55 ± 0.49} & \textbf{64.48 ± 47.18} & \textbf{64.51 ± 47.21} & \textbf{92.19 ± 0.27} & \textbf{91.76 ± 0.16} \\
80 & \textbf{63.26 ± 46.09} & \textbf{91.58 ± 0.14} & 90.23 ± 0.18 & 90.98 ± 0.19 & \textbf{64.39 ± 47.10} & 10.00 ± 0.00 & 10.00 ± 0.00 & 83.95 ± 0.31 & \textbf{91.24 ± 0.66} & \underline{\textbf{91.65 ± 0.30}} & \textbf{64.44 ± 47.15} & 90.50 ± 0.10 \\
90 & 88.12 ± 1.63 & \underline{\textbf{91.54 ± 0.10}} & 10.00 ± 0.00 & 10.00 ± 0.00 & \textbf{59.96 ± 43.27} & 10.00 ± 0.00 & 10.00 ± 0.00 & 10.00 ± 0.00 & 64.24 ± 18.53 & 10.00 ± 0.00 & 88.84 ± 0.39 & 10.00 ± 0.00 \\
95 & \underline{\textbf{86.45 ± 0.92}} & 10.00 ± 0.00 & 10.00 ± 0.00 & 10.00 ± 0.00 & 10.00 ± 0.00 & 10.00 ± 0.00 & 10.00 ± 0.00 & 10.00 ± 0.00 & 10.00 ± 0.00 & 10.00 ± 0.00 & 10.00 ± 0.00 & 10.00 ± 0.00 \\
98 & \underline{\textbf{82.58 ± 2.12}} & 10.00 ± 0.00 & 10.00 ± 0.00 & 10.00 ± 0.00 & 10.00 ± 0.00 & 10.00 ± 0.00 & 10.00 ± 0.00 & 10.00 ± 0.00 & 10.00 ± 0.00 & 10.00 ± 0.00 & 10.00 ± 0.00 & 10.00 ± 0.00 \\
99 & \underline{\textbf{78.18 ± 2.27}} & 10.00 ± 0.00 & 10.00 ± 0.00 & 10.00 ± 0.00 & 10.00 ± 0.00 & 10.00 ± 0.00 & 10.00 ± 0.00 & 10.00 ± 0.00 & 10.00 ± 0.00 & 10.00 ± 0.00 & 10.00 ± 0.00 & 10.00 ± 0.00 \\
\bottomrule
\end{tabular}}
\end{sc}
\end{small}
\end{center}
\vskip -0.1in
\end{table}
\begin{table}[h]
\caption{Performance of different pruning methods for CIFAR-10 on VGG-19 with warmup. Bold and underlined values highlight the top performer. Bold values show methods whose confidence intervals overlap with the best performer.}
\label{tab:vgg19_cifar10_compressors_warmup_1}
\vskip 0.15in
\begin{center}
\begin{small}
\begin{sc}
\resizebox{\textwidth}{!}{%
\begin{tabular}{ccccccccccc}
\toprule
Sparsity (\%) & GN & SNIP & GraSP & SynFlow & \textbf{FD} & \textbf{FP} & \textbf{FTS} & \textbf{HD} & \textbf{HP} & \textbf{HTS}\\
\midrule
80 & \textbf{90.75 ± 2.09} & \textbf{91.69 ± 0.83} & 90.74 ± 0.64 & 10.00 ± 0.00 & \underline{\textbf{92.37 ± 0.35}} & \textbf{91.09 ± 1.53} & 89.75 ± 2.14 & 91.47 ± 0.12 & 91.14 ± 0.36 & \textbf{91.85 ± 0.23} \\
90 & 92.04 ± 0.47 & 91.95 ± 0.16 & 90.98 ± 0.45 & 10.00 ± 0.00 & 46.87 ± 41.19 & \underline{\textbf{93.04 ± 0.14}} & \textbf{64.46 ± 47.17} & 91.18 ± 0.20 & 91.00 ± 0.38 & 91.62 ± 0.48 \\
95 & \underline{\textbf{92.25 ± 0.19}} & \textbf{91.89 ± 0.70} & 90.58 ± 0.19 & 10.00 ± 0.00 & 10.00 ± 0.00 & 10.00 ± 0.00 & \textbf{92.04 ± 0.35} & 91.24 ± 0.26 & 91.05 ± 0.35 & 91.52 ± 0.47 \\
98 & 59.50 ± 43.46 & 37.15 ± 47.03 & \underline{\textbf{90.63 ± 0.42}} & 10.00 ± 0.00 & 10.00 ± 0.00 & 10.00 ± 0.00 & 23.68 ± 23.69 & \textbf{90.17 ± 0.58} & \textbf{90.38 ± 0.26} & 32.14 ± 23.81 \\
99 & 10.00 ± 0.00 & 10.00 ± 0.00 & \underline{\textbf{89.10 ± 0.37}} & 10.00 ± 0.00 & 10.00 ± 0.00 & 10.00 ± 0.00 & 10.00 ± 0.00 & \textbf{89.06 ± 0.36} & 66.05 ± 7.84 & 10.00 ± 0.00 \\
\bottomrule
\end{tabular}}
\end{sc}
\end{small}
\end{center}
\vskip -0.1in
\end{table}
Table~\ref{tab:vgg19_cifar10_compressors_warmup_0} shows the complete sparsity spectrum for VGG-19 with CIFAR-10. We observe that performance drastically degrades for data-dependent methods as sparsity increases, ultimately leading to layer collapse. As discussed in the Results section of the main body, we propose a warm-up phase that updates the batch norm statistics to prevent collapse and stabilize pruning performance. Table~\ref{tab:vgg19_cifar10_compressors_warmup_1} demonstrates the effectiveness of this approach to mitigate layer collapse.

%% file: sections/appendix/07_cifar10_resnet_50.tex
\newpage
\subsection{ResNet50}
\label{appendix:CIFAR10_ResNet50}
\begin{table}[h]
\caption{Performance of different pruning methods for CIFAR-10 on ResNet-50 without warmup. Bold and underlined values highlight the top performer. Bold values indicate methods whose confidence intervals overlap with those of the best performer. Baseline accuracy (no pruning): 93.17 ± 0.25.}
\label{tab:resnet50_cifar10_compressors_warmup_0}
\vskip 0.15in
\begin{center}
\begin{small}
\begin{sc}
\resizebox{\textwidth}{!}{%
\begin{tabular}{ccccccccccccc}
\toprule
Sparsity (\%) & Random & Magnitude & GN & SNIP & GraSP & SynFlow & \textbf{FD} & \textbf{FP} & \textbf{FTS} & \textbf{HD} & \textbf{HP} & \textbf{HTS}\\
\midrule
10 & \textbf{93.52 ± 0.18} & \textbf{93.63 ± 0.22} & \textbf{93.45 ± 0.19} & \textbf{93.45 ± 0.34} & 91.91 ± 0.13 & \underline{\textbf{93.80 ± 0.47}} & \textbf{93.50 ± 0.30} & \textbf{93.23 ± 0.26} & 92.91 ± 0.15 & \textbf{92.78 ± 1.05} & 92.96 ± 0.26 & \textbf{93.35 ± 0.70} \\
20 & 92.91 ± 0.38 & 93.23 ± 0.35 & 93.27 ± 0.16 & 93.34 ± 0.18 & 90.70 ± 1.12 & \underline{\textbf{93.92 ± 0.13}} & 92.95 ± 0.07 & 92.96 ± 0.52 & 93.43 ± 0.04 & 92.91 ± 0.82 & 93.24 ± 0.33 & \textbf{93.40 ± 0.55} \\
30 & 93.32 ± 0.14 & 93.11 ± 0.11 & 92.82 ± 0.56 & 93.21 ± 0.47 & 90.72 ± 2.88 & \underline{\textbf{94.07 ± 0.06}} & 93.01 ± 0.35 & 93.19 ± 0.10 & 93.36 ± 0.31 & 92.90 ± 0.42 & 93.10 ± 0.32 & 93.24 ± 0.14 \\
40 & 93.20 ± 0.03 & 93.18 ± 0.36 & 93.07 ± 0.34 & 92.96 ± 0.51 & 87.50 ± 4.27 & \underline{\textbf{94.20 ± 0.06}} & 93.11 ± 0.76 & 92.91 ± 0.21 & 93.24 ± 0.47 & 93.15 ± 0.42 & 93.32 ± 0.15 & 93.42 ± 0.21 \\
50 & 93.05 ± 0.51 & 93.25 ± 0.38 & 93.22 ± 0.33 & 92.98 ± 0.62 & 89.63 ± 2.63 & \underline{\textbf{94.28 ± 0.01}} & 93.01 ± 0.48 & 93.18 ± 0.30 & 93.01 ± 0.31 & 92.98 ± 0.29 & 93.11 ± 0.37 & 93.14 ± 0.40 \\
60 & \textbf{93.07 ± 0.47} & \textbf{92.93 ± 0.84} & \textbf{93.18 ± 0.40} & \underline{\textbf{93.43 ± 0.61}} & 88.74 ± 2.41 & 10.00 ± 0.00 & \textbf{93.01 ± 0.22} & \textbf{93.01 ± 0.46} & \textbf{93.07 ± 0.15} & \textbf{93.18 ± 0.57} & \textbf{93.16 ± 0.38} & \textbf{93.02 ± 0.40} \\
70 & 93.17 ± 0.16 & \underline{\textbf{93.78 ± 0.15}} & 92.83 ± 0.33 & 92.96 ± 0.41 & 87.86 ± 2.54 & 10.00 ± 0.00 & 92.45 ± 0.54 & 91.82 ± 0.63 & 92.94 ± 0.13 & 93.30 ± 0.13 & 93.16 ± 0.11 & 92.96 ± 0.01 \\
80 & \textbf{92.71 ± 0.31} & 90.88 ± 0.21 & \textbf{93.00 ± 0.09} & 92.29 ± 0.10 & 89.63 ± 2.43 & 10.00 ± 0.00 & 92.32 ± 0.49 & 92.67 ± 0.06 & \textbf{92.79 ± 0.43} & 92.69 ± 0.10 & \underline{\textbf{93.13 ± 0.16}} & 92.29 ± 0.18 \\
90 & 91.80 ± 0.10 & 80.13 ± 0.22 & 92.51 ± 0.10 & 92.18 ± 0.23 & 88.76 ± 1.14 & 10.00 ± 0.00 & 91.75 ± 0.06 & 92.19 ± 0.17 & \underline{\textbf{92.73 ± 0.04}} & 92.18 ± 0.08 & \textbf{92.54 ± 0.35} & 92.32 ± 0.36 \\
95 & 89.94 ± 0.23 & 10.00 ± 0.00 & 90.32 ± 0.29 & \textbf{91.57 ± 0.31} & 87.69 ± 1.71 & 10.00 ± 0.00 & 88.18 ± 0.60 & 91.05 ± 0.21 & \underline{\textbf{91.80 ± 0.11}} & 90.80 ± 0.33 & \textbf{91.43 ± 0.39} & \textbf{91.38 ± 0.35} \\
98 & 86.96 ± 0.38 & 10.00 ± 0.00 & 81.79 ± 1.48 & \textbf{88.53 ± 0.30} & \textbf{88.38 ± 0.61} & 10.00 ± 0.00 & 69.92 ± 4.22 & 85.49 ± 0.18 & \textbf{88.31 ± 0.20} & 87.65 ± 0.57 & \underline{\textbf{88.90 ± 0.55}} & \textbf{88.42 ± 0.21} \\
99 & 66.14 ± 27.98 & 10.00 ± 0.00 & 66.28 ± 3.00 & 82.41 ± 0.99 & \underline{\textbf{85.96 ± 0.32}} & 10.00 ± 0.00 & 52.05 ± 2.02 & 74.83 ± 2.06 & 82.15 ± 0.88 & 79.84 ± 0.86 & 84.02 ± 0.78 & 83.70 ± 0.36 \\
\bottomrule
\end{tabular}}
\end{sc}
\end{small}
\end{center}
\vskip -0.1in
\end{table}
Table~\ref{tab:resnet50_cifar10_compressors_warmup_0} shows the complete sparsity spectrum for ResNet-50 with CIFAR-10. Compared to the smaller ResNet18 results, this deeper and wider architecture exhibits a faster decline in performance as the sparsity increases. Notably, the SynFlow criterion performs well for low to moderate sparsity, but suffers from layer collapse as sparsity increases. The proposed HP criteria (in bold) remain top performers for high sparsity levels, reinforcing the idea that there is no silver bullet for PaI, and that criteria are complementary. However, we emphasize that our proposed criteria effectively cover a broader part of the spectrum.

\begin{figure}[b]
    \centering
    \includegraphics[width=0.85\linewidth]{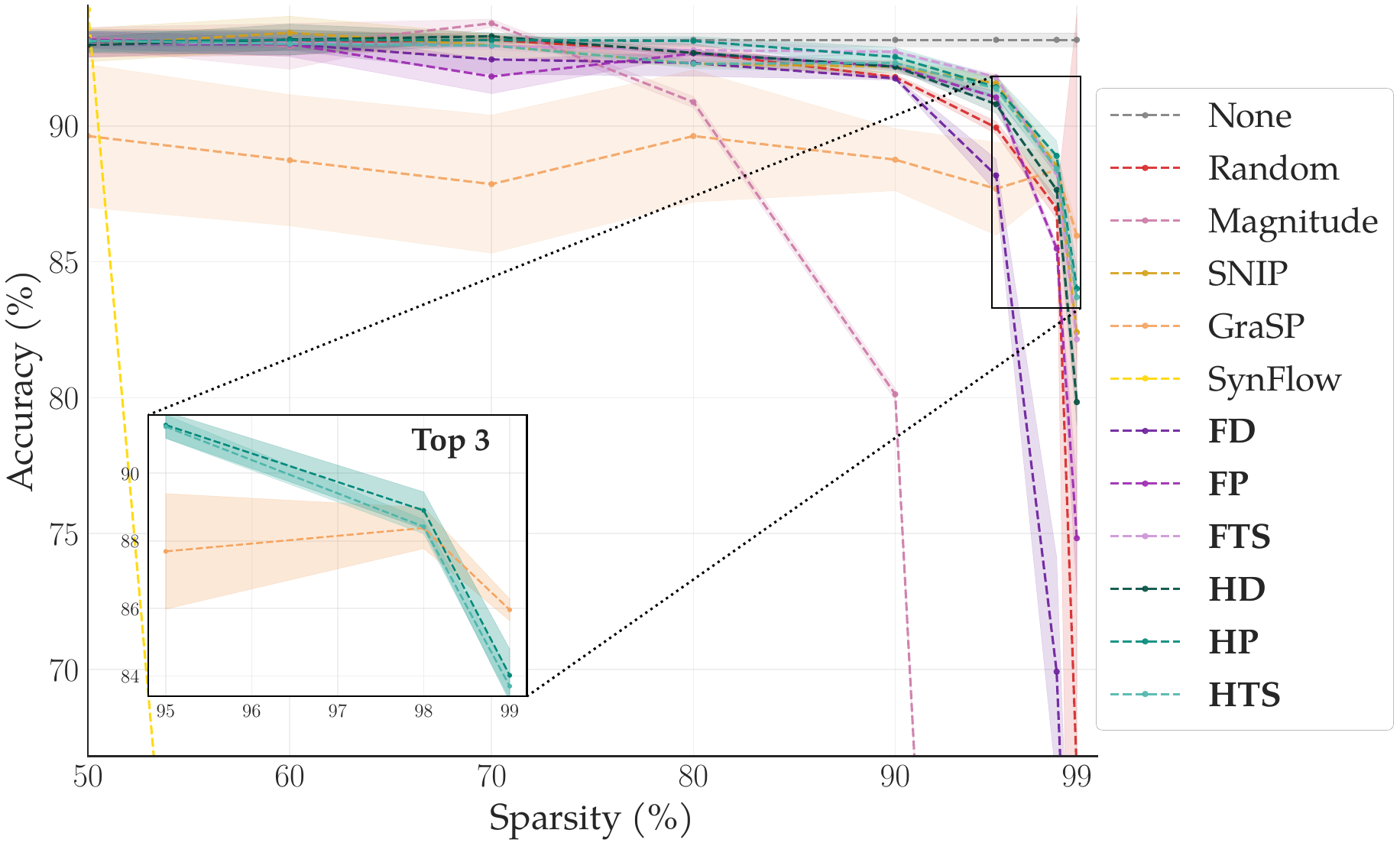}
    \caption{Test accuracy for ResNet50 for the CIFAR-10 dataset only without warmup. For extreme sparsities, we zoom in on the top 3 performers.}
    \label{fig:ResNet_50_CIFAR_10_w=0}
\end{figure}

%% file: sections/appendix/08_cifar100_resnet_18.tex
\newpage
\section{Results CIFAR100}
\subsection{ResNet18}
\label{sec:resnet_cifar-100}
\begin{table}[h]
\caption{Performance of different pruning methods for CIFAR-100 on ResNet-18 without warmup. Bold and underlined values highlight the top performer. Bold values show methods whose confidence intervals overlap with the best performer. Baseline accuracy (no pruning): 74.19 ± 0.14.}
\label{tab:resnet18_cifar100_compressors_warmup_0}
\vskip 0.15in
\begin{center}
\begin{small}
\begin{sc}
\resizebox{\textwidth}{!}{%
\begin{tabular}{ccccccccccccc}
\toprule
Sparsity (\%) & Random & Magnitude & GN & SNIP & GraSP & SynFlow & \textbf{FD} & \textbf{FP} & \textbf{FTS} & \textbf{HD} & \textbf{HP} & \textbf{HTS}\\
\midrule
10 & \textbf{74.19 ± 0.28} & \textbf{74.51 ± 0.14} & 74.25 ± 0.06 & \textbf{74.50 ± 0.18} & 71.35 ± 0.41 & \textbf{74.49 ± 0.53} & \textbf{74.37 ± 0.30} & \underline{\textbf{74.55 ± 0.15}} & \textbf{74.40 ± 0.13} & 74.06 ± 0.32 & 74.09 ± 0.08 & \textbf{74.35 ± 0.34} \\
20 & \textbf{74.19 ± 0.44} & 74.06 ± 0.03 & \textbf{74.40 ± 0.34} & 73.93 ± 0.18 & 71.34 ± 0.16 & 73.80 ± 0.16 & 74.06 ± 0.14 & \textbf{74.31 ± 0.39} & \underline{\textbf{74.43 ± 0.10}} & 74.00 ± 0.08 & 73.92 ± 0.26 & \textbf{74.39 ± 0.05} \\
30 & \textbf{74.11 ± 0.09} & \textbf{74.18 ± 0.39} & \textbf{74.14 ± 0.03} & \textbf{73.89 ± 0.33} & 70.82 ± 0.32 & \textbf{73.83 ± 0.52} & \textbf{74.11 ± 0.15} & \textbf{74.21 ± 0.17} & \underline{\textbf{74.28 ± 0.21}} & \textbf{73.96 ± 0.16} & \textbf{73.96 ± 0.25} & \textbf{74.16 ± 0.57} \\
40 & 73.69 ± 0.27 & 73.85 ± 0.12 & 73.90 ± 0.13 & 73.90 ± 0.14 & 70.09 ± 0.16 & \textbf{73.78 ± 0.40} & 73.87 ± 0.19 & \underline{\textbf{74.37 ± 0.26}} & \textbf{73.97 ± 0.36} & 73.71 ± 0.20 & 73.75 ± 0.10 & \textbf{74.18 ± 0.27} \\
50 & \textbf{73.55 ± 0.18} & \textbf{73.82 ± 0.49} & \textbf{73.59 ± 0.25} & \textbf{73.86 ± 0.51} & 70.33 ± 0.52 & \textbf{73.55 ± 0.26} & 73.25 ± 0.04 & \textbf{73.59 ± 0.30} & \underline{\textbf{74.00 ± 0.32}} & \textbf{73.62 ± 0.25} & \textbf{73.45 ± 0.30} & \textbf{73.99 ± 0.25} \\
60 & \textbf{73.28 ± 0.27} & 73.09 ± 0.26 & \textbf{73.51 ± 0.29} & \textbf{73.16 ± 0.41} & 69.80 ± 0.63 & \textbf{73.38 ± 0.14} & 72.70 ± 0.35 & \underline{\textbf{73.54 ± 0.15}} & \textbf{73.25 ± 0.17} & 73.24 ± 0.11 & \textbf{73.21 ± 0.22} & \textbf{73.29 ± 0.15} \\
70 & 72.49 ± 0.12 & \underline{\textbf{73.30 ± 0.22}} & 72.61 ± 0.19 & 72.91 ± 0.14 & 69.70 ± 0.27 & 1.00 ± 0.00 & 71.84 ± 0.33 & 72.48 ± 0.05 & 72.82 ± 0.06 & 72.87 ± 0.05 & 72.78 ± 0.28 & \textbf{72.83 ± 0.41} \\
80 & 71.07 ± 0.30 & \underline{\textbf{72.49 ± 0.31}} & 71.34 ± 0.19 & \textbf{72.16 ± 0.20} & 69.30 ± 0.39 & 1.00 ± 0.00 & 70.85 ± 0.37 & 71.53 ± 0.09 & 71.53 ± 0.19 & 71.42 ± 0.13 & 71.80 ± 0.18 & 71.98 ± 0.08 \\
90 & 68.46 ± 0.21 & \textbf{70.17 ± 0.15} & 68.81 ± 0.35 & 69.88 ± 0.05 & 67.72 ± 0.27 & 1.00 ± 0.00 & 68.64 ± 0.24 & 69.70 ± 0.15 & \textbf{70.07 ± 0.31} & 69.10 ± 0.36 & \underline{\textbf{70.20 ± 0.20}} & \textbf{69.97 ± 0.42} \\
95 & 64.17 ± 0.23 & 67.54 ± 0.24 & 67.46 ± 0.31 & \textbf{67.70 ± 0.85} & 66.32 ± 0.43 & 1.00 ± 0.00 & 66.48 ± 0.16 & 67.63 ± 0.20 & 67.65 ± 0.46 & 66.68 ± 0.39 & \underline{\textbf{68.47 ± 0.31}} & 67.69 ± 0.33 \\
98 & 59.72 ± 0.69 & 63.51 ± 0.20 & 63.12 ± 0.38 & 64.51 ± 0.47 & 63.83 ± 0.43 & 1.00 ± 0.00 & 62.06 ± 0.45 & 64.63 ± 0.10 & 64.70 ± 0.37 & 64.01 ± 0.11 & \underline{\textbf{65.23 ± 0.14}} & 64.57 ± 0.12 \\
99 & 53.23 ± 0.08 & 49.15 ± 0.32 & 57.23 ± 0.45 & 59.43 ± 0.24 & 59.60 ± 0.43 & 1.00 ± 0.00 & 54.24 ± 0.66 & 58.02 ± 0.28 & 59.73 ± 0.33 & 59.79 ± 0.33 & \underline{\textbf{61.63 ± 0.59}} & 60.29 ± 0.14 \\
\bottomrule
\end{tabular}}
\end{sc}
\end{small}
\end{center}
\vskip -0.1in
\end{table}
We tested the CIFAR-100 dataset to extend our evaluation with a higher-complexity task. Table~\ref{tab:resnet18_cifar100_compressors_warmup_0} shows that although some of our proposed metrics remain top performers at high to extreme sparsities. Additionally, HP criteria take a significant distance from the rest in the extreme sparsity setting, suggesting that second-order information could capture relevant information to push the boundaries of PaI forward.
\begin{table}[h]
\caption{Performance of different pruning methods for CIFAR-100 on ResNet-18 with warmup. Bold and underlined values highlight the top performer. Bold values show methods whose confidence intervals overlap with the best performer.}
\label{tab:resnet18_cifar100_compressors_warmup_1}
\vskip 0.15in
\begin{center}
\begin{small}
\begin{sc}
\resizebox{\textwidth}{!}{%
\begin{tabular}{ccccccccccc}
\toprule
Sparsity (\%) & GN & SNIP & GraSP & SynFlow & \textbf{FD} & \textbf{FP} & \textbf{FTS} & \textbf{HD} & \textbf{HP} & \textbf{HTS}\\
\midrule
80 & 71.85 ± 0.31 & \textbf{72.43 ± 0.26} & 69.60 ± 0.36 & 1.00 ± 0.00 & 70.56 ± 0.47 & \textbf{72.52 ± 0.34} & \textbf{72.68 ± 0.12} & 72.22 ± 0.14 & \textbf{72.23 ± 0.38} & \underline{\textbf{72.69 ± 0.13}} \\
90 & 69.78 ± 0.19 & 70.36 ± 0.34 & 68.30 ± 0.52 & 1.00 ± 0.00 & 65.46 ± 0.82 & 70.04 ± 0.26 & \textbf{70.68 ± 0.25} & \textbf{70.77 ± 0.34} & \textbf{71.01 ± 0.54} & \underline{\textbf{71.19 ± 0.27}} \\
95 & 66.85 ± 0.49 & 68.04 ± 0.27 & 66.71 ± 0.75 & 1.00 ± 0.00 & 2.07 ± 1.86 & 65.20 ± 0.32 & 67.75 ± 0.14 & \underline{\textbf{69.45 ± 0.29}} & 68.61 ± 0.27 & 68.20 ± 0.16 \\
98 & 58.96 ± 0.91 & 63.73 ± 0.66 & 63.06 ± 0.26 & 1.00 ± 0.00 & 1.00 ± 0.00 & 62.84 ± 0.66 & 63.78 ± 0.21 & \underline{\textbf{66.38 ± 0.30}} & 65.65 ± 0.03 & 64.45 ± 0.15 \\
99 & 14.97 ± 10.97 & 59.56 ± 0.25 & 57.34 ± 0.74 & 1.00 ± 0.00 & 1.00 ± 0.00 & 29.25 ± 16.18 & 59.17 ± 0.44 & \underline{\textbf{62.08 ± 0.35}} & 61.23 ± 0.35 & 60.72 ± 0.29 \\
\bottomrule
\end{tabular}}
\end{sc}
\end{small}
\end{center}
\vskip -0.1in
\end{table}

\begin{figure*}[b]
    \centering
    \begin{subfigure}{0.85\textwidth}
        \includegraphics[width=\linewidth]{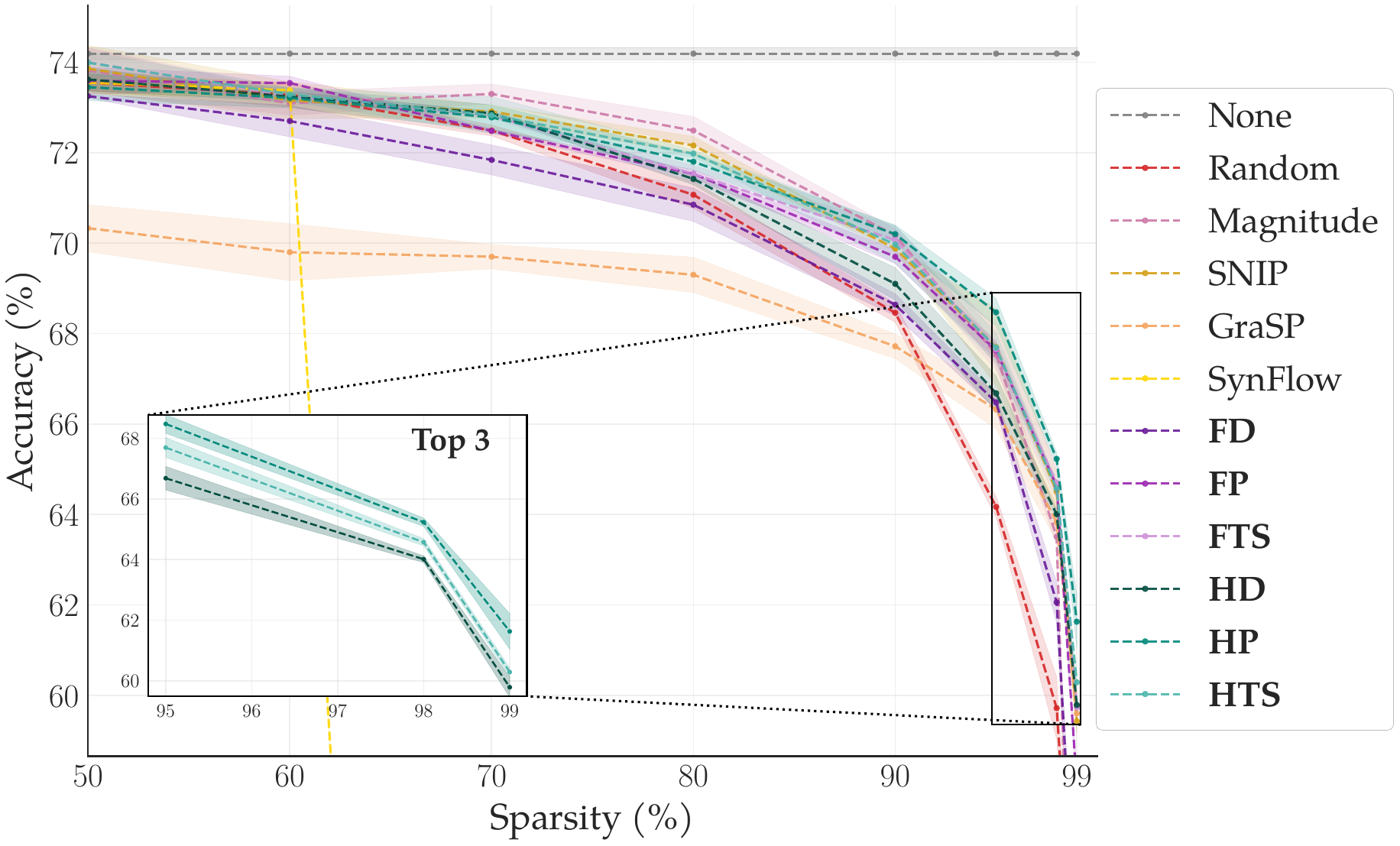}
        \caption{Without warmup}
        \label{fig:ResNet_18_CIFAR_100_w=0}
    \end{subfigure}
    \hfill
    \begin{subfigure}{0.85\textwidth}
        \includegraphics[width=\linewidth]{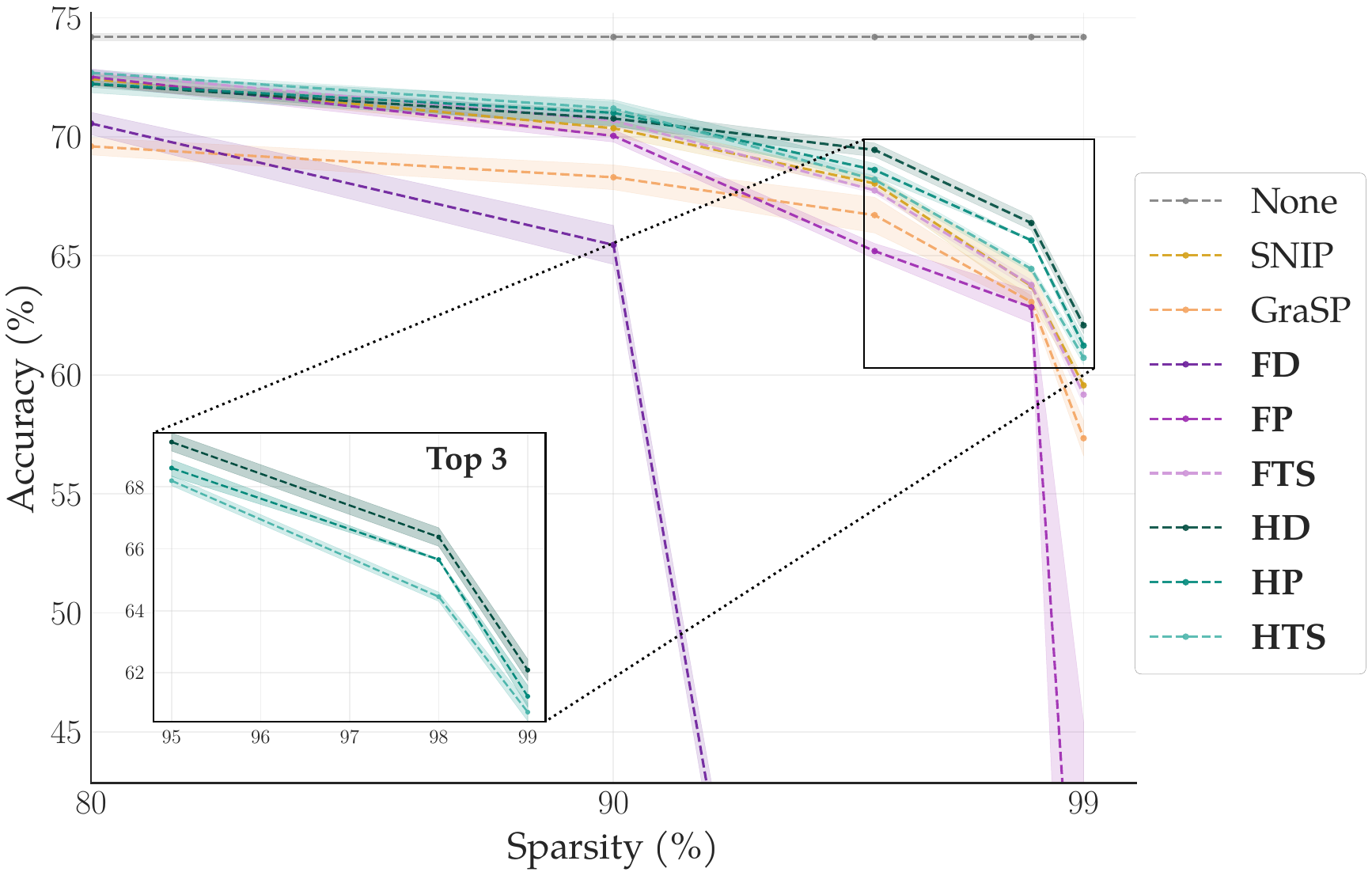}
        \caption{With warmup}
        \label{fig:ResNet_18_CIFAR_100_w=1}
    \end{subfigure}
    \caption{Test accuracy for ResNet-18 for the CIFAR-100 dataset with and without warmup. For extreme sparsities, we zoom in on the top 3 performers.}
    \label{fig:ResNet_18_CIFAR_100}
\end{figure*}

%% file: sections/appendix/09_cifar100_vgg_19.tex
\clearpage
\subsection{VGG19}
\label{appendix:vgg_cifar-100}
\begin{table}[h]
\caption{Performance of different pruning methods for CIFAR-100 on VGG-19 without warmup. Bold and underlined values highlight the top performer. Bold values show methods whose confidence intervals overlap with the best performer. Baseline accuracy (no pruning): 67.92 ± 0.86.}
\label{tab:vgg19_cifar100_compressors_warmup_0}
\vskip 0.15in
\begin{center}
\begin{small}
\begin{sc}
\resizebox{\textwidth}{!}{%
\begin{tabular}{ccccccccccccc}
\toprule
Sparsity (\%) & Random & Magnitude & GN & SNIP & GraSP & SynFlow & \textbf{FD} & \textbf{FP} & \textbf{FTS} & \textbf{HD} & \textbf{HP} & \textbf{HTS}\\
\midrule
10 & 66.94 ± 2.19 & 67.17 ± 0.05 & \textbf{68.96 ± 0.72} & 68.48 ± 0.64 & 66.19 ± 0.51 & \underline{\textbf{69.69 ± 0.19}} & 68.08 ± 0.55 & 68.42 ± 0.94 & 68.16 ± 1.02 & 67.39 ± 1.04 & 67.92 ± 0.18 & 68.54 ± 0.83 \\
20 & \textbf{67.33 ± 1.32} & \textbf{67.51 ± 0.50} & \textbf{68.28 ± 0.66} & \textbf{45.11 ± 38.24} & \textbf{68.17 ± 0.64} & 1.00 ± 0.00 & \textbf{45.11 ± 38.20} & \underline{\textbf{68.36 ± 0.97}} & \textbf{68.21 ± 0.34} & \textbf{67.39 ± 0.57} & \textbf{67.50 ± 0.67} & \textbf{67.82 ± 0.95} \\
30 & \textbf{44.64 ± 37.81} & \textbf{68.23 ± 0.68} & \textbf{68.17 ± 0.72} & \textbf{45.37 ± 38.44} & \textbf{67.99 ± 0.60} & 1.00 ± 0.00 & \textbf{67.61 ± 0.32} & \textbf{68.11 ± 0.71} & \textbf{68.00 ± 1.03} & \underline{\textbf{68.38 ± 0.70}} & \textbf{68.35 ± 0.36} & \textbf{67.94 ± 0.32} \\
40 & 67.24 ± 0.46 & \textbf{68.13 ± 0.43} & \textbf{46.31 ± 39.25} & \textbf{68.77 ± 0.43} & \textbf{68.03 ± 0.92} & 1.00 ± 0.00 & \textbf{68.38 ± 0.71} & \textbf{45.61 ± 38.65} & \underline{\textbf{68.81 ± 0.99}} & \textbf{68.25 ± 0.83} & \textbf{68.11 ± 0.75} & \textbf{46.18 ± 39.12} \\
50 & \textbf{67.16 ± 1.15} & \textbf{67.41 ± 1.33} & \textbf{67.52 ± 2.24} & \underline{\textbf{68.72 ± 0.80}} & \textbf{45.40 ± 38.45} & 1.00 ± 0.00 & \textbf{67.93 ± 2.00} & \textbf{67.62 ± 1.67} & \textbf{68.49 ± 0.76} & \textbf{68.67 ± 0.73} & \textbf{67.93 ± 0.44} & \textbf{66.12 ± 4.87} \\
60 & 67.57 ± 0.43 & 67.97 ± 0.64 & \textbf{68.68 ± 0.72} & \textbf{67.53 ± 1.91} & 67.90 ± 0.63 & 1.00 ± 0.00 & \textbf{68.22 ± 0.82} & \textbf{68.61 ± 0.85} & \textbf{68.21 ± 1.67} & \textbf{68.67 ± 0.37} & \underline{\textbf{68.76 ± 0.12}} & \textbf{68.30 ± 1.25} \\
70 & 67.30 ± 0.06 & \textbf{68.58 ± 1.02} & 67.19 ± 0.04 & 66.23 ± 1.13 & \textbf{67.72 ± 1.34} & 1.00 ± 0.00 & 66.12 ± 1.51 & 67.03 ± 0.16 & 66.68 ± 0.40 & \underline{\textbf{68.79 ± 0.16}} & 68.19 ± 0.20 & 67.10 ± 1.40 \\
80 & 65.50 ± 1.46 & \underline{\textbf{68.36 ± 1.26}} & 63.31 ± 0.49 & 65.46 ± 0.36 & \textbf{67.63 ± 0.84} & 1.00 ± 0.00 & 1.00 ± 0.00 & 10.79 ± 16.96 & 64.67 ± 0.44 & \textbf{67.49 ± 0.32} & \textbf{67.71 ± 0.50} & 63.72 ± 0.74 \\
90 & 63.92 ± 1.26 & \underline{\textbf{68.48 ± 0.42}} & 1.00 ± 0.00 & 1.00 ± 0.00 & 51.43 ± 2.00 & 1.00 ± 0.00 & 1.00 ± 0.00 & 1.00 ± 0.00 & 9.38 ± 14.51 & 1.00 ± 0.00 & 61.08 ± 2.03 & 1.00 ± 0.00 \\
95 & \underline{\textbf{60.28 ± 0.49}} & 1.00 ± 0.00 & 1.00 ± 0.00 & 1.00 ± 0.00 & 1.00 ± 0.00 & 1.00 ± 0.00 & 1.00 ± 0.00 & 1.00 ± 0.00 & 1.00 ± 0.00 & 1.00 ± 0.00 & 1.00 ± 0.00 & 1.00 ± 0.00 \\
98 & \underline{\textbf{55.46 ± 0.69}} & 1.00 ± 0.00 & 1.00 ± 0.00 & 1.00 ± 0.00 & 1.00 ± 0.00 & 1.00 ± 0.00 & 1.00 ± 0.00 & 1.00 ± 0.00 & 1.00 ± 0.00 & 1.00 ± 0.00 & 1.00 ± 0.00 & 1.00 ± 0.00 \\
99 & \underline{\textbf{50.41 ± 0.47}} & 1.00 ± 0.00 & 1.00 ± 0.00 & 1.00 ± 0.00 & 1.00 ± 0.00 & 1.00 ± 0.00 & 1.00 ± 0.00 & 1.00 ± 0.00 & 1.00 ± 0.00 & 1.00 ± 0.00 & 1.00 ± 0.00 & 1.00 ± 0.00 \\
\bottomrule
\end{tabular}}
\end{sc}
\end{small}
\end{center}
\vskip -0.1in
\end{table}
\begin{table}[h]
\caption{Performance of different pruning methods for CIFAR-100 on VGG-19 with warmup. Bold and underlined values highlight the top performer. Bold values show methods whose confidence intervals overlap with the best performer.}
\label{tab:vgg19_cifar100_compressors_warmup_1}
\vskip 0.15in
\begin{center}
\begin{small}
\begin{sc}
\resizebox{\textwidth}{!}{%
\begin{tabular}{ccccccccccc}
\toprule
Sparsity (\%) & GN & SNIP & GraSP & SynFlow & \textbf{FD} & \textbf{FP} & \textbf{FTS} & \textbf{HD} & \textbf{HP} & \textbf{HTS}\\
\midrule
80 & 23.60 ± 39.14 & \textbf{68.86 ± 0.91} & 66.59 ± 0.27 & 1.00 ± 0.00 & \textbf{39.33 ± 34.18} & \underline{\textbf{69.39 ± 0.41}} & 67.86 ± 0.49 & \textbf{68.00 ± 1.19} & 66.04 ± 0.66 & \textbf{45.56 ± 38.60} \\
90 & \underline{\textbf{69.32 ± 0.72}} & \textbf{68.36 ± 1.01} & \textbf{43.74 ± 37.01} & 1.00 ± 0.00 & 1.00 ± 0.00 & 1.00 ± 0.00 & \textbf{69.05 ± 0.92} & 66.94 ± 1.28 & 67.31 ± 0.75 & \textbf{68.95 ± 0.65} \\
95 & \underline{\textbf{68.49 ± 0.58}} & \textbf{68.23 ± 0.72} & 65.87 ± 0.92 & 1.00 ± 0.00 & 1.00 ± 0.00 & 1.00 ± 0.00 & \textbf{68.11 ± 0.54} & 64.62 ± 2.28 & \textbf{67.39 ± 1.06} & \textbf{68.19 ± 0.43} \\
98 & 51.50 ± 8.64 & 23.12 ± 26.58 & \textbf{64.99 ± 0.18} & 1.00 ± 0.00 & 1.00 ± 0.00 & 1.00 ± 0.00 & 1.00 ± 0.00 & \textbf{65.06 ± 1.14} & \underline{\textbf{65.41 ± 0.58}} & 42.73 ± 15.33 \\
99 & 1.00 ± 0.00 & 1.00 ± 0.00 & 38.84 ± 32.86 & 1.00 ± 0.00 & 1.00 ± 0.00 & 1.00 ± 0.00 & 1.00 ± 0.00 & \underline{\textbf{63.49 ± 0.29}} & 52.11 ± 5.11 & 1.00 ± 0.00 \\
\bottomrule
\end{tabular}}
\end{sc}
\end{small}
\end{center}
\vskip -0.1in
\end{table}
The results in VGG19 with CIFAR-100 exhibit a similar trend to those observed in CIFAR-10. Table~\ref{tab:vgg19_cifar100_compressors_warmup_0} shows the occurrence of layer collapse in extreme sparsities when no warm-up is applied, leading to a significant drop in accuracy. Introducing a simple warm-up phase effectively resolves this issue as seen in Table~\ref{tab:vgg19_cifar100_compressors_warmup_1}, restoring the pruning performance in all evaluated criteria.

%% file: sections/appendix/11_tinyimagenet_resnet_18.tex
\newpage
\section{Results TinyImageNet}
\subsection{ResNet18}
\label{appendix:TinyImageNet_ResNet18}
\begin{table}[h]
\caption{Performance of different pruning methods for TinyImageNet on ResNet-18 without warmup. Bold and underlined values highlight the top performer. Bold values indicate methods whose confidence intervals overlap with those of the best performer. Baseline accuracy (no pruning): 60.63 ± 0.12.}
\label{tab:resnet18_tinyimagenet_tinyimagenet_compressors_warmup_0}
\vskip 0.15in
\begin{center}
\begin{small}
\begin{sc}
\resizebox{\textwidth}{!}{%
\begin{tabular}{ccccccccccccc}
\toprule
Sparsity (\%) & Random & Magnitude & GN & SNIP & GraSP & SynFlow & \textbf{FD} & \textbf{FP} & \textbf{FTS} & \textbf{HD} & \textbf{HP} & \textbf{HTS}\\
\midrule
10 & 60.23 ± 0.17 & \textbf{60.54 ± 0.25} & \textbf{60.34 ± 0.14} & \underline{\textbf{60.83 ± 0.36}} & 59.23 ± 0.23 & \textbf{60.28 ± 0.22} & \textbf{60.41 ± 0.31} & \textbf{60.30 ± 0.23} & \textbf{60.58 ± 0.22} & \textbf{60.64 ± 0.15} & \textbf{60.48 ± 0.43} & \textbf{60.66 ± 0.20} \\
20 & \textbf{60.45 ± 0.33} & \textbf{60.30 ± 0.26} & \textbf{60.17 ± 0.21} & \textbf{60.48 ± 0.31} & 58.68 ± 0.70 & 59.93 ± 0.17 & \textbf{60.19 ± 0.38} & \underline{\textbf{60.58 ± 0.20}} & \textbf{60.33 ± 0.36} & \textbf{60.53 ± 0.24} & \textbf{60.37 ± 0.10} & \textbf{60.24 ± 0.51} \\
30 & \textbf{60.06 ± 0.19} & \textbf{59.88 ± 0.31} & \textbf{59.97 ± 0.24} & \textbf{60.03 ± 0.15} & 57.95 ± 0.53 & 59.65 ± 0.25 & 59.87 ± 0.17 & \textbf{60.17 ± 0.12} & \textbf{60.14 ± 0.19} & \textbf{59.92 ± 0.30} & \underline{\textbf{60.33 ± 0.25}} & \textbf{60.00 ± 0.34} \\
40 & \textbf{60.21 ± 0.05} & \textbf{60.16 ± 0.16} & \textbf{59.97 ± 0.25} & \textbf{60.00 ± 0.27} & 57.77 ± 0.61 & \textbf{59.79 ± 0.41} & \textbf{59.66 ± 0.56} & 59.82 ± 0.15 & \textbf{59.99 ± 0.26} & \underline{\textbf{60.22 ± 0.03}} & \textbf{60.15 ± 0.04} & \textbf{59.99 ± 0.37} \\
50 & \textbf{60.12 ± 0.08} & \textbf{59.97 ± 0.40} & \textbf{59.86 ± 0.24} & 59.73 ± 0.22 & 57.73 ± 0.19 & 59.29 ± 0.12 & 59.78 ± 0.16 & 59.71 ± 0.12 & \textbf{59.89 ± 0.38} & \textbf{59.87 ± 0.40} & \underline{\textbf{60.18 ± 0.10}} & \textbf{60.05 ± 0.19} \\
60 & \underline{\textbf{59.96 ± 0.34}} & \textbf{59.82 ± 0.13} & 59.17 ± 0.27 & 59.42 ± 0.14 & 57.77 ± 0.52 & 57.86 ± 0.21 & 58.90 ± 0.71 & \textbf{59.27 ± 0.43} & 59.40 ± 0.21 & \textbf{59.66 ± 0.33} & \textbf{59.64 ± 0.07} & \textbf{59.59 ± 0.37} \\
70 & \textbf{59.37 ± 0.86} & \underline{\textbf{59.64 ± 0.15}} & 58.69 ± 0.23 & 59.02 ± 0.30 & 57.21 ± 0.64 & 0.50 ± 0.00 & 58.46 ± 0.38 & 58.86 ± 0.15 & 59.10 ± 0.16 & 59.14 ± 0.20 & \textbf{59.52 ± 0.23} & 59.09 ± 0.12 \\
80 & 58.40 ± 0.11 & \underline{\textbf{59.28 ± 0.42}} & 57.85 ± 0.21 & \textbf{58.23 ± 0.70} & 56.10 ± 0.27 & 0.50 ± 0.00 & 56.95 ± 0.32 & 57.98 ± 0.09 & 58.24 ± 0.24 & \textbf{58.76 ± 0.28} & \textbf{58.93 ± 0.24} & 58.43 ± 0.16 \\
90 & 57.16 ± 0.19 & 54.92 ± 0.15 & 55.52 ± 0.44 & 56.57 ± 0.30 & 54.47 ± 0.32 & 0.50 ± 0.00 & 54.77 ± 0.44 & 55.88 ± 0.30 & 56.57 ± 0.06 & 57.19 ± 0.16 & \underline{\textbf{57.83 ± 0.12}} & 57.04 ± 0.34 \\
95 & 54.39 ± 0.29 & 52.30 ± 0.06 & 51.69 ± 0.41 & 53.28 ± 0.04 & 52.37 ± 0.62 & 0.50 ± 0.00 & 50.56 ± 0.29 & 52.90 ± 0.33 & 53.37 ± 0.15 & \textbf{54.45 ± 0.52} & \underline{\textbf{55.37 ± 0.40}} & 54.27 ± 0.36 \\
98 & 47.38 ± 0.14 & 44.89 ± 0.31 & 44.43 ± 0.65 & 47.54 ± 0.25 & 46.38 ± 0.29 & 0.50 ± 0.00 & 42.97 ± 0.38 & 46.62 ± 0.06 & 47.50 ± 0.43 & 49.56 ± 0.16 & \underline{\textbf{51.05 ± 0.16}} & 48.30 ± 0.44 \\
99 & 39.81 ± 0.66 & 32.90 ± 0.32 & 37.61 ± 0.39 & 41.12 ± 0.23 & 40.25 ± 0.78 & 0.50 ± 0.00 & 36.66 ± 0.24 & 40.46 ± 0.05 & 41.40 ± 0.36 & 44.12 ± 0.20 & \underline{\textbf{44.51 ± 0.08}} & 41.93 ± 0.72 \\
\bottomrule
\end{tabular}}
\end{sc}
\end{small}
\end{center}
\vskip -0.1in
\end{table}
We tested the TinyImageNet dataset to extend our evaluation with a higher-complexity task. Table~\ref{tab:resnet18_tinyimagenet_tinyimagenet_compressors_warmup_0} and Table~\ref{tab:resnet18_tinyimagenet_tinyimagenet_compressors_warmup_1} shows that the metrics based on the Hutchinson approximation demonstrate a clear dominance in a more complex task, HP being the best-performing metric across the sparsity spectrum.
\begin{table}[h]
\caption{Performance of different pruning methods for TinyImageNet on ResNet-18 with warmup. Bold and underlined values highlight the top performer. Bold values indicate methods whose confidence intervals overlap with those of the best performer.}
\label{tab:resnet18_tinyimagenet_tinyimagenet_compressors_warmup_1}
\vskip 0.15in
\begin{center}
\begin{small}
\begin{sc}
\resizebox{\textwidth}{!}{%
\begin{tabular}{ccccccccccc}
\toprule
Sparsity (\%) & GN & SNIP & GraSP & SynFlow & \textbf{FD} & \textbf{FP} & \textbf{FTS} & \textbf{HD} & \textbf{HP} & \textbf{HTS}\\
\midrule
80 & 57.13 ± 0.30 & 58.48 ± 0.44 & 50.99 ± 0.77 & 0.50 ± 0.00 & 57.26 ± 0.29 & \textbf{58.85 ± 0.27} & 58.69 ± 0.21 & \textbf{59.25 ± 0.02} & \underline{\textbf{59.49 ± 0.39}} & \textbf{59.28 ± 0.11} \\
90 & 53.82 ± 0.76 & 56.33 ± 0.29 & 51.16 ± 0.73 & 0.50 ± 0.00 & 51.95 ± 1.01 & 55.50 ± 0.32 & 56.17 ± 0.30 & \textbf{58.36 ± 0.45} & \underline{\textbf{58.50 ± 0.36}} & 57.58 ± 0.27 \\
95 & 48.37 ± 0.90 & 52.28 ± 0.11 & 49.01 ± 1.02 & 0.50 ± 0.00 & 30.99 ± 13.47 & 50.30 ± 0.55 & 52.58 ± 0.48 & \underline{\textbf{56.58 ± 0.11}} & 56.00 ± 0.21 & 53.95 ± 0.14 \\
98 & 33.66 ± 4.50 & 44.98 ± 0.02 & 41.49 ± 1.10 & 0.50 ± 0.00 & 0.50 ± 0.00 & 44.76 ± 0.49 & 44.93 ± 0.45 & \underline{\textbf{52.07 ± 0.61}} & 48.94 ± 0.13 & 46.92 ± 0.57 \\
99 & 11.44 ± 8.65 & 39.67 ± 0.67 & 34.43 ± 1.94 & 0.50 ± 0.00 & 0.50 ± 0.00 & 34.56 ± 1.48 & 39.33 ± 0.18 & \underline{\textbf{45.75 ± 0.42}} & 42.32 ± 0.18 & 42.02 ± 0.23 \\
\bottomrule
\end{tabular}}
\end{sc}
\end{small}
\end{center}
\vskip -0.1in
\end{table}

\newpage
\begin{figure*}
    \centering
    \begin{subfigure}{0.85\textwidth}
        \includegraphics[width=\linewidth]{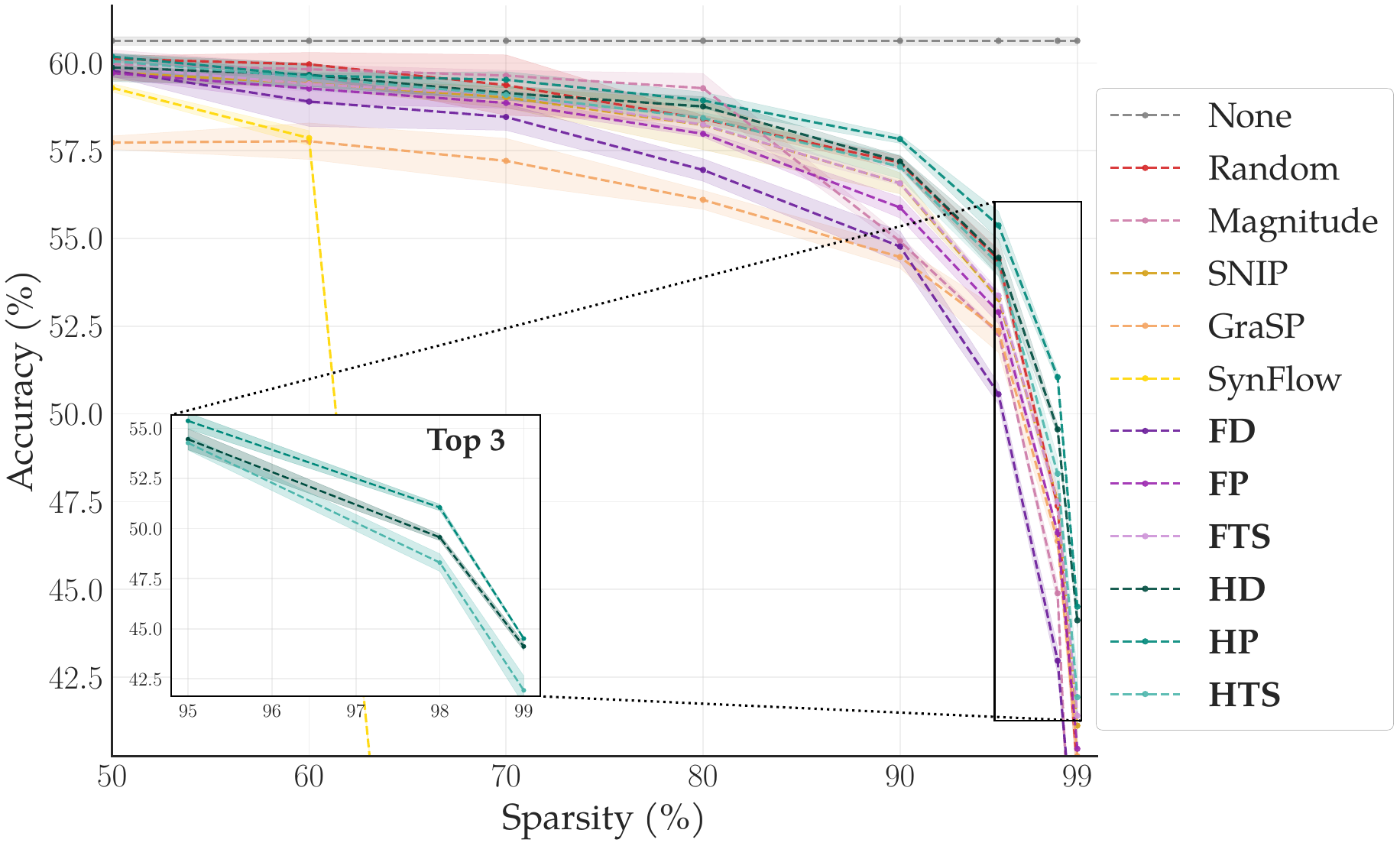}
        \caption{Without warmup}
        \label{fig:ResNet_18_TinyImageNet_w=0}
    \end{subfigure}
    \hfill
    \begin{subfigure}{0.85\textwidth}
        \includegraphics[width=\linewidth]{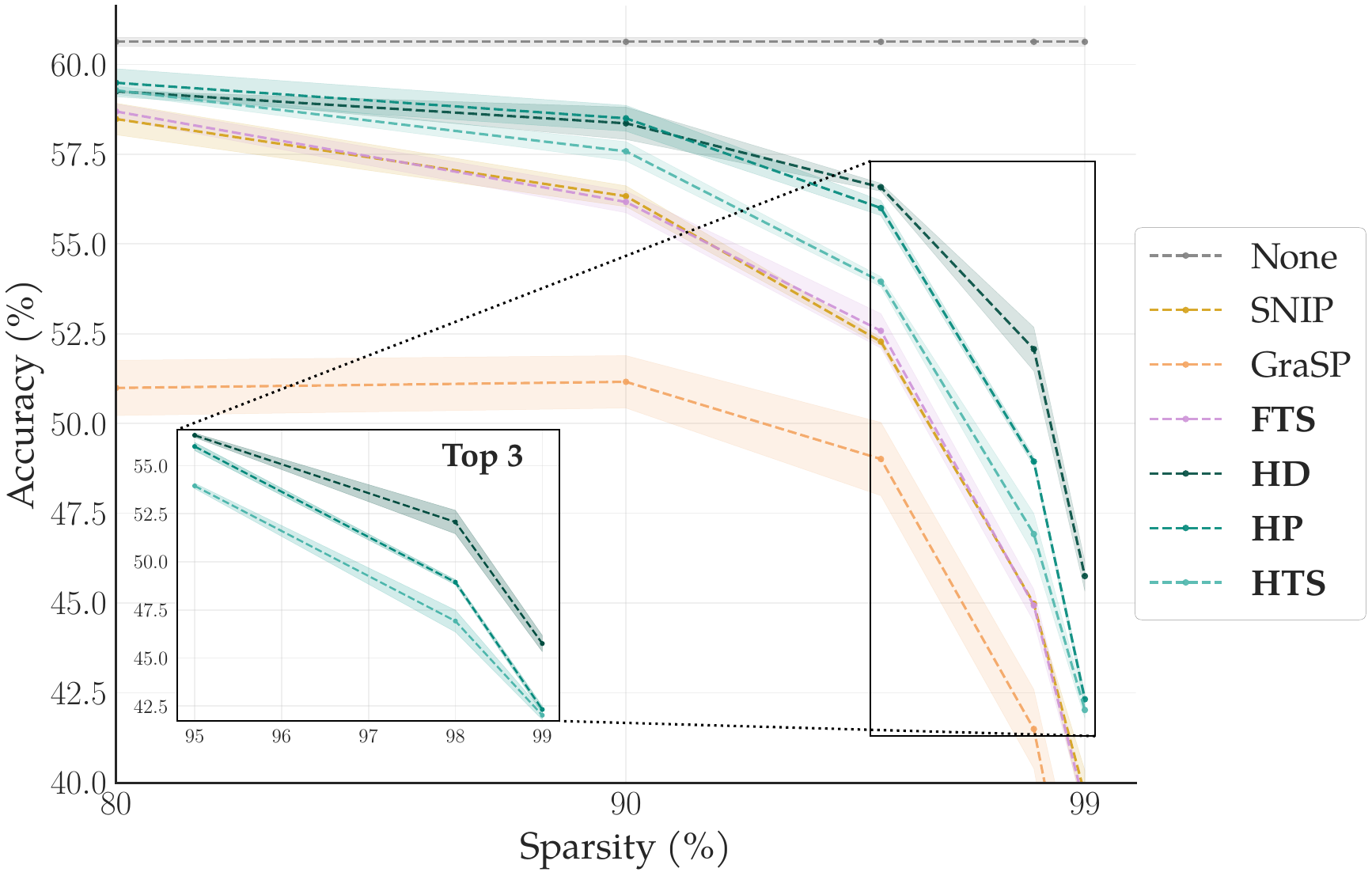}
        \caption{With warmup}
        \label{fig:ResNet_18_TinyImageNet_w=1}
    \end{subfigure}
    \caption{Test accuracy for ResNet-18 for the TinyImagenet dataset with and without warmup. For extreme sparsities, we zoom in on the top 3 performers.}
    \label{fig:ResNet_18_TinyImageNet}
\end{figure*}
\clearpage

%% file: sections/appendix/12_imagenet_resnet_50.tex
\newpage
\section{ImageNet}
\label{appendix:ImageNet}

\begin{table}[h]
\caption{Performance of different pruning methods for ImageNet on ResNet50. Bold and underlined values highlight the top performer. Bold values indicate methods whose confidence intervals overlap with those of the best performer. Baseline accuracy (no pruning): 73.95 ± 0.18.}
\label{tab:resnet50_imagenet_compressors_warmup_0}
\vskip 0.15in
\begin{center}
\begin{small}
\begin{sc}
\resizebox{\textwidth}{!}{%
\begin{tabular}{ccccccccccc}
\toprule
Sparsity (\%) & Random & Magnitude & SNIP & GraSP & \textbf{FTS} & \textbf{HD} & \textbf{HP} & \textbf{HTS}\\
\midrule
60 & 71.61 ± 0.48 & \underline{\textbf{72.70 ± 0.48}} & 72.03 ± 0.17 & 70.24 ± 0.21 & \textbf{72.33 ± 0.40} & \textbf{71.76 ± 0.83} & 71.08 ± 0.79 & 72.12 ± 0.09 \\
70 & 70.25 ± 0.36 & \underline{\textbf{71.47 ± 0.07}} & 70.69 ± 0.25 & 68.55 ± 1.26 & 71.09 ± 0.15 & 70.91 ± 0.02 & 70.94 ± 0.17 & \textbf{71.06 ± 0.47} \\
80 & 68.84 ± 0.22 & \textbf{69.64 ± 0.29} & \textbf{69.63 ± 0.23} & 67.63 ± 0.52 & \underline{\textbf{69.68 ± 0.15}} & 68.83 ± 0.12 & 68.90 ± 0.18 & \textbf{69.45 ± 0.13} \\
90 & 64.21 ± 0.46 & 63.72 ± 0.43 & \textbf{65.47 ± 0.50} & 62.52 ± 0.50 & \textbf{65.75 ± 0.04} & \textbf{65.08 ± 0.33} & \textbf{66.01 ± 0.03} & \underline{\textbf{66.10 ± 0.85}} \\
95 & 59.17 ± 0.71 & 48.08 ± 9.25 & 61.88 ± 0.36 & 57.55 ± 1.09 & 61.50 ± 0.06 & 59.53 ± 0.26 & \textbf{60.94 ± 2.05} & \underline{\textbf{62.87 ± 0.23}} \\
98 & 49.03 ± 1.12 & 0.10 ± 0.00 & 54.36 ± 0.89 & 50.19 ± 0.72 & 54.25 ± 1.01 & 49.71 ± 1.65 & \textbf{55.55 ± 0.07} & \underline{\textbf{56.09 ± 0.73}} \\
99 & 38.64 ± 0.63 & 0.10 ± 0.00 & 47.38 ± 0.06 & 42.07 ± 3.93 & \textbf{47.47 ± 1.03} & 32.79 ± 3.32 & \underline{\textbf{48.49 ± 0.50}} & \textbf{48.29 ± 1.68} \\
\bottomrule
\end{tabular}}
\end{sc}
\end{small}
\end{center}
\vskip -0.1in
\end{table}

\begin{figure}[b]
    \centering
    \includegraphics[width=0.85\textwidth]{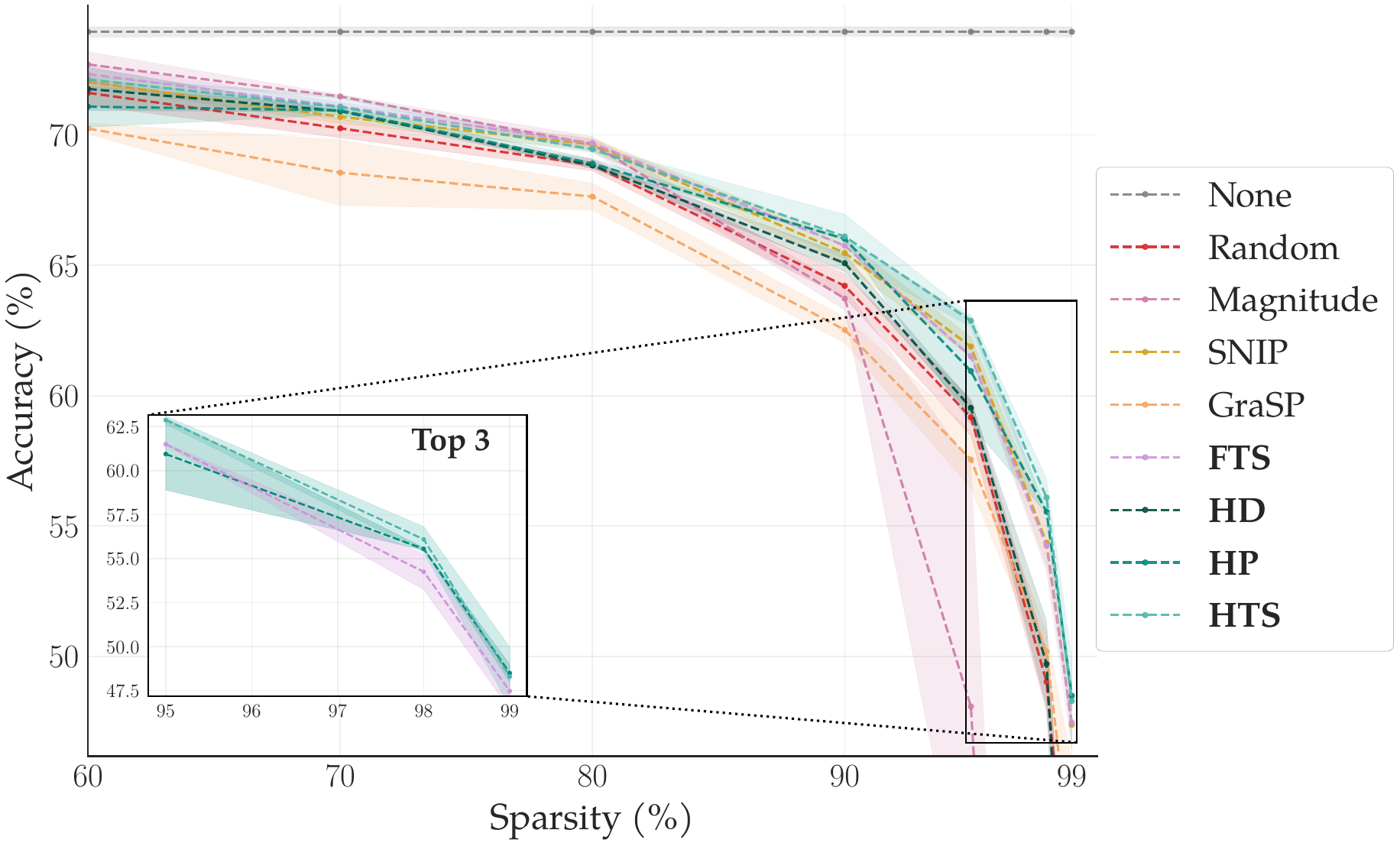}
    \caption{Test accuracy for ResNet50 for the Imagenet-1k dataset only without warmup. For extreme sparsities, we zoom in on the top 3 performers.}
    \label{fig:ResNet_50_ImageNet_w=0}
\end{figure}

%% file: sections/appendix/13_pat_resnet_18.tex
\newpage
\section{Prune After Training CIFAR-10}
\subsection{ResNet18}
\label{appendix:PAT_CIFAR10_ResNet18}
\begin{table}[h]
\caption{Performance of different pruning methods for CIFAR-10 on ResNet18 (PaT). Bold and underlined values highlight the top performer. Bold values show methods whose confidence intervals overlap with the best performer. Baseline accuracy (no pruning): 93.73 ± 0.20.}
\label{tab:resnet18_cifar10_pat_compressors}
\vskip 0.15in
\begin{center}
\begin{small}
\begin{sc}
\resizebox{\textwidth}{!}{%
\begin{tabular}{cccccccccccccc}
\toprule
Sparsity (\%) & Random & Magnitude & GN & SNIP & GraSP & SynFlow & \textbf{FD} & \textbf{FP} & \textbf{FTS} & \textbf{HD} & \textbf{HP} & \textbf{HTS}\\
\midrule
10 & 93.60 ± 0.06 & \textbf{93.84 ± 0.06} & \textbf{93.76 ± 0.13} & \textbf{93.80 ± 0.08} & 92.76 ± 0.05 & \textbf{93.88 ± 0.11} & \textbf{93.83 ± 0.16} & \textbf{93.80 ± 0.04} & \textbf{93.78 ± 0.07} & \textbf{93.86 ± 0.11} & \underline{\textbf{93.92 ± 0.08}} & \textbf{93.77 ± 0.11}  \\
20 & \textbf{93.60 ± 0.15} & \textbf{93.83 ± 0.06} & \textbf{93.83 ± 0.09} & \textbf{93.80 ± 0.06} & 91.85 ± 0.17 & \textbf{93.79 ± 0.14} & \textbf{93.76 ± 0.10} & \textbf{93.85 ± 0.07} & \textbf{93.83 ± 0.07} & \underline{\textbf{93.88 ± 0.24}} & \textbf{93.83 ± 0.09} & \textbf{93.79 ± 0.10}  \\
30 & 93.61 ± 0.09 & \textbf{93.82 ± 0.07} & \textbf{93.83 ± 0.16} & \textbf{93.81 ± 0.05} & 91.69 ± 0.50 & 93.49 ± 0.04 & \textbf{93.76 ± 0.11} & \textbf{93.79 ± 0.07} & \textbf{93.79 ± 0.06} & \textbf{93.63 ± 0.16} & \underline{\textbf{93.88 ± 0.15}} & \textbf{93.77 ± 0.05}  \\
40 & 93.39 ± 0.20 & \underline{\textbf{93.84 ± 0.06}} & \textbf{93.79 ± 0.14} & \textbf{93.76 ± 0.05} & 91.83 ± 0.40 & 93.49 ± 0.08 & \textbf{93.75 ± 0.10} & \textbf{93.77 ± 0.05} & \textbf{93.83 ± 0.04} & \textbf{93.74 ± 0.15} & \textbf{93.79 ± 0.17} & \textbf{93.81 ± 0.05}  \\
50 & 93.44 ± 0.20 & \underline{\textbf{93.86 ± 0.08}} & \textbf{93.77 ± 0.08} & \textbf{93.74 ± 0.05} & 91.47 ± 0.28 & 93.54 ± 0.14 & 93.60 ± 0.06 & \textbf{93.71 ± 0.12} & \textbf{93.77 ± 0.10} & 93.59 ± 0.18 & 93.58 ± 0.01 & \textbf{93.71 ± 0.15}  \\
60 & 93.19 ± 0.06 & \textbf{93.83 ± 0.09} & \textbf{93.84 ± 0.12} & \textbf{93.71 ± 0.13} & 91.34 ± 0.24 & \textbf{93.56 ± 0.21} & \textbf{93.84 ± 0.08} & \textbf{93.68 ± 0.07} & \textbf{93.75 ± 0.04} & \underline{\textbf{93.88 ± 0.18}} & \textbf{93.61 ± 0.09} & 93.59 ± 0.10  \\
70 & 92.85 ± 0.27 & \underline{\textbf{93.80 ± 0.04}} & \textbf{93.69 ± 0.12} & 93.49 ± 0.14 & 91.19 ± 0.10 & 10.00 ± 0.00 & \textbf{93.74 ± 0.14} & 93.50 ± 0.03 & 93.57 ± 0.09 & \textbf{93.69 ± 0.09} & 93.60 ± 0.15 & 93.49 ± 0.06  \\
80 & 91.07 ± 0.35 & 93.55 ± 0.07 & 93.60 ± 0.04 & \textbf{93.55 ± 0.21} & 91.29 ± 0.21 & 10.00 ± 0.00 & \textbf{93.58 ± 0.17} & \textbf{93.67 ± 0.07} & \textbf{93.50 ± 0.19} & 93.43 ± 0.13 & \underline{\textbf{93.72 ± 0.06}} & \textbf{93.60 ± 0.19}  \\
90 & 89.12 ± 0.30 & 92.95 ± 0.12 & \textbf{93.36 ± 0.20} & \textbf{93.30 ± 0.10} & 90.55 ± 0.39 & 10.00 ± 0.00 & \underline{\textbf{93.39 ± 0.15}} & \textbf{93.32 ± 0.18} & \textbf{93.18 ± 0.26} & 92.90 ± 0.26 & \textbf{93.37 ± 0.30} & \textbf{93.25 ± 0.16}  \\
95 & 87.53 ± 0.29 & 92.29 ± 0.11 & 92.36 ± 0.14 & \textbf{92.56 ± 0.19} & 89.95 ± 0.38 & 10.00 ± 0.00 & 92.19 ± 0.27 & \textbf{92.50 ± 0.17} & \textbf{92.69 ± 0.21} & 92.13 ± 0.24 & \underline{\textbf{92.70 ± 0.07}} & \textbf{92.56 ± 0.28} \\
98 & 84.45 ± 0.37 & 90.91 ± 0.17 & 90.46 ± 0.17 & 90.84 ± 0.30 & 88.46 ± 0.66 & 10.00 ± 0.00 & 90.14 ± 0.23 & 91.16 ± 0.15 & 91.07 ± 0.21 & 90.16 ± 0.07 & \underline{\textbf{91.41 ± 0.02}} & 90.81 ± 0.22 \\
99 & 81.01 ± 0.36 & \textbf{89.30 ± 0.52} & 88.70 ± 0.17 & \textbf{89.01 ± 0.55} & 87.19 ± 0.14 & 10.00 ± 0.00 & 87.85 ± 0.29 & \textbf{89.10 ± 0.33} & \textbf{89.10 ± 0.27} & 88.69 ± 0.42 & \underline{\textbf{89.52 ± 0.25}} & 88.87 ± 0.35 \\
\bottomrule
\end{tabular}}
\end{sc}
\end{small}
\end{center}
\vskip -0.1in
\end{table}
\begin{table}[h]
\caption{Performance delta (PaT - PaI with warmup) for CIFAR-10 on ResNet-18. Positive values indicate PaT performs better, negative values indicate PaI performs better.}
\label{tab:resnet18_cifar10_pat_vs_pbt_delta}
\vskip 0.15in
\begin{center}
\begin{small}
\begin{sc}
\resizebox{\textwidth}{!}{%
\begin{tabular}{ccccccccccccc}
\toprule
Sparsity (\%) & Random & Magnitude & GN & SNIP & GraSP & SynFlow & \textbf{FD} & \textbf{FP} & \textbf{FTS} & \textbf{HD} & \textbf{HP} & \textbf{HTS}\\
\midrule
80 & -1.57 ± 0.35 & 0.20 ± 0.07 & 0.28 ± 0.04 & 0.25 ± 0.21 & -0.44 ± 0.21 & 0.00 ± 0.00 & 0.06 ± 0.17 & 0.31 ± 0.07 & 0.03 ± 0.19 & 0.15 ± 0.13 & 0.40 ± 0.06 & 0.08 ± 0.19 \\
90 & -2.26 ± 0.30 & 0.09 ± 0.12 & 0.65 ± 0.20 & 0.18 ± 0.10 & -1.04 ± 0.39 & 0.00 ± 0.00 & 0.50 ± 0.15 & 0.23 ± 0.18 & 0.18 ± 0.26 & 0.26 ± 0.26 & 0.50 ± 0.30 & 0.10 ± 0.16 \\
95 & -2.06 ± 0.29 & 0.24 ± 0.11 & 0.20 ± 0.14 & 0.24 ± 0.19 & -0.94 ± 0.38 & 0.00 ± 0.00 & 12.61 ± 0.27 & 0.98 ± 0.17 & 0.42 ± 0.21 & -0.07 ± 0.24 & 0.52 ± 0.07 & 0.33 ± 0.28 \\
98 & -1.84 ± 0.37 & 2.46 ± 0.17 & 0.18 ± 0.17 & 0.55 ± 0.30 & -1.32 ± 0.66 & 0.00 ± 0.00 & 80.14 ± 0.23 & 1.85 ± 0.15 & 0.80 ± 0.21 & -1.08 ± 0.07 & 0.86 ± 0.02 & 0.56 ± 0.22 \\
99 & -1.91 ± 0.36 & 12.76 ± 0.52 & 28.63 ± 0.17 & 0.65 ± 0.55 & -0.79 ± 0.14 & 0.00 ± 0.00 & 77.85 ± 0.29 & 3.71 ± 0.33 & 1.08 ± 0.27 & -0.78 ± 0.42 & 0.73 ± 0.25 & 0.81 ± 0.35 \\
\bottomrule
\end{tabular}}
\end{sc}
\end{small}
\end{center}
\vskip -0.1in
\end{table}
Frankle et al.~\citep{frankle2020pruning} previously criticized PaI methods for consistently underperforming compared to magnitude PaT. However, their analysis did not assess how the PaI-designed criteria perform in the PaT setting. Table~\ref{tab:resnet18_cifar10_pat_compressors} shows that second-order-based criteria remain top performers across sparsities in the PaT setting. Table~\ref{tab:resnet18_cifar10_pat_vs_pbt_delta} presents the test accuracy delta between PaI and PaT settings. As we increase the sparsity, it is clear that the advantage of magnitude PaT comes at the expense of training a fully dense model and the required retraining.

%% file: sections/appendix/14_vit_structured.tex
\clearpage
\section{Pre-Finetune Pruning for Transformers}
\label{appendix:structured}

\textbf{Training Setup.}
All models were fine-tuned using the same training protocol and hyperparameters. As transformers are known to be data-hungry, we pruned the default initialization in Torch (from ImageNet-1K) before fine-tuning for 30 epochs, similar to the approach in \cite{dosovitskiy2021imageworth16x16words}. As sparsity increases, the model's size and FLOPs are drastically reduced. We emphasize that our goal in this experiment is not to optimize transformer compression performance, but to assess whether pruning-at-initialization metrics, originally designed for unstructured pruningm remain informative when applied to structured pruning decisions in transformer architectures, something demonstrated in \cite{sun2025optimalbrainapoptosis}.
\nl
\textbf{Implementation.}
Structured pruning was implemented using the library \href{https://github.com/VainF/Torch-Pruning}{torch-pruning} as done also in \cite{sun2025optimalbrainapoptosis}, which provides graph-based dependency tracking for structured parameter removal in PyTorch models. This framework enables pruning entire structured units while preserving computational consistency across dependent layers, which is particularly important for transformer architectures where attention and projection layers are tightly coupled. We employed per-layer structured pruning, ensuring that pruning decisions adhere to layer-wise constraints and avoid invalid graph states.
\nl
Pruning was applied at initialization to a pretrained \textit{ViT-B/16} model (5.61 billion parameters), before fine-tuning on CIFAR-10. For each sparsity level, the same global sparsity ratio was enforced across layers by pruning an equal proportion of structured units per layer; the encoder's multilayer perception (MLP) layers shrink at a squared sparsity ratio. All pruning criteria were used solely to rank structured units; the pruning procedure itself was identical across methods.
\nl
As unstructured PaI methods implicitly assume that unstructured sparsity is a suitable target for neural networks. However, this assumption becomes problematic in transformer-based architectures. Unstructured pruning removes individual weights and produces highly irregular sparse patterns, but transformer computation is dominated by dense linear algebra operations, so these patterns rarely lead to meaningful reductions in FLOPs or wall-clock inference time on standard hardware without specialized sparse engines. Furthermore, transformers exhibit strong inter-parameter dependencies due to tightly coupled components, such as multi-head attention and residual connections. Removing individual weights without accounting for these interactions can disrupt key network pathways and degrade performance unless carefully compensated for during training or fine-tuning. 
\nl
Prior work on pruning transformers also highlights that unstructured criteria often fail to align with structural redundancies in these models, while structured pruning that targets entire attention heads, blocks, or projection subspaces can better respect the model’s computational topology and yield more stable efficiency–performance trade-offs. These observations suggest that while our work centers on unstructured PaI, it is valuable to test whether our proposed pruning metrics can remain informative in a structured setting where decisions operate over larger, computationally meaningful units rather than individual weights.
\newpage
\begin{table}[t]
\caption{Summary results of structured pruning: we used different criteria to perform per-layer pruning for the model ViT-B/16 in the CIFAR-10 dataset. Bold values indicate the best average performance among three random seeds. Baseline accuracy (no pruning): 85.78 ± 1.56.}
\label{tab:tbd}
\vskip 0.15in
\begin{center}
\begin{small}
\begin{sc}
\resizebox{\textwidth}{!}{%
\begin{tabular}{ccccccccccccc}
\toprule
Sparsity (\%) & FLOPs & Runtime (min) & Random & Magnitude & SNIP & GraSP & \textbf{FD} & \textbf{FP} & \textbf{FTS} & \textbf{HD} & \textbf{HP} & \textbf{HTS}\\
\midrule
    20 & 4.49G & 55 & 85.75 ± 0.20 & 85.79 ± 0.36 & \textbf{86.34 ± 0.44} & 84.86 ± 0.52 & 86.07 ± 0.86 & 85.63 ± 0.66 & 85.77 ± 0.38 & 85.21 ± 0.64 & 85.28 ± 0.60 & 86.01 ± 0.48 \\
    36 & 3.59G & 44 & 85.18 ± 0.28 & 84.84 ± 0.46 & 85.43 ± 0.39 & 83.42 ± 1.81 & 85.31 ± 0.16 & 85.15 ± 0.45 & \textbf{85.45 ± 0.24} & 84.96 ± 0.65 & 84.99 ± 0.04 & 85.21 ± 0.42 \\
    52 & 2.69G & 38 & 83.75 ± 0.56 & 83.91 ± 0.09 & 83.82 ± 0.21 & 82.62 ± 0.79 & 83.88 ± 0.14 & 83.58 ± 0.57 & 83.83 ± 0.47 & 83.73 ± 1.16 & 83.75 ± 0.48 & \textbf{84.37 ± 0.28} \\
    64 & 2.02G & 29 & 83.31 ± 0.45 & 82.95 ± 0.29 & 82.91 ± 0.62 & 82.19 ± 0.59 & 82.75 ± 0.17 & 83.26 ± 0.10 & 82.80 ± 0.21 & 82.32 ± 1.67 & 83.02 ± 0.25 & \textbf{83.60 ± 0.35} \\
    75 & 1.40G & 22 & 82.07 ± 0.50 & \textbf{82.29 ± 0.61} & 81.89 ± 0.57 & 81.44 ± 0.35 & 81.86 ± 0.17 & 81.64 ± 0.73 & 81.70 ± 0.59 & 82.07 ± 1.05 & 81.95 ± 0.51 & 81.88 ± 0.65 \\
    84 & 0.90G & 18 & 76.79 ± 0.32 & 74.31 ± 1.28 & \textbf{79.34 ± 0.55} & 76.10 ± 2.20 & 78.87 ± 0.56 & 78.67 ± 0.42 & 79.15 ± 0.24 & 78.76 ± 0.43 & 78.41 ± 0.20 & 78.71 ± 0.80 \\
    91 & 0.50G & 13 & 75.18 ± 0.69 & 69.23 ± 0.91 & 77.20 ± 0.73 & 76.05 ± 0.37 & 77.31 ± 0.67 & 76.37 ± 0.95 & 77.10 ± 0.89 & 76.78 ± 0.99 & 76.69 ± 0.70 & \textbf{77.62 ± 0.73} \\
    99 & 0.06G & 10 & 72.57 ± 0.08 & 69.79 ± 1.14 & 73.00 ± 0.57 & 73.42 ± 0.62 & 73.28 ± 0.24 & 73.47 ± 0.97 & 73.32 ± 0.74 & 72.90 ± 1.24 & 73.18 ± 0.82 & \textbf{73.48 ± 1.39} \\
\bottomrule
\end{tabular}}
\end{sc}
\end{small}
\end{center}
\vskip -0.1in
\end{table}

\begin{figure}
    \centering
    \includegraphics[width=0.85\linewidth]{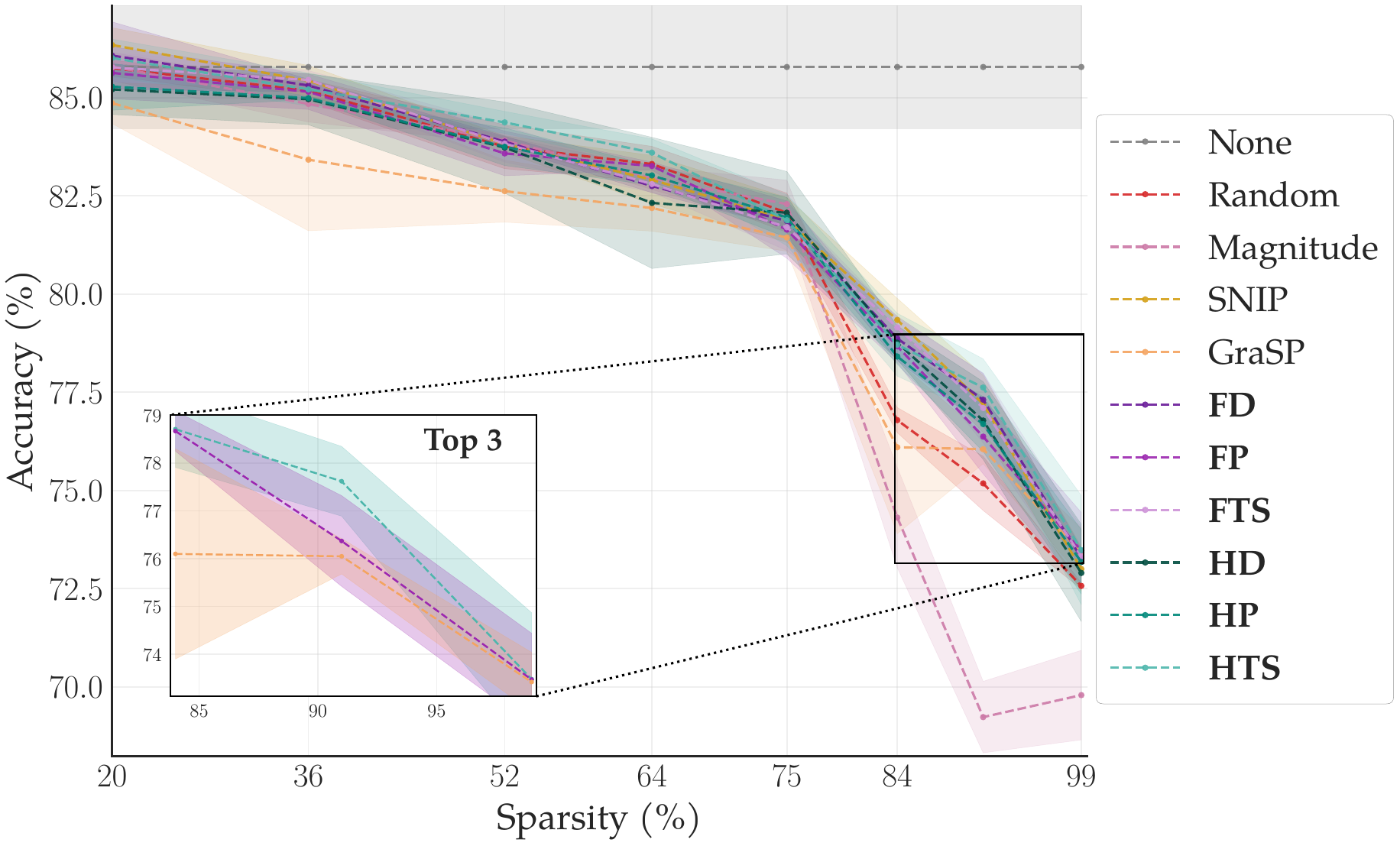}
    \caption{Structured pruning of ViT-B/16 in CIFAR-10: Test accuracy across sparsities for all the importance-based metrics.}
    \label{fig:vit_cifar10_lineplot}
    \label{fig:placeholder}
\end{figure}

%% file: sections/appendix/15_criteria_vs_magnitude.tex
\clearpage
\section{Comparison of our criteria with magnitude-based pruning}
\label{appendix:parameter_selection_comparison}
\begin{figure}[htp]
    \centering
    \includegraphics[width=0.9\linewidth]{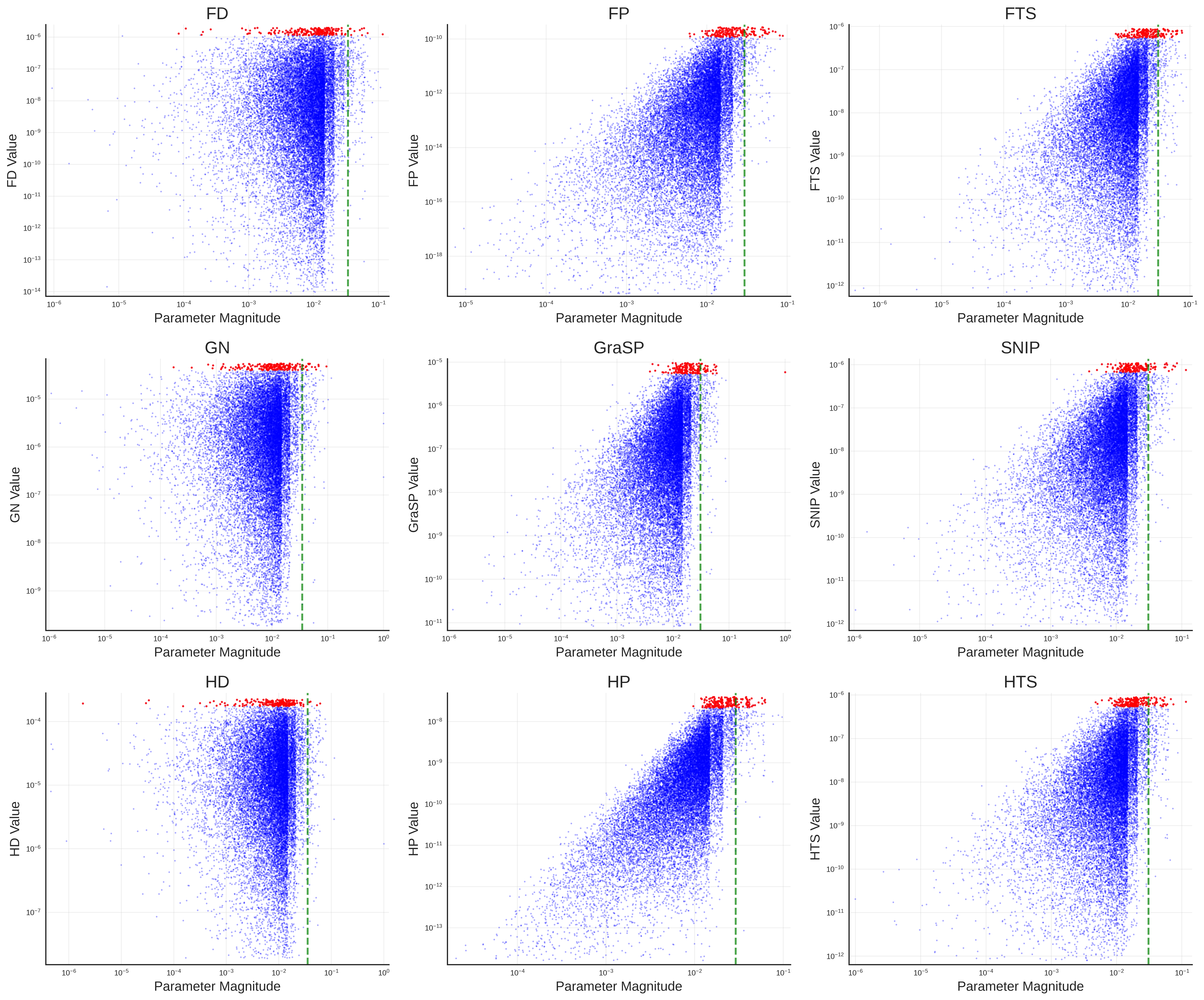}
    \caption{Proposed criteria vs. Magnitude parameter selection for 99\% sparsity (ResNet18, CIFAR-10, Seed 0)} 
    \label{fig:our_criterion_vs_magnitude}
\end{figure}
Figure \ref{fig:our_criterion_vs_magnitude} illustrates the relationship between parameter magnitude and different sensitivity-based pruning metrics. Each point represents a model parameter, with red points indicating the top-$1\%$ parameters selected for retention by each criterion. The green dashed line marks the 99th percentile of parameter magnitudes. A key observation is that the success of data-based methods relies on the ability to consider small-magnitude parameters that are discarded by magnitude-based pruning.